\pdfoutput=1

\documentclass[11pt]{article}

\usepackage[final, nohyperref]{acl}
\usepackage{times}
\usepackage{multicol}
\usepackage{latexsym}
\usepackage{booktabs}
\usepackage{amssymb}
\usepackage{amsmath}
\usepackage{bbding}
\usepackage{subfig}
\usepackage{graphicx}
\usepackage{multirow}
\usepackage[ruled,linesnumbered]{algorithm2e}
\usepackage{colortbl}
\usepackage{rotating}
\usepackage[inkscapelatex=false]{svg}
\usepackage{url}
\usepackage{diagbox}
\usepackage{pifont}
\usepackage{listings}
\definecolor{citec}{HTML}{882810}
\definecolor{refc}{HTML}{4658cf}
\definecolor{urlc}{HTML}{39a85c}
\usepackage[colorlinks,
            linkcolor=refc,
            anchorcolor=refc,
            urlcolor=refc,
            citecolor=refc]{hyperref}
\usepackage{orcidlink}
\usepackage[T1]{fontenc}

\usepackage[utf8]{inputenc}

\usepackage{microtype}

\usepackage{inconsolata}

\newcommand{\M}{Hidden Calibration}
\newcommand{\SER}{\texttt{SemEval 2014-Task 4 Restaurant}}
\newcommand{\SEL}{\texttt{SemEval 2014-Task 4 Laptops}}
\newcommand{\OPT}{OPT-2.7B}
\newcommand{\LLAMA}{Llama 2-6.9B}
\newcommand{\GPT}{GPT2-XL}
\renewcommand{\paragraph}{\vspace{0.25em}\noindent\textbf}

\title{Token-based Decision Criteria Are Suboptimal in In-context Learning}

\author{Hakaze Cho\orcidlink{0000-0002-7127-1954}${}^{1,\text{\ding{73}}}$\phantom{11 111}Yoshihiro Sakai${}^{1}$\phantom{111111}Mariko Kato${}^{1}$\\ 
\textbf{Kenshiro Tanaka}${}^{1}$\phantom{1111111}\textbf{Akira Ishii}${}^{1}$\phantom{11111111}\textbf{Naoya Inoue}${}^{1,2}$\phantom{1}\\
${}^{1}$Japan Advanced Institute of Science and Technology\phantom{1111}${}^{2}$RIKEN\\
${}^{\text{\ding{73}}}$\hyperref[Appendix:ACS]{Primary Contributor}, Correspondence to: \texttt{yfzhao@jaist.ac.jp}}

\begin{document}
\maketitle
\begin{abstract}
\textbf{I}n-\textbf{C}ontext \textbf{L}earning (ICL) typically utilizes classification criteria from output probabilities of manually selected label tokens. However, we argue that such token-based classification criteria lead to suboptimal decision boundaries, despite delicate calibrations through translation and constrained rotation applied. To address this problem, we propose \M, which renounces token probabilities and uses the nearest centroid classifier on the LM's last hidden states. In detail, we assign the label of the nearest centroid previously estimated from a calibration set to the test sample as the predicted label. Our experiments on 6 models and 10 classification datasets indicate that \M~consistently outperforms current token-based baselines by about 20\%$\sim$50\%, achieving a strong state-of-the-art in ICL. Our further analysis demonstrates that \M~finds better classification criteria with less inter-class overlap, and LMs provide linearly separable intra-class clusters with the help of demonstrations, which supports \M{} and gives new insights into the principle of ICL. Our official code implementation can be found at~\url{https://github.com/hc495/Hidden_Calibration}.
\end{abstract}


\section{Introduction}

\begin{figure}[t]
    \centering
    \includegraphics[width=1\linewidth]{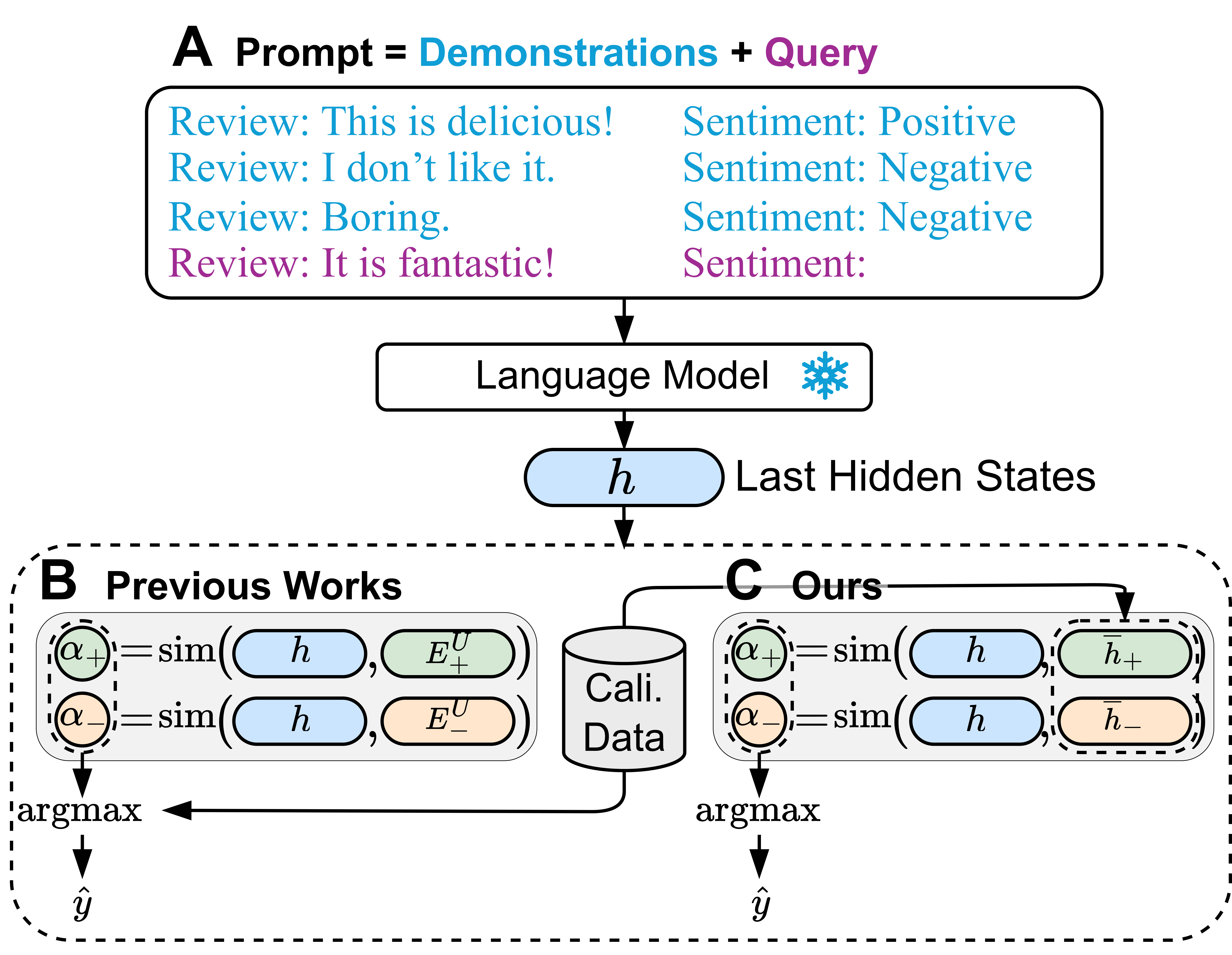}
    \caption{In an ICL diagram, \textbf{A.} The prompt of ICL consists of a concatenation of \textcolor[HTML]{0f9ed5}{demonstrations} and \textcolor[HTML]{a02b93}{a query}. LMs encode the prompt into the last hidden state $h$, then \textbf{B.} (previous works) use the un-embedding vectors of the label tokens ($E^U_+$, $E^U_-$) to decode the $h$ to prediction $\hat{y}$, then calibrations are used to adjust the predicted logits. \textbf{C.} Our work uses the calibration dataset to calculate centroids ($\Bar{h}_+$, $\Bar{h}_-$) to decode the $h$.}
    \label{fig:1_Method_Abs}
\end{figure}

\textbf{I}n-\textbf{c}ontext \textbf{L}earning (ICL)~\cite{dong2022survey} is a few-shot learning paradigm without model parameter updates on \textbf{L}anguage \textbf{M}odels (LMs). In detail, as shown in Fig.~\ref{fig:1_Method_Abs}-(A, B), given a prompt consisting of demonstrations and a query, LMs conduct causal language modeling operation from the prompt to assign probabilities to the label token candidates designed by hand, and ICL chooses the one with the highest probability as the prediction. 

One well-known issue of ICL is that the predicted probabilities are biased (\textit{under-calibrated}), leading to a decrease in prediction performance~\cite{fei2023mitigating,han2022prototypical,zhao2021calibrate,zhou2023batch}. To address this issue, previous work \textit{calibrates} the predicted label token probabilities by performing affine transformations with estimated parameters to adjust these probabilities for more precise predictions.

These previous works and also the vanilla ICL are based on a potential assumption: the affine manifolds spanned by the decoding vectors in the LM head (\textit{un-embedding vectors}) of \textit{manually selected} label tokens are good subspaces of the hidden space to distinguish hidden vectors (i.e., the last hidden states to be fed to the LM head) onto various label appropriately, so that the label token probabilities decoded from these subspaces are accurate classification logits. However, although using the label un-embedding with task-related semantics (e.g.\ ``positive" and ``negative") seems intuitive, it should be noted that we have no reason to believe that these label un-embeddings have any explicit guarantee for decoding the last hidden states into accurate classification logits (token-based decision criterion), even if various delicate calibrations are used to move these boundaries \textit{inside the subspaces} (see \S\ref{sec:3.1}). Also, some works have pointed out that randomly changing label spaces doesn't critically influence ICL performance~\cite{min2022rethinking, wei2023symbol}, which means the selected label subspaces are trivial and arbitrary, making a suspicion of: using manually selected label un-embeddings to decode the last hidden states, i.e., \textbf{utilizing manually selected label probabilities as classification criteria may not be good ICL practices}. 

Previous work has shown that using the output probabilities of the \textit{full vocabulary} increases ICL performance~\cite{xu2022k, abbas2024enhancing}. This is a good start to avoid the manually selected classification criteria, but there is still doubt that output probability distributions are not informative enough for classification (see \S\ref{sec:3.2}). Therefore, we utilize the last hidden states instead, which are informative precursors of the token probabilities. 

Concretely, we propose \M, training centroid classifiers on the last hidden states of ICL prompts. As shown in Fig.~\ref{fig:3_Method}, \textbf{during the training}, we build standard ICL prompts similarly to Fig.~\ref{fig:1_Method_Abs}-A from a supervised calibration set and input them into the LM to get the last hidden states of the last tokens of ICL prompts. Then, we calculate the centroids of the last hidden states w.r.t.\ the queries' label to get a centroid for each label, as an anchor for inference. \textbf{During the inference}, we input the test prompt, find the nearest centroid to the last hidden states of the test prompt, and assign the corresponding label of the centroid as the prediction.

Empirically, \M~improves the ICL performance by approximately more than 20\% on 10 text classification datasets and 6 modern LMs (\S\ref{sec:4.1}), with an equal computational cost with previous calibration methods. To the best of the author's knowledge, \M~consistently outperforms the calibration baselines, achieving a strong state-of-the-art in ICL. Additional experiments indicate that \M~effectively alleviates the stress of prompt engineering, performing robust accuracy under different prompt designs.

Moreover, our subsequent analysis indicates that \M~does find better logits-mapping subspaces that effectively separate data points. In detail, we find that the distribution of classification logits calculated from \M~have less inter-class overlapping than from label probabilities, while such overlapping is proportional to the lower bound of the classification error. This suggests \M~finds subspaces with essentially better classification performance.

Furthermore, we investigate the principle of \M, that is, the reason why a simple centroid-based linear decision boundary can divide the ICL hidden state properly. We find that LMs provide linearly separable clusters in the hidden states w.r.t.\ query labels, while more demonstrations can promote such a process.

\vspace{0.8em}
\noindent\textbf{Our contributions can be summarized as:}
\vspace{-0.4\baselineskip}
\begin{itemize}
    \item We analyze the previous calibration practices on ICL, and find their consistent limitations: using predicted probabilities of manually selected label tokens for classification criteria, which is often under-guaranteed.
    \vspace{-0.4\baselineskip}
    \item We propose \M~to address the problem before, eliminating the unreliable decoding on the hand-selected label, and using a centroid classifier on the LM's last hidden states. Our experiments suggest that \M~reaches a strong state-of-the-art.
    \vspace{-0.4\baselineskip}
    \item Our further analysis indicates that \M~does find better classification criteria with less inter-class overlap, and LMs provide linearly separable intra-class clusters with the help of demonstrations, which supports \M{} to classify samples accurately.
\end{itemize}

\section{Background}

This section reviews previous work on ICL and denotes their mathematical descriptions as an introduction to the main motivation of this work.

\subsection{In-context Learning}
\label{sec:2.1}

\paragraph{Prompting.} Given a few-shot natural language classification dataset (\textit{demonstration set}) $\mathcal{D}=\left\{\left(x^{(i)}, y^{(i)}\right)\in\mathcal{X}\times\mathcal{Y}\right\}_{i=1}^n$, where $x^{(i)}$ and $y^{(i)}$ are the input sequence and label token of $i$-th data point, and $\mathcal{X}, \mathcal{Y}$ is the input and label space, respectively, we sample a set of $k$ samples $\mathcal{D}^{de}=\left\{\left(x^{(c_i)}, y^{(c_i)}\right)\right\}_{i=1}^k$ from $\mathcal{D}$ with an index set $\left\{c_i\right\}_{i=1}^k$ for a given query $x^q$ as the demonstrations. Then, we use a template $T$ to concatenate them in a natural language form into a prompt token sequence: $s = T\left(\mathcal{D}^{de}, x^q\right)$, as shown in Fig.~\ref{fig:1_Method_Abs}-A. 

\paragraph{Encoding.} A decoder-structured LM receives the prompt token sequence $s$ and encodes it into the \textit{last} (from the last Transformer layer) hidden state matrix as $H\in\mathbb{R}^{\vert s\vert\times d}$ with a length of token $\vert s\vert$ and embedding dimension of $d$. We denote the hidden state of the last token as $h = H_{\vert s\vert}\in\mathbb{R}^d$.

\paragraph{Constrained Decoding.} In a typical ICL setup, one chooses the un-embedding vectors of the label candidates in the output head\footnote{We omit the bias term in the output head (if any) for the sake of simplicity, which can be overridden by a fixed-to-one dimension, or covered by the calibration described below.} to decode $h$ as the prediction. That is, for each label $l$, the similarity $\alpha_l = \mathrm{sim}(h, E_l^U)$ (usually the dot-product similarity) between $h$ and the un-embedding vector $E_l^U$ is calculated as the output logits $\alpha_l$, as shown in Fig.~\ref{fig:1_Method_Abs}-B for a binary classification example. Then, the label with the highest logits is chosen as the prediction $\hat{y}$, that is: $\hat{y} = \mathop{\mathrm{argmax}}\limits_{l\in \mathcal{Y}}\mathrm{sim}(h, E_l^U)$. 

\begin{figure}[t]
    \centering
    \includegraphics[width=1.00\linewidth]{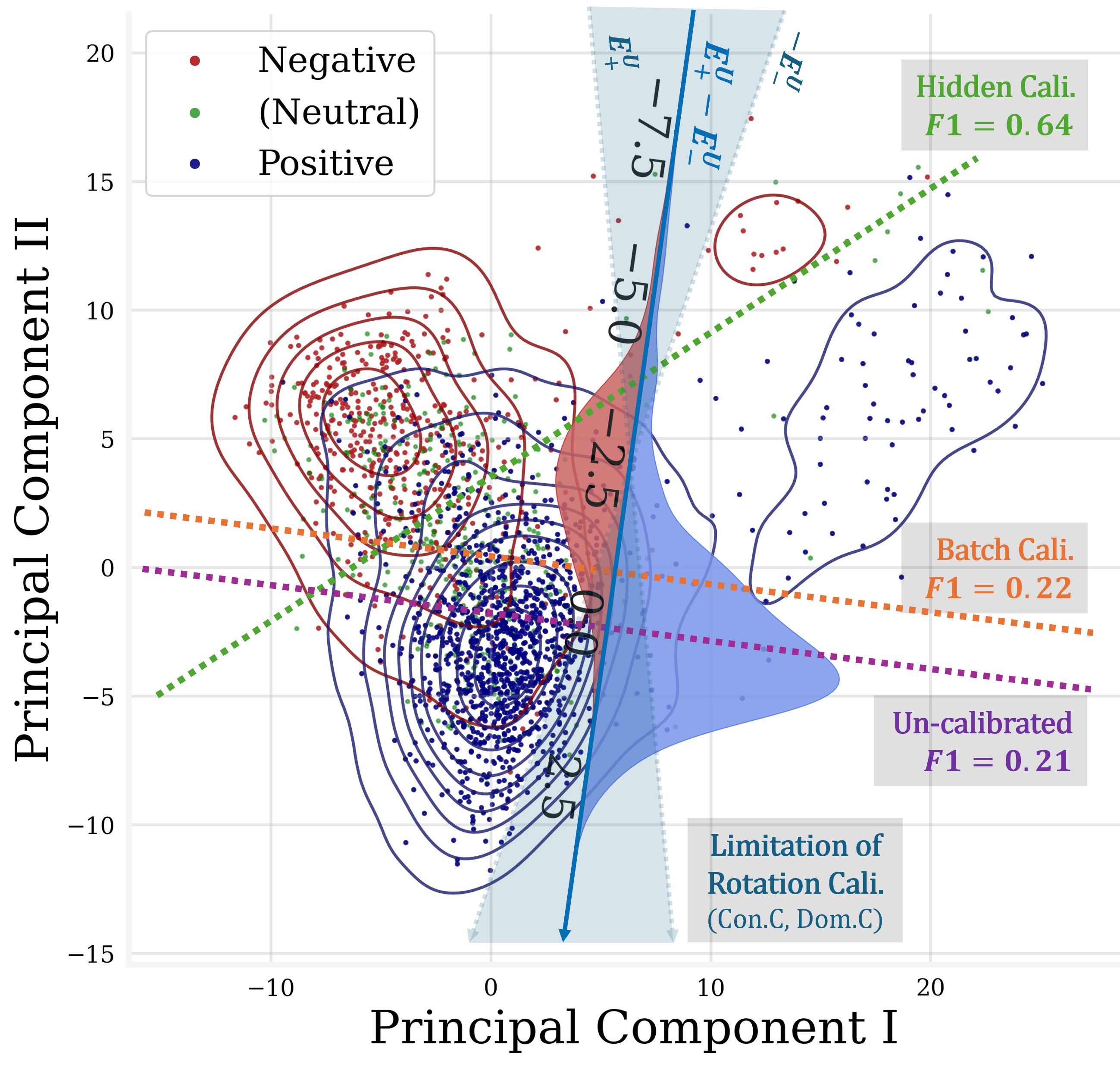}
    \vspace{-1.5\baselineskip}
    \caption{Token probability-based decision boundaries (\textcolor[HTML]{a02b93}{original} \& \textcolor[HTML]{e97132}{batch calibrated}) are suboptimal comparing to \textcolor[HTML]{4ea72e}{centroid-based boundary}. Points and contour lines are ICL's last hidden states and kernel densities mapped by Principal Component Analysis. \textcolor[HTML]{0070c0}{Oblique coordinate axis} is the direction of the un-embedding difference of label tokens $\left(E_+^U-E_-^U\right)$, where the kernel densities of mapped data points are plotted. The rotating calibration by $A\neq \mathbf{1}$ (e.g.\ Contextual Calibration, Domain Calibration) has a \colorbox[HTML]{d8e4ea}{limited feasible mapping direction\textsuperscript{\ref{footnote:3}}}.}
    \label{fig:2_Motivation}
\end{figure}

\vspace{-0.5\baselineskip}
\subsection{Token-probability Calibration for ICL}

However, \newcite{zhao2021calibrate} find that simply using the original logits for classification can not lead to a good ICL practice, since these logits have considerable prior bias and often tend towards specific labels even if the query is blank or meaningless~\cite{zhao2021calibrate, fei2023mitigating}. Some calibrations have been proposed to mitigate such bias in a linear form: first, the logits are transformed into probabilities as $p = \mathrm{softmax}\left(\left[\alpha_1, \alpha_2, \dots, \alpha_{\vert\mathcal{Y}\vert}\right]\right)$, then affine-transformed as calibrated classification criteria $p'=A\odot p+B$, where $A, B\in\mathbb{R}^{\vert\mathcal{Y}\vert}$ is the calibration terms estimated from $m$ training examples from a calibration set, and $\odot$ is the Hadamard multiplication. Various estimations for $A$ and $B$ are used: some practices use examples with pseudo queries terms~\cite{fei2023mitigating, zhao2021calibrate}, while other practices use Gaussian estimation on real prompts~\cite{han2022prototypical} or the mean value of $p$ during the inference~\cite{zhou2023batch}.

However, as to be discussed in~\S\ref{sec:3.1} current calibrations are affine transformations on label token probability, without modifying the $E^U_l$, causing only limited improvement to ICL performance.


\section{Methodology}

Based on the above background, in this section, we demonstrate the limitations of the above calibrations, and then propose \M~to address such limitations fundamentally.

\subsection{Token Probabilities Are Not Good Classification Criteria}
\label{sec:3.1}


To better understand the limitations of the label token probability-based ICL, we show a prototypical visualization of the hidden states of ICL prompts (aforementioned $h$). Specifically, we input 2,048 ICL prompts (with $k=8$) built from of \SER~\cite{pontikietal2014semeval} into \OPT~\cite{zhang2022opt} and plot the $h$ on a 2D-Principal Component plane in Fig.~\ref{fig:2_Motivation} (detailed in Appendix~\ref{Appendix:Visual}).

As a simple 2-way case, focusing on the data points labeled ``positive'' and ``negative'', we plot the \textcolor[HTML]{0070c0}{difference direction $\left(E_{+}^U-E_{-}^U\right)$} between the un-embedding vectors of these two label tokens\footnote{Notice that Principal Component Analysis is an orthogonal transformation, keeping the dot-product and normal line fixed (In fact, beyond orthogonal transformations, they are also centralized. Therefore, the \textcolor[HTML]{0070c0}{projection axis} does not necessarily pass through the coordinate origin). See Appendix~\ref{Appendix:Visual}.}. Then, the coordinates of the projected hidden states in this direction are the difference of predicted logits between these two labels, serving as a \textit{token-based classification criteria}, i.e., when the coordinate is positive, a ``positive'' label will be assigned, and vise versa. Therefore, in this visualized scenario, the orthogonal line at the zero point is the \textcolor[HTML]{a02b93}{original decision boundary}, points below this boundary are classified as ``positive'', and vise versa. The \textcolor[HTML]{e97132}{batch calibrated boundary}~\cite{zhou2023batch} is always parallel to the original one, and the other calibrations (Contextual Calibration~\cite{zhao2021calibrate}, Domain Calibration~\cite{fei2023mitigating}) produce \textcolor[HTML]{156082}{rotated mapping directions $\left(A_+E_+^U - A_-E_-^U\right)$}, by positive-definite term $A$ and thus rotated decision boundaries, with limited direction\footnote{In current practices, the $A$ are calculated from reciprocals of probabilities, which are positive-definite (Note that the calibration is trivial when $A$ is not positively definite: the label with negative $A$ components will never be assigned), and usually do not have significant relative values.\label{footnote:3}} between $E_+^U$ and $-E_-^U$.

Intuitively, as shown in Fig.~\ref{fig:2_Motivation}, the token-based decision boundaries cannot effectively classify these data points, which is due to the inherent direction of the token un-embedding vectors, regardless of limited affine transformation by calibration. A straightforward better linear boundary is the \textcolor[HTML]{4ea72e}{equidistant points} between both classes' centroids, we try to find it as follows.

\subsection{Hidden Calibration}
\label{sec:3.2}

\begin{figure}[t]
    \centering
    \includegraphics[width=1\linewidth]{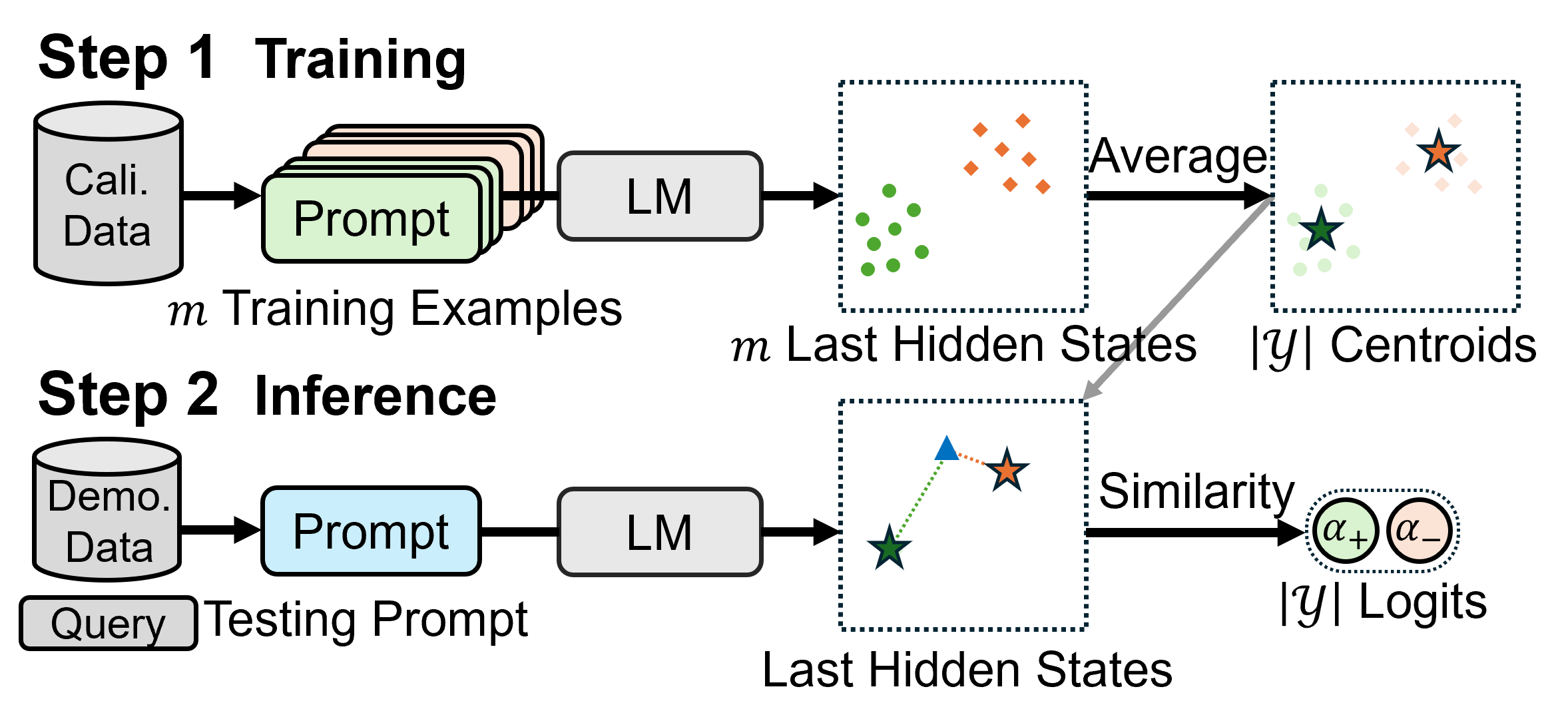}
    \vspace{-1.8\baselineskip}
    \caption{The diagram of \M. \textbf{Step 1:} Calculating the hidden state centroid of each label. \textbf{Step 2:} Find the label of the nearest centroid of the text sample to be the prediction.}
    \vspace{-0.9\baselineskip}
    \label{fig:3_Method}
\end{figure}

Motivated by the visualization, we propose \M, using the centroid similarity as the classification logits. In practice, we use a 2-step paradigm as shown in Fig. \ref{fig:3_Method}: first, as training, we calculate the centroid of the last hidden states of data points within each class on some (of amount $m$) prompt-label training examples. Then, in the inference, we select the closest centroid of the test prompt's hidden state as the prediction. 

In detail, \textbf{(1) Training:} Given a \textit{calibration set} with $m$ supervised prompt-label pair $\left\{\left(s^{(i)}, y^{(i)}\right)\right\}_{i=1}^m$, where the $s^{(i)}$s (\textit{training examples}) are standard ICL prompts with $k$ demonstrations, and $y^{(i)}$s are the ground-truth labels of corresponding $s^{(i)}$s' query, we input each training example $s^{(i)}$ to LM, and extract the last hidden state $h^{(i)}$. Repeating on the whole training example set, we can get a supervised hidden state set $\mathcal{H} = \left\{\left(h^{(i)}, y^{(i)}\right)\right\}_{i=1}^m$. Then, we calculate the centroids of label $l$ as: $\Bar{h}_{l} = \mathbb{E}_{\left(h^{(i)},y^{(i)}\right)\in\mathcal{H}, y^{(i)}=l}\left[h^{(i)}\right]$.

Then, we utilize the calculated centroids in \textbf{(2) Inference:} Given a test ICL prompt, we input it into the LM and get the last hidden state $h$, then calculate the similarity between $h$ and every centroid $\Bar{h}_{l}$ as the centroid-based logits $\alpha_l$. In practice, the additive inverse of Euclidean distance is used as the similarity (that is, $\alpha_l = -\left\Vert h-\Bar{h}_l\right\Vert_2^{\frac{1}{2}}$), while Appendix~\ref{Appendix:Similarity} shows that \M~acts equally on cosine similarity. We assign the label with the highest logits as the prediction.

\textbf{``Why hidden states?''} Notice that another intuitive solution to the problem in \S\ref{sec:3.1} is utilizing the logits or probabilities of the \textit{whole vocabulary}, as shown in previous works~\cite{xu2022k,abbas2024enhancing}. However, since the input, hidden states, and logits form a Markov chain, no input-relevant information gain is propagated to the full-vocabulary logits. Moreover, the dimensionality of the full-vocabulary logits is typically significantly larger than the hidden states, therefore we choose the hidden states, a dense and informative precursor of token probabilities, as the classification feature.

\textbf{``Why centroid classifier?''} Moreover, more complex classifiers, such as a KNN classifier, or a multi-layer perceptron, can be used on the last hidden states instead of a centroid classifier. However, we choose the centroid classifier as the simplest implementation to avoid attribution confusion, that is, if even a rudimentary classifier on hidden states still outperforms, it is powerful enough to demonstrate that our hypothesis is robust. Also, a centroid classifier has a minimal computation cost to fit the scenario of low-resource (\S\ref{sec:4.3}).

\section{Experiments \& Main Results}
\label{sec:4}

In this section, we empirically prove the effectiveness of \M~by classification performance on 6 models and 10 datasets. \M~outperforms all the baselines and reaches a strong state-of-the-art of ICL with high efficiency in calculation, data, and prompt engineering.

\begin{figure*}[t]
    \centering
    \includegraphics[width=0.9\linewidth]{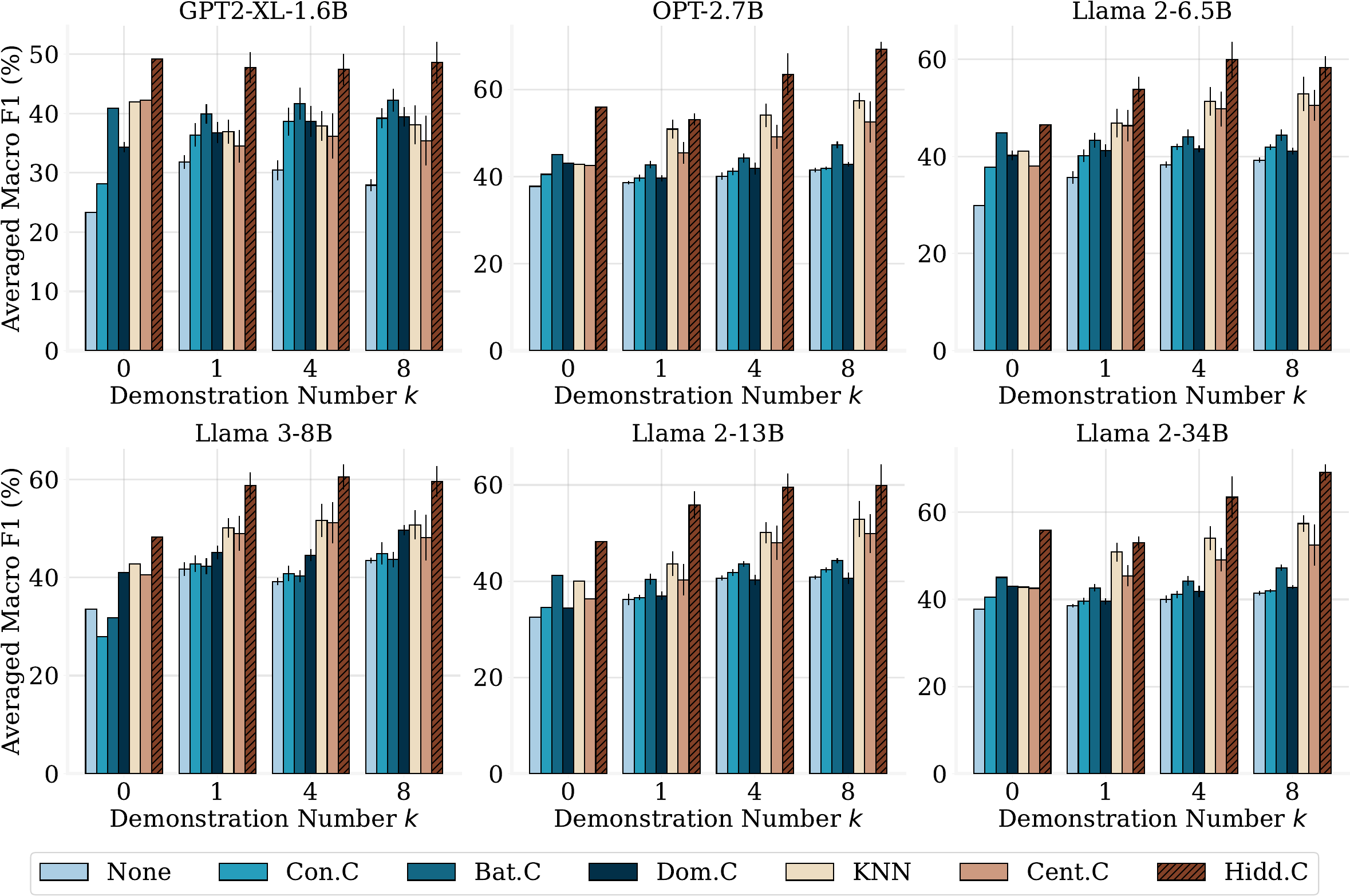}
    \vspace{-0.5\baselineskip}
    \caption{The classification performance (Macro F1(\%)) of 6 models averaged on 10 datasets. \M~(\textcolor[HTML]{834127}{Hidd.C}) is a new state-of-the-art of ICL, where demonstrations consistently improve the performance.}
    \label{fig:4_Mainres}
\end{figure*}

\subsection{Experimental Settings}
\label{sec:4.1}

\paragraph{Models.} We use 6 models: \OPT~\cite{zhang2022opt}, Llama 2~\cite{touvron2023llama} (6.9B, 13B, 34B), Llama 3~\cite{llama3modelcard} (8B) and \GPT~\cite{radford2019language} (1.6B). Models larger than 10B are quantized.

\paragraph{Baselines.} We use 6 baselines from the previous works, with 4 label token-based methods: \textbf{Vanilla ICL} (None)~\cite{radford2019language}, \textbf{Contextual Calibration} (Con.C)~\cite{zhao2021calibrate}, \textbf{Batch Calibration} (Bat.C)~\cite{zhou2023batch}, and \textbf{Domain Calibration} (Dom.C)~\cite{fei2023mitigating}; 2 whole vocabulary probabilities-based methods KNN~\cite{xu2022k} and \textbf{Centroid Calibration} (Cent.C), which we propose as a fair comparison with the same processing on the whole output vocabulary probability vectors instead of the hidden states. Details can be found in Appendix~\ref{Appendix:Baselines}. 

\paragraph{Datasets.} We use 10 commonly used classification datasets with some of the overlength data points excluded. See Appendix~\ref{Appendix:Dataset} for details.

\paragraph{Other details.} All the model checkpoints and datasets are loaded from \verb|HuggingFace|. Macro F1 is used as the classification metric. We use a simple template to generate the prompt, see Appendix~\ref{Appendix:Prompts}. We set $m=16|\mathcal{Y}|$ training examples (16 examples per class), and for fairness, every baseline is given equal training examples for calibration. All the experiments are repeated 5 times.

\subsection{Main Results: \M~is A New State-of-the-art of ICL}
\label{sec:4.2}

The tested classification performance of \M~and baselines is shown in Fig.~\ref{fig:4_Mainres}, where \M~(\textcolor[HTML]{834127}{Hidd.C}) consistently outperforms all the label token-based or vocabulary-based baselines. Comparing to the vanilla ICL (\textcolor[HTML]{abcee4}{None}), \M~produces an improvement up to around 100\%. In general, compared to the strongest baseline, \M~improves the performance by approximately 20\%. Detailed numeric and Accuracy results are in Appendix~\ref{Appendix:Detailed_Results_mainres}.

Especially, compared to the \textcolor[HTML]{cd9a80}{Cent.C} baseline proposed by us for a controlled trial, which conducts the same calculation but uses the whole output token probabilities instead of the hidden states, \M~outperforms, which confirms our idea that token probability distribution is a less informative classification feature mentioned in \S\ref{sec:3.2}.


\subsection{Efficiency: Low Complexity towards Time, Space, Data, and Prompting}
\label{sec:4.3}

\paragraph{Time and Space Complexity.} Intuitively, \M~has little additional computational cost compared to the calibration baselines, since they require almost equivalent feedforward calculations, making it competitively efficient as listed in Table~\ref{table:1_cost}. Here, we are most concerned about the inference time cost, and~\M~is the fastest among all the non-label-based methods since the product $\vert\mathcal{Y}\vert d$ is usually not very large.

\begin{figure*}[t]
    \newlength{\figurelength}
    \setlength{\figurelength}{0.90\linewidth}
    \centering
    \includegraphics[width=0.3285\figurelength]{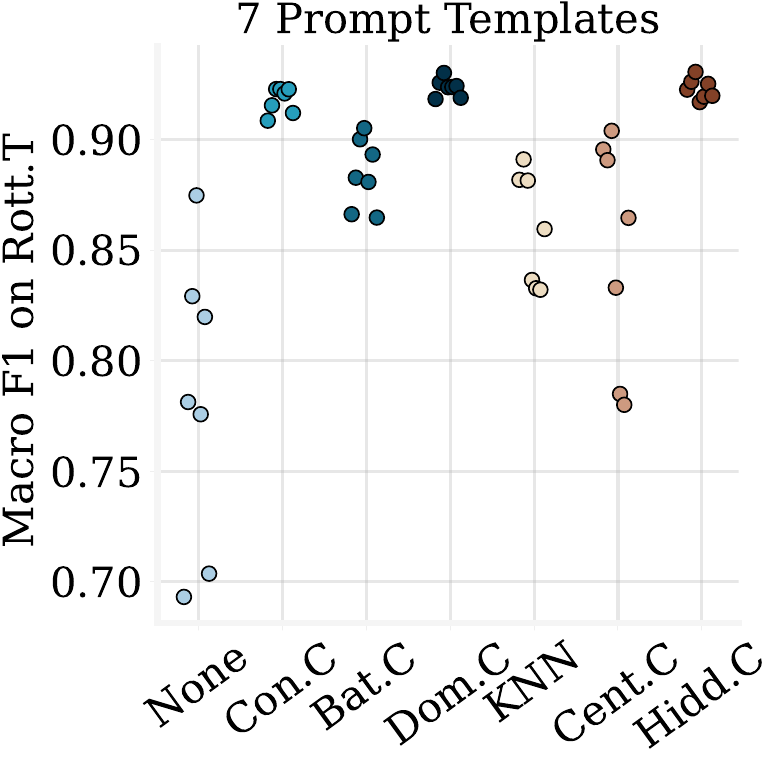} \hspace{3mm}
    \includegraphics[width=0.315\figurelength]{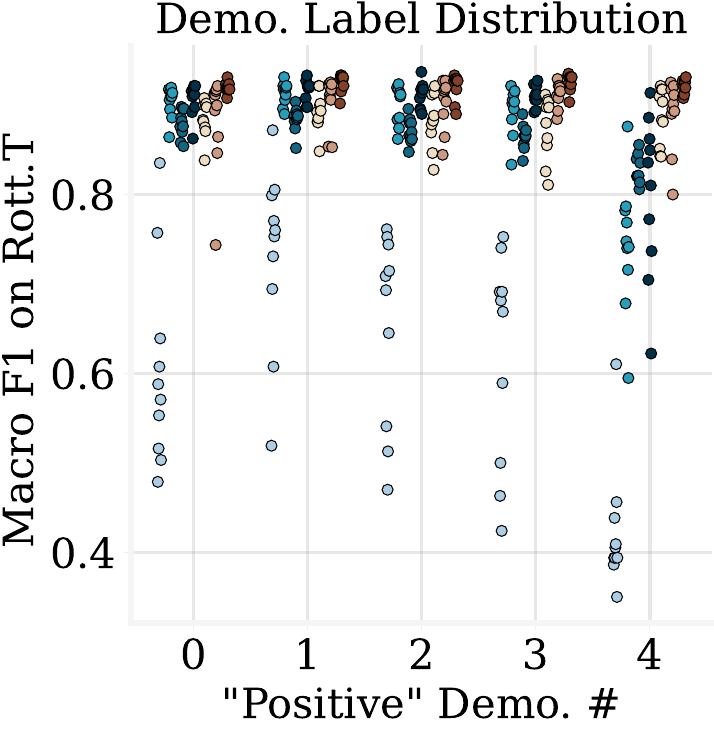} \hspace{3mm}
    \includegraphics[width=0.315\figurelength]{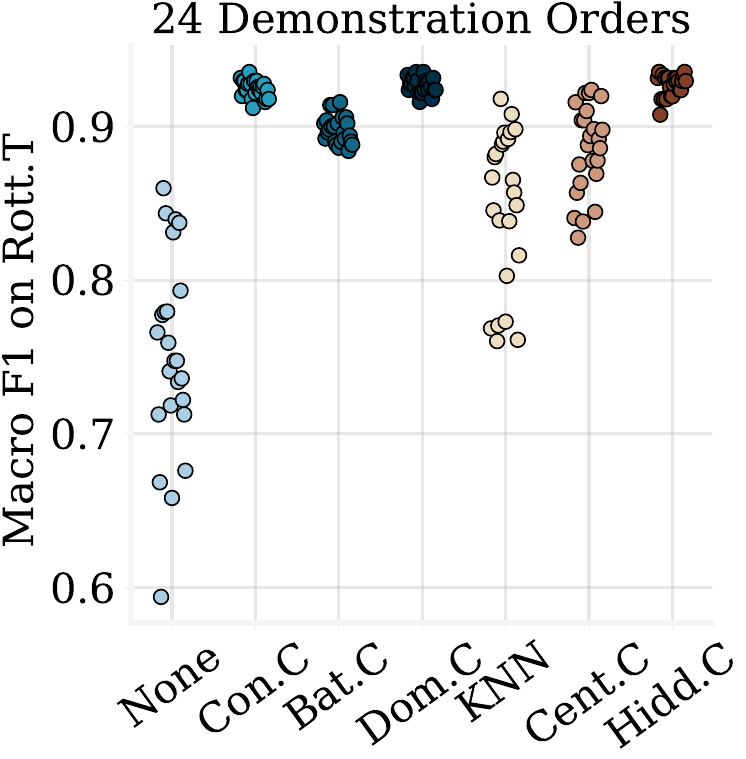}
    \vspace{-0.3\baselineskip}
    \caption{Sensitivities on (\textbf{left}) prompt template, (\textbf{middle}) demonstration label distribution, and (\textbf{right}) demonstration order on \LLAMA~and \texttt{Rotten\_Tomatoes}. Legend is consistent with Fig.~\ref{fig:4_Mainres}, omitted.}
    \label{fig:sense}
    \vspace{-0.8\baselineskip}
\end{figure*}

\begin{figure}
    \centering
    \includegraphics[width=0.89\linewidth]{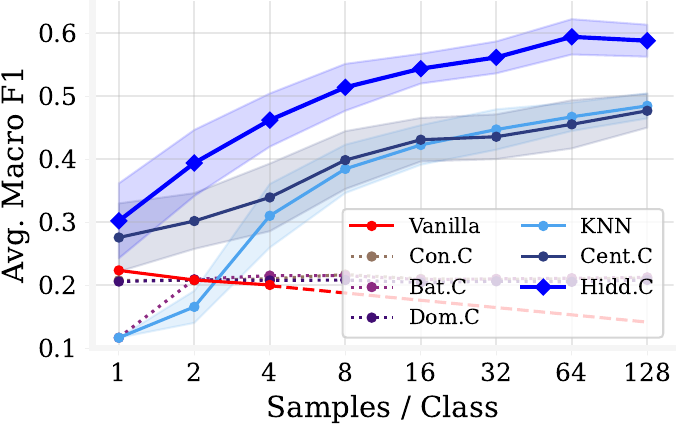}
    \vspace{-0.5\baselineskip}
    \caption{Classification performance against the number of training examples ($m$) of calibrations. As a comparison, we plotted the results of vanilla ICL using equivalent demonstrations, due to the quadratic overhead against the context length, we can test up to $k=4\vert\mathcal{Y}\vert$.}
    \label{fig:11_Data_eff}
    \vspace{-0.7\baselineskip}
\end{figure}

\paragraph{Training Data Complexity.} \M~requires additional annotated data compared to label token-based calibration methods of the same scale in an acceptable range. In detail, in Con.C and Dom.C of $k$ demonstrations, $k$ supervised data is needed with a synthetic query to build a training example, while \M~needs a real query \textbf{for each label}, requiring one more supervised data. However, for classification tasks, preparing an example for each label can be easily done whether in an industry or laboratory scenario, furthermore, these data and trained centroids can be reused (Appendix~\ref{sec.appendix.transf}) to further reduce the requirement of annotated data.

\paragraph{Training Sample Efficiency.} Regarding the efficiency of training examples, we repeat the experiments with various $m$ on \OPT~(see Appendix~\ref{Appendix:ExpDetails_5.3} for details), from 1 to 128 calibration examples \textbf{per class}. Also, for the vanilla ICL, we give equivalent demonstrations for a fair comparison. The results are shown in Fig.~\ref{fig:11_Data_eff}, which indicate that \M~stably benefits from the size of the calibration set, while even one sample per class can still make it outperform. Meanwhile, vanilla ICL and label token-based methods can not benefit from more available data, making \M~a better practice no matter how much supervised data can be accessed: data can be used to estimate the centroid to improve the classification in a linear cost, rather than increase the demonstrations in a quadratic cost with less benefit.

\begin{table}[t]
\centering
\caption{The \textbf{additional} (compare to vanilla ICL) time and space on calibration and inference cost of various methods. \M~has a similar cost upper bound to other calibrations. $\vert\mathbb{V}\vert$ is the vocabulary size.}
\vspace{-0.6\baselineskip}
\label{table:1_cost}
\resizebox{0.9\columnwidth}{!}{
\begin{tabular}{@{}cccc@{}}
\toprule
\multirow{2}{*}{\textbf{Method}} & \multicolumn{2}{c}{\textbf{Training Cost}}   & \textbf{Inference Cost} \\ \cmidrule(l){2-3} \cmidrule(l){4-4} 
                        & Add. Space                 & Add. Time & Add. Time      \\ \midrule
None                    & $0$                        & $0$       & $0$            \\
Con.C                   & $O(\vert\mathcal{Y}\vert)$ & $O(m)$      & $O(\vert\mathcal{Y}\vert)$          \\
Bat.C                   & $0$                          & $0$        & $O(m\vert\mathcal{Y}\vert)$          \\
Dom.C                   & $O(\vert\mathcal{Y}\vert)$    & $O(m)$     & $O(\vert\mathcal{Y}\vert)$            \\ \midrule
KNN                     & $O(m\vert\mathbb{V}\vert)$  & $O(m)$      & $O(m\vert\mathbb{V}\vert)$         \\
Cent.C                  & $O(\vert\mathcal{Y}\vert\vert\mathbb{V}\vert)$ & $O(m)$ & $O(\vert\mathcal{Y}\vert\vert\mathbb{V}\vert)$ \\
\cellcolor[HTML]{EFEFEF}\textbf{Hidd.C}         & \cellcolor[HTML]{EFEFEF}$O(\vert\mathcal{Y}\vert d)$ & \cellcolor[HTML]{EFEFEF}$O(m)$ & \cellcolor[HTML]{EFEFEF}$O(\vert\mathcal{Y}\vert d)$ \\ \bottomrule
\end{tabular}
}
\vspace{-0.5\baselineskip}
\end{table}

\paragraph{Prompting Complexity.} We find \M~reduces the pressure on prompt engineering for \M~and baselines on \LLAMA~and \texttt{Rotten\_Tomatoes} ($k=4$) in three aspects: \textbf{(1) Prompt template}~\cite{voronov2024mind}. We select 7 different prompt templates (shown in Appendix~\ref{Appendix:Prompts}) and test ICL performance on them, shown in Fig.~\ref{fig:sense} (left). \textbf{(2) Label distribution in demonstrations.} We construct prompts with various numbers of ``positive'' demonstrations presented, and test ICL performance shown in Fig.~\ref{fig:sense} (middle), \textbf{(3) Demonstration order}~\cite{lu2022fantastically}. We enumerate the full arrangements from a fixed demonstration set of $k=4$, and test the ICL performance using each demonstration arrangement, shown in Fig.~\ref{fig:sense} (right). All the results show that: compared to baselines, \M~keeps narrow and high-performance distribution against all the three variables, i.e., \M~stably works for various contexts, providing higher efficiency on prompt designing.

\begin{figure*}
    \centering
    \includegraphics[width=\linewidth]{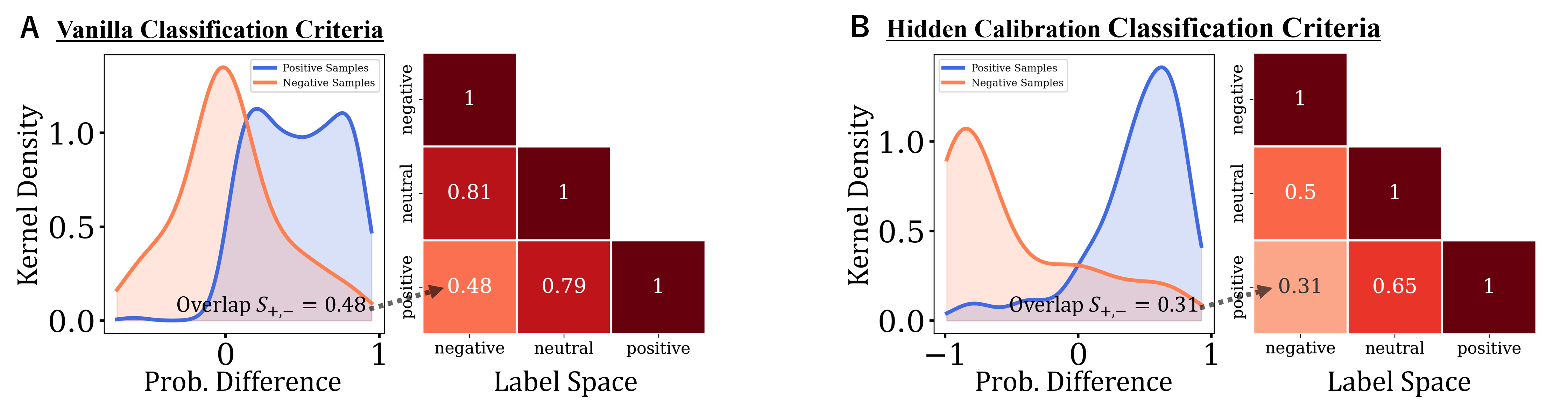}
    \vspace{-2\baselineskip}
    \caption{Diagrammatic sketch of the overlap calculation with \GPT~on \SER, $k=4$. \textbf{Curves:} The kernel density of probability difference of \textcolor[HTML]{456be1}{$l_1=$``positive''} and \textcolor[HTML]{fd7f53}{$l_2=$``negative''}. \textbf{Heatmaps:} The overlap of 2-combinations (we plot the combination with the same label with overlap 1, but omit them in averaging).}
    \label{fig:5_Ana1_intro}
    \vspace{-0.6\baselineskip}
\end{figure*}

\begin{figure*}[t]
    \centering
    \includegraphics[width=1\linewidth]{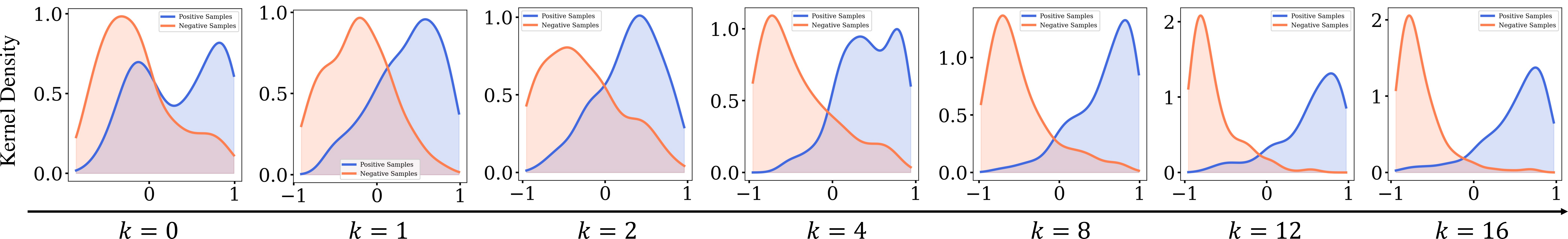}
    \vspace{-1.2\baselineskip}
    \caption{ICL hidden states clustering dynamics visualized on \OPT~and \SEL~\cite{pontikietal2014semeval}. The densities of data points appear in clusters responding to their query labels originally when no demonstrations are given, and gradually converge to the centroid w.r.t.\ the demonstrations number ($k$).}
    \label{fig:7_dynam_intro}
    \vspace{-0.6\baselineskip}
\end{figure*}

\section{Analysis}

This section attempts to enhance our understanding of \M~through comprehensive observations: \textbf{(1)} Similar to Fig.~\ref{fig:2_Motivation}, we measure the inter-class overlapping area on data points projected into classification criteria, to find whether \M~maps data into logits with lower inter-class overlap, i.e., better separability. \textbf{(2)} We further investigate why simple linear boundaries can effectively classify ICL hidden states, as observed in \M. We find that LMs provide a primary linear clustering in hidden states responding to query classes, and such clustering is enhanced by more demonstrations.

\subsection{Effectiveness: Hidden Calibration Finds Criteria with Lower Overlap}\label{sec:5.1}

\begin{figure*}[t]
    \begin{minipage}[t]{0.30\linewidth}
        \flushright
        \includegraphics[width=1\linewidth]{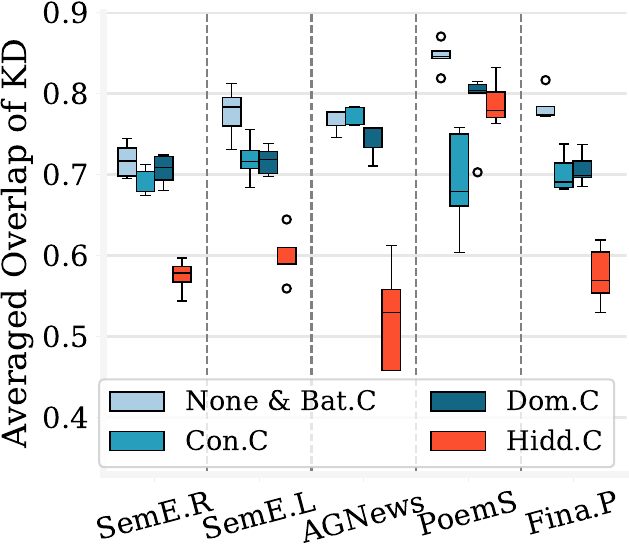}
        \vspace{-1.5\baselineskip}
        \caption{The Averaged Overlap of 4 inference methods on \GPT~and 5 datasets.}
        \label{fig:6_Ana1_OPT_res}
    \end{minipage} \hfill
    \begin{minipage}[t]{0.31\linewidth}
        \includegraphics[width=1\linewidth]{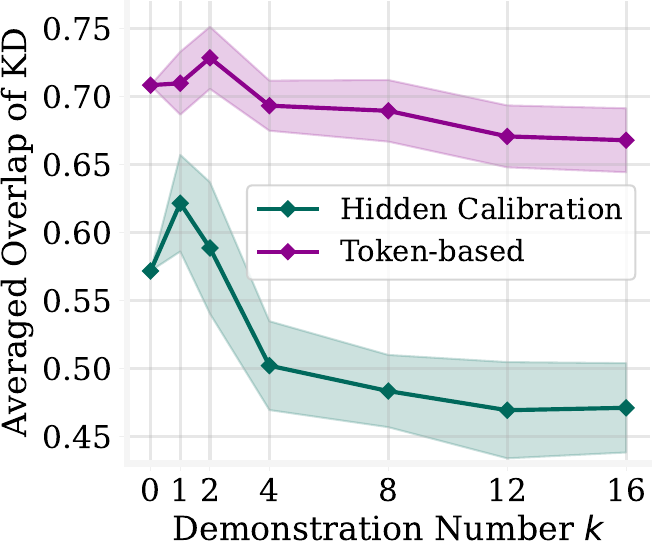}
        \vspace{-1.5\baselineskip}
        \caption{The Averaged Overlap on \OPT~and 5 datasets against the demonstrations number.}
        \label{fig:8_Dynam_Overlap}
    \end{minipage} \hfill
    \begin{minipage}[t]{0.33\linewidth}
        \includegraphics[width=1\linewidth]{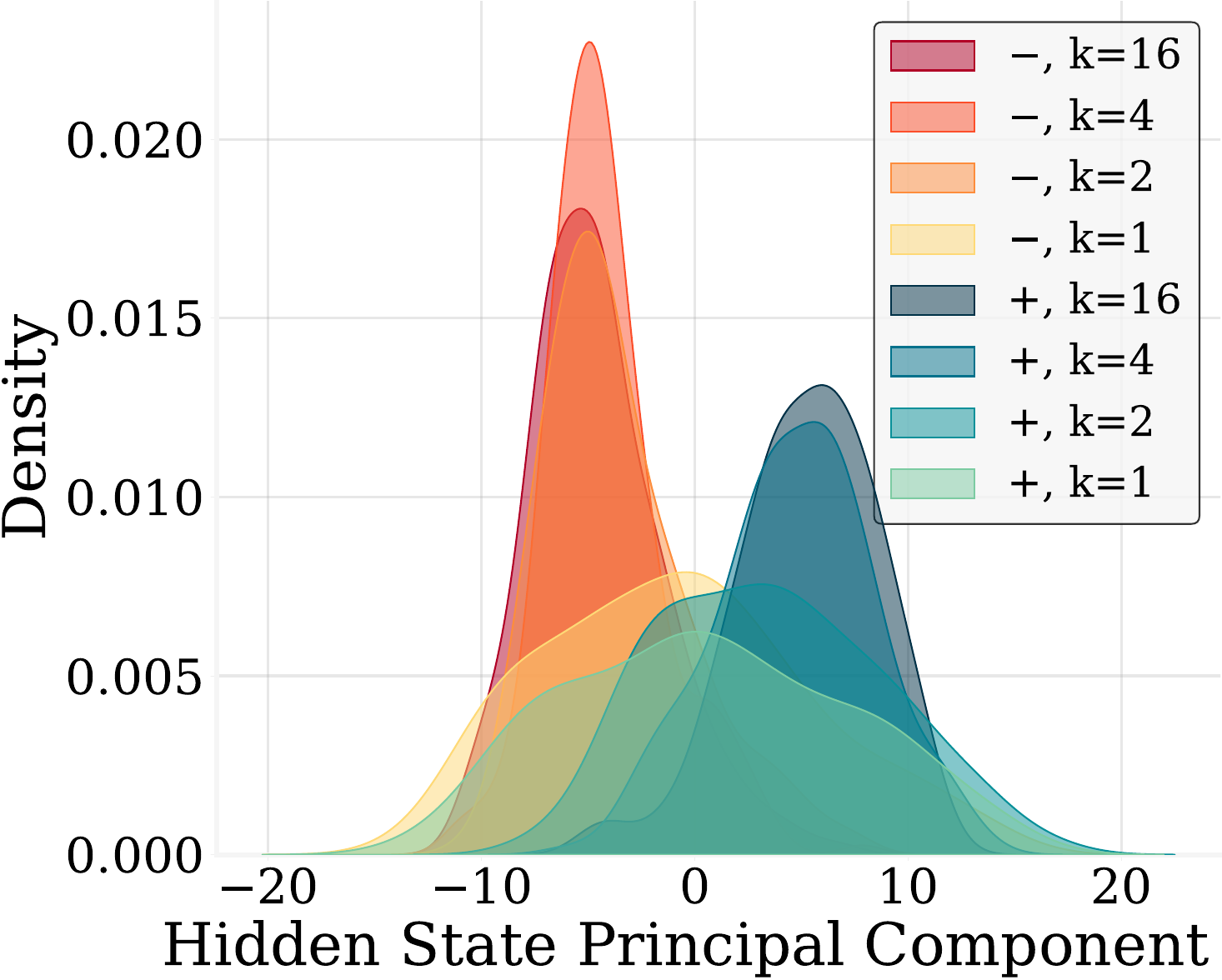}
        \vspace{-1.5\baselineskip}
        \caption{Hidden state clustering w.r.t.\ $k$ of Fig.~\ref{fig:7_dynam_intro} visualized on the direction of principal component II.}
        \label{fig:9_Dynam_PCAII}
    \end{minipage} 
    \vspace{-0.6\baselineskip}
\end{figure*}


In Fig.~\ref{fig:2_Motivation}, we projected the data points into the difference of the label logits (\textit{vanilla classification criteria}) on the \textcolor[HTML]{0070c0}{oblique coordinate axis}, then a significant \textit{overlap} between the projected data point cloud in two classes can be observed, making it difficult to find suitable classification boundaries vertical to the projection direction. Therefore, such overlap can be used to evaluate classification criteria, so, in this section, we quantify the intuitive observation as the area of overlap serving as a metric for classification criteria. 

In detail, we first decompose the multi-way classification dataset into all possible binary classification combinations w.r.t.\ the ground-truth labels (for example, in a binary combination with labels ``positive'' and ``negative'', only ``positive'' and ``negative''-labeled data is obtained). Then, for each combination, we build standard ICL prompts with queries labeled with a specific one of the selected binary combination. Input these prompts into the LM and map the last hidden state of the prompts onto the normal vector of the decision boundary formed by the calibration method, we get the mapped coordinate, as what is shown in the \textcolor[HTML]{0070c0}{oblique coordinate axis} of Fig.~\ref{fig:2_Motivation}. To get a continuous distribution of the distance, we run kernel density estimations on the calculated coordinate, then repeat this processing on the other label in the binary combination, and get two density estimations for both labels in a binary combination, as shown in Fig.~\ref{fig:5_Ana1_intro} (curves). Then, we calculate the overlap area of these two kernel density curves. We repeat such processing for each binary combination as shown in Fig.~\ref{fig:5_Ana1_intro} (heat maps), and the final Averaged Overlap is the macro average of overlap area among all possible binary combinations (operation details are in Appendix~\ref{Appendix:ExpDetails_5.1}).

The overlap area of the two distribution curves is double to the \textit{lower bound} of the classifier's error rate among these two labels (Appendix~\ref{Appendix:Proof}), so Averaged Overlap is an intuitive metric of the classification criteria: the larger the overlap, the more difficult it is for the classifier, even (further) calibrated or ideal, to classify data points correctly, resulting in a potential decrease in accuracy.

We measure the Averaged Overlap of 4 un-quantized models on 5 datasets (see Appendix~\ref{Appendix:OverlapExpDetails} for experimental details). The result on \GPT~is shown in Fig. \ref{fig:6_Ana1_OPT_res} (see Appendix~\ref{Appendix:ResDetails_5.1} for other models), where the Averaged Overlaps from token-based methods are consistently high, causing that better classification performance cannot be achieved on such methods, which confirms our hypothesis in~\S\ref{sec:3.1}. Meanwhile, the overlaps from \M~is much less than from token-based methods, meaning that \M~produces better classification criteria with better \textit{possible} classification performance than the token-based methods, even if delicate calibrations transfer or rotate these classification boundaries.

\subsection{Principle: The Inner Linear-separability}
\label{sec:5.2}


In the practice of \M, simple linear boundaries are used to classify ICL examples, raising curious on the linear separability of hidden states. In this section, We find that LMs primarily produce linearly separable hidden state clusters corresponding to the ground-truth label, and the demonstrations facilitate this process.

As an intuitive visualization, we plot curves the same as the Fig.~\ref{fig:5_Ana1_intro} but with various numbers of demonstrations $k$ to visualize the \textit{clustering dynamics} of hidden states in Fig.~\ref{fig:7_dynam_intro}, where we find that: \textbf{(1)} the data points have a little linear separability when $k=0$, and \textbf{(2)} such linear separability is being enhanced among the increment of $k$, performing increasing intra-class converging tendency.

We further characterize this process. First, we calculate the Averaged Overlap similar to~\S\ref{sec:5.1} against $k$ in Fig.~\ref{fig:8_Dynam_Overlap}. We find that the token-based overlaps remain high and stable w.r.t.\ $k$, which indicates that the token-based methods can not benefit much from the demonstrations. However, the overlaps from \M~significantly decrease with the increase of $k$, indicating that \M~benefits from the demonstrations as expected, aligning with our observations in~\S\ref{sec:4.2}.

More generally, we visualize the distribution of the last hidden states from similar inputs of Fig.~\ref{fig:7_dynam_intro} on the \textbf{second} principal components of hidden states to get an essential observation in Fig.~\ref{fig:9_Dynam_PCAII}, where as $k$ increases, the hidden state shows more clear intra-class clustering, enabling separability through a linear boundary.

\begin{figure*}[t]
        \includegraphics[width=0.2376\linewidth]{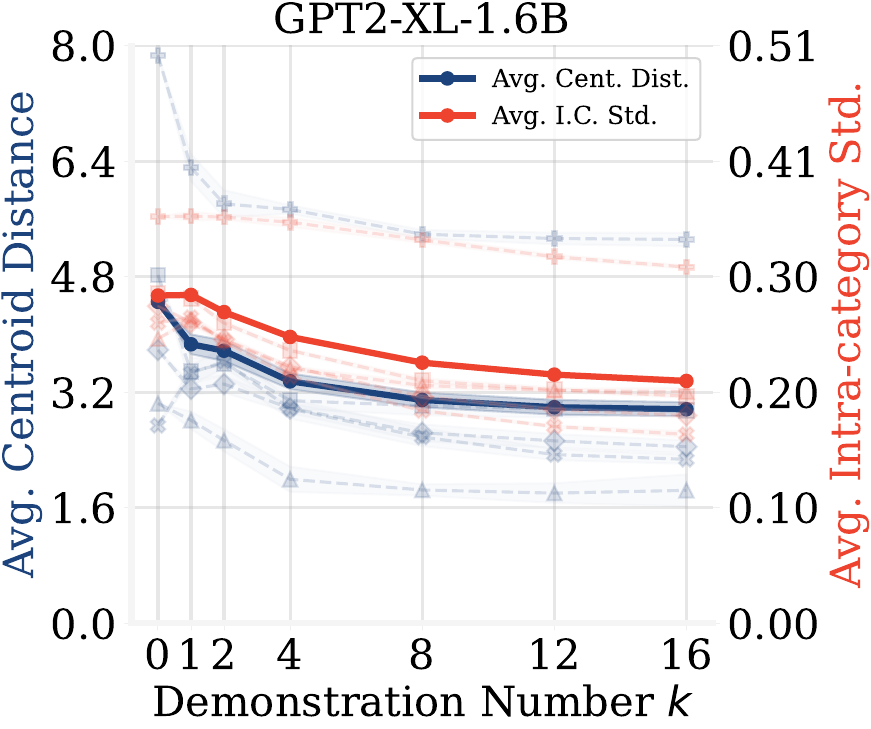}
        \includegraphics[width=0.244\linewidth]{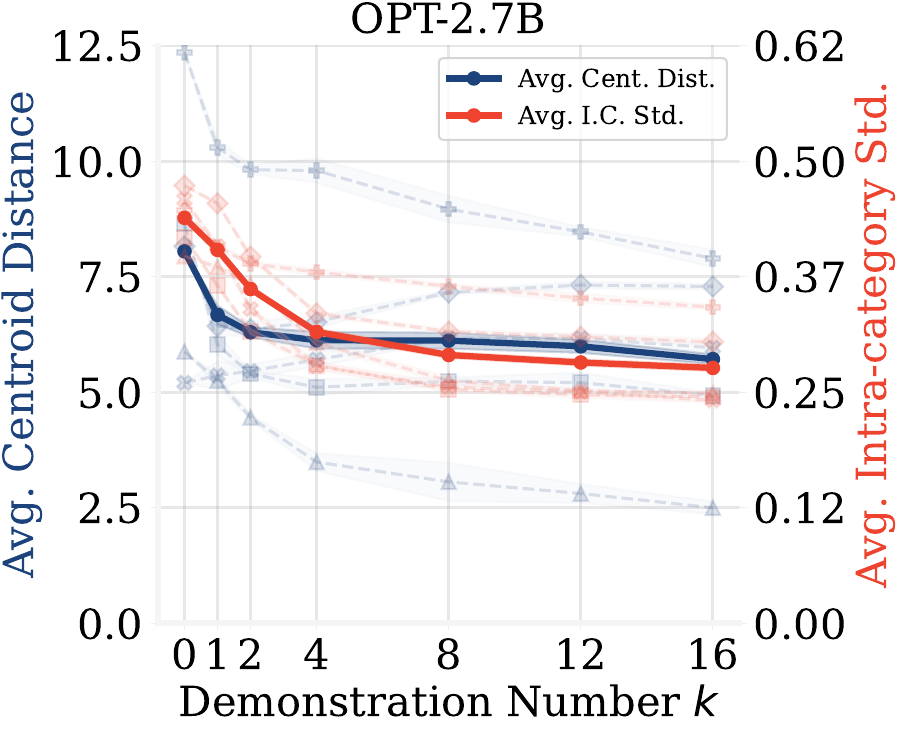}
        \includegraphics[width=0.244\linewidth]{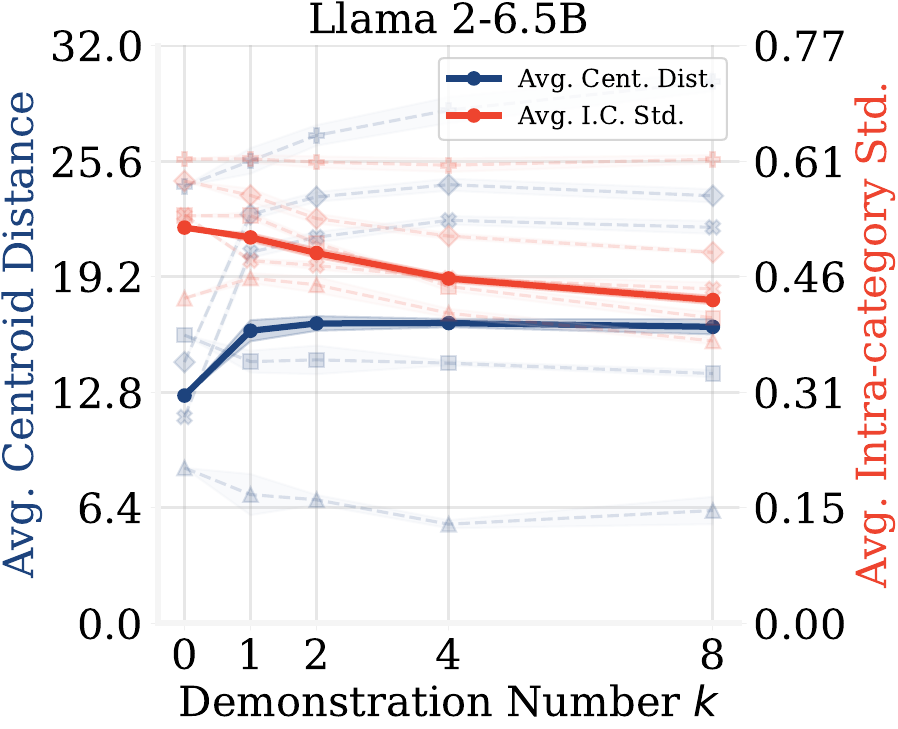}
        \includegraphics[width=0.234\linewidth]{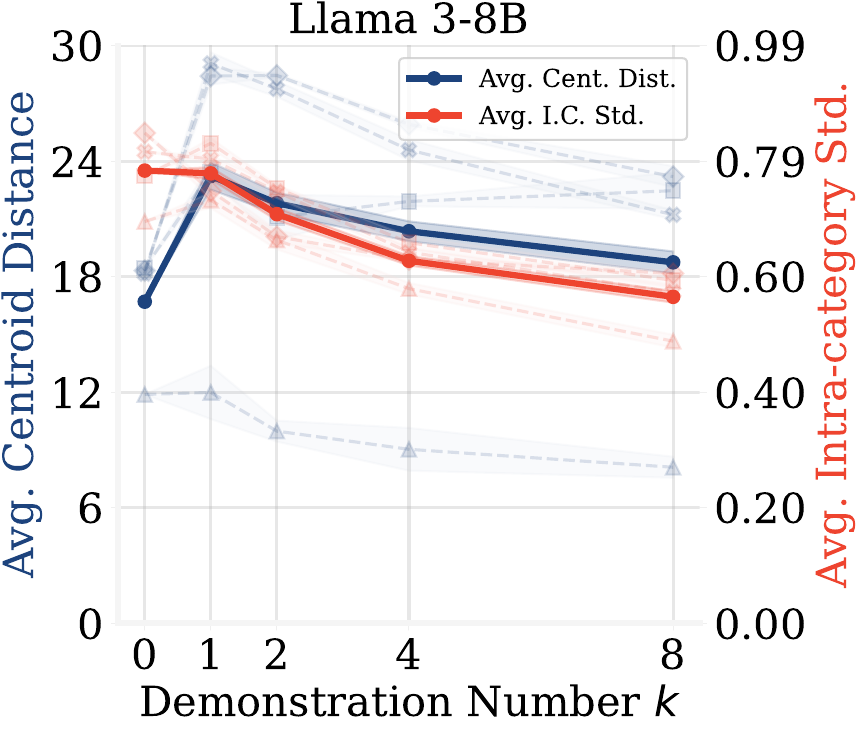}
        \caption{The \textcolor[HTML]{ee442f}{averaged intra-class standard error} of data points and the \textcolor[HTML]{1d437d}{inter-class averaged centroid distance} against $k$. \textbf{Solid curves}: means on 5 datasets; \textbf{Dashed and pale curves}: Individual results for each dataset.}
        \label{fig:10_Dynam_COV} 
\end{figure*}

More directly, on the last hidden states, we measure the intra-class standard error and the inter-class averaged centroid distance against $k$ (see Appendix~\ref{Appendix:ExpDetails_5.2} for details), both are a first-order moment for a joint measurement of intra-class clustering and inter-class clustering. The results are shown in Fig.~\ref{fig:10_Dynam_COV}, where the two curves are both diminishing, showing an obvious \textcolor[HTML]{ee442f}{intra-} and \textcolor[HTML]{1d437d}{inter-class} clustering trend w.r.t.\ $k$. However, the \textcolor[HTML]{1d437d}{inter-class} clustering has weaker and less persistent decreasing trends, presenting only in early demonstrations, or even ascending, which indicates that demonstration enhances \textcolor[HTML]{ee442f}{intra-class clustering} stronger than the \textcolor[HTML]{1d437d}{inter-class clustering}, which is beneficial to linear classification. Moreover, a model with more parameters shows a stronger difference between these clustering.

\section{Related Works}
\label{Appendix:RW}

Given the topic of enhancing in-context learning, we classify the literature into 3 categories. 

\paragraph{Model parameter update-based method:} Although it is pointed out that the ICL objective is implicitly included in pre-training data~\cite{han2023understanding}, explicitly fixing the gap between the ICL objective and causal language modeling objective can still be beneficial. Such methods are usually based on supervised fine-tuning~\cite{min2022metaicl, gu2023pre, wei2021finetuned, wei2023symbol, iyer2022opt, wang2022super}, and also self-supervised training~\cite{chen2022improving} and non-gradient method~\cite{zhao2024noisyicl}. Such methods usually require huge amounts of computation and data overhead to update billions of LM parameters. 

\vspace{0.2em}

\noindent In contrast, lightweight solutions focus on \textbf{classification criteria-based method (calibration)}. Such methods focus on re-calculating output label probabilities, keeping the main feed-forward calculation processes and model parameters un-modified. The original motivation for these works is to eliminate prior bias and unfaithful confidence in ICL, by calibrating the output label probabilities~\cite{holtzman2021surface, shi2022nearest, fei2023mitigating, zhao2021calibrate, han2022prototypical, zhou2023batch, jiang2023generative}. While, as described in the main text, some practices without the usage of label-specific probabilities have also been proposed~\cite{xu2022k, abbas2024enhancing, min2022noisy}.

\vspace{0.2em}

\noindent Also, a careful \textbf{design of input prompts} can help improve the ICL performance. \textbf{(1) Demonstration selection.}~\citet{gonen2023demystifying} finds that selecting the demonstrations with lower perplexity improves the ICL performance, similarly, \citet{kim2022self} generate the demonstrations from pre-trained LMs, etc. \textbf{(2) Demonstration ordering.} It is found that the ordering of demonstrations can significantly influence the performance~\cite{lu2022fantastically, liu2024let, xu2024context}, as also shown in our experiments in Fig.~\ref{fig:sense}. Specifically,~\citet{lu2022fantastically} detect the optimal demonstration ordering by some synthetic detecting sequences, while~\citet{liu2024let} orders the demonstrations from easy to hard, following a curriculum learning form.

\section{Discussion}

\paragraph{Conclusion.} In this paper, we analyze the current token-based ICL decision boundaries and point out a limitation of using token probability for ICL prediction. To address such a drawback, We propose \M~by decoding the classification logits from centroid classifiers on LM's last hidden states. Our experiments show that \M~is a new state-of-the-art of ICL, with high efficiency on time \& space, data, and prompt engineering. Then, we confirm that \M~indeed creates better classification logits by reducing the inter-class overlap. Moreover, we discover the hidden state convergence promoted by demonstrations, as an explanation of the principle of the performance improvement by a single linear classification boundary in \M. We hope this work can inspire exploration of the ICL by investigating the hidden state instead of token probabilities, and update the community's understanding of ICL calibration.

\paragraph{Comparison to Previous Works.} \textbf{(1) Comparison to Probe Methods.} One concern is that our work can be regarded as a degraded linear probe~\cite{abbas2024enhancing} of the hidden states. However, we believe our work has more advantages: In terms of application, we use fewer samples and require no gradient-based training, which makes our method more user-friendly, efficient, elegant, and interpretable. Moreover, compared to fitting a universal approximation~\cite{hornik1989multilayer}, our method and settings fully utilize the hidden state convergence on decoder LMs (described in~\S\ref{sec:5.2}), making it a true ICL practice. \textbf{(2) Comparison to Supervised Fine-tuning.} Some practices~\cite{gu2023pre, min2022metaicl} build training objectives to fine-tune models for better ICL performance. These efforts are efficient but costly, while our work avoids such an enormous overhead, making it more usable and elegant. \textbf{(3) Comparison to Other Calibrations.} Our method can be seen as a disruptive innovation for methods based on token probability (even the ones based on the whole vocabulary). Experimental comparisons of these methods have been given throughout this paper.

\newpage

\section{Limitations}

Due to computability limitations, we are not able to compare the performance of \M~with the baseline based on supervised fine-tuning. However, we believe that \M~is not within the same methodology as the fine-tuning method, due to the significant difference in computational cost. So such a lack of comparison will not seriously hurt the soundness of this paper.

We argue that human intuition in the label token choice is not reliable. However, we have not eliminated such human intuition completely from the ICL loop: when we build prompts, we still choose the label token. How to automatically select the optimal label token in the prompt will be an important issue, leaving as future research directions for improving the performance of ICL further.

Other label probability calibrations (e.g.\ Batch Calibration) can be combined with \M~for further performance improvements, since the $0$-point is not necessarily an exact classification boundary, as shown in Fig.~\ref{fig:7_dynam_intro}. Also, more complex prompts can be used. However, due to space constraints, we have not attempted this incremental approach, remaining it for an empirically possible practice.

Observation in~\S\ref{sec:5.2} needs more theoretical and experimental analysis. As we can see, some models (\GPT) do not benefit from demonstrations even through the lens of hidden state clustering or \M, which needs to be explained. An explanation of ``why such clustering occurs or not'', and ``how to enhance the intra-class clustering by fine-tuning or prompt engineering'' will be considerably beneficial for understanding ICL.

\section*{Acknowledgments}

This work is supported by the Nakajima Foundation.


\newpage

\bibliography{custom}

\appendix

\section{Experimental Details}
\subsection{Datasets}
\label{Appendix:Dataset}
In this paper, 10 datasets are used as shown in Table~\ref{tab:dataset}. Some datasets do not provide valid splitting, so we randomly split all of them into calibration sets and test sets: For each dataset, we first shuffle it with the random seed \verb|42|. Then, we choose the 512 data at the tail as the testing data, and the 512 data at the head (all the datasets have more than 1024 examples) as the calibration data. Each data point in a test set is used once for each experiment trial to build a prompt example and test for performance.

\texttt{AGNews} and \texttt{GLUE-RTE} have over-length examples. So, in the main experiments, we filter out those examples: for~\LLAMA, when $k=8$, we filter out all the examples with a string length greater than 512 in \texttt{AGNews} and 128 in \texttt{GLUE-RTE}. Also, for Llama 3-8B, when $k=8$, we filter out all the examples with a string length greater than 128 in GLUE-RTE and omit the experiments on \texttt{AGNews}. In the experiments in~\S\ref{sec:5.2}, for all the models, we filter out all the examples with a string length greater than 256 for all the $k$.

\subsection{Baselines}
\label{Appendix:Baselines}

6 baselines (1 vanilla and 5 improved) are used in this paper. Here we introduce the 5 improved baseline.

\paragraph{Contextual Calibration (Con.C).} Proposed by~\citet{zhao2021calibrate}, Con.C uses empty queries with normal demonstrations as calibration samples to estimate the calibration term $A$. In detail, Con.C inputs $m$ samples with empty queries into the model and gets the averaged normalized label probabilities $\Bar{p}'$ among $m$ samples. We take the reciprocal of the probabilities as calibration term $A=\Bar{p}'^{-1}$, while the $B=\mathbf{0}$.

\paragraph{Batch Calibration (Bat.C).} Proposed by~\citet{zhou2023batch}, Bat.C is an inference-time calibration, using the negative averaged normalized label probabilities $-\Bar{p}$ of $m$ samples in inference time as the calibration term $B=-\Bar{p}$, while the $A=\mathbf{1}$, where $\mathbf{1}$ is the all-one vector. 

\paragraph{Domain Calibration (Dom.C).} Proposed by~\citet{fei2023mitigating}, Dom.C acts similarly to the Con.C, with the difference that it uses a random sequence sampled on the random tokens from the calibration dataset as queries instead of empty ones. We fix the sampled length to 32.

\paragraph{KNN Prompt (KNN).} Proposed by~\citet{xu2022k}, KNN uses the whole output vocabulary probability distribution as the classification feature, instead of the label tokens. In detail, first, features of calibration examples are calculated as $k$-NN anchors. Then, during the inference, a $k$-NN classifier is used to classify the feature from the test samples. We use $m$ examples to calculate the anchors for k-NN, and the nearest neighbor number is set to 3.

\paragraph{Central Calibration (Cent.C).} This is the control method proposed by us with a calculation process completely consistent with the \M, except that the usage of the hidden state is replaced by the whole output vocabulary probability distribution consistent with KNN. This method compares with \M~to prove that the output probability distribution is not a good classification feature for ICL in a controlled setting.

\vspace{0.2em}

Notice that: these label-probability-based methods (Con.C, Bat.C, Dom.C) use $A$ or $B$ \textit{along}, which may be another major drawback of these calibration methods: According to Fig.~\ref{fig:2_Motivation}, if a calibration rotates the \textcolor[HTML]{0070c0}{mapping direction} suitably, and transfer the $0$-point properly, a decision boundary close to the \textcolor[HTML]{4ea72e}{\M} can be found. This also leads to a new research direction for calibration: the simultaneous usage of translation and rotation methods.

\subsection{Prompts}
\label{Appendix:Prompts}

\begin{table}[t] 
    \centering
    \caption{Datasets and Abbreviations used in this paper.}
    \vspace{-0.5\baselineskip}
    \label{tab:dataset}
    \resizebox{\columnwidth}{!}{
    \begin{tabular}{lc}
    \toprule
      \textbf{Dataset} & \textbf{Abbr.} \\
    \midrule
      \texttt{AGNews}~\cite{Zhang2015CharacterlevelCN} & AGNews \\
      \SER~\cite{pontikietal2014semeval} & SemE.R\\
      \SEL~\cite{pontikietal2014semeval} & SemE.L\\
      \texttt{Poem Sentiment}~\cite{sheng2020investigating} & PoemS\\
      \texttt{GLUE-RTE}~\cite{wang2019glue} & RTE\\
      \texttt{tweet\_eval\_emotion}~\cite{mohammad2018semeval} & TEE\\
      \texttt{tweet\_eval\_hate}~\cite{basile-etal-2019-semeval} & TEH \\
      \texttt{tweet\_eval\_sentiment}~\cite{rosenthal2017semeval} & TES\\
      \texttt{financial\_phrasebank (all agree)}~\cite{Malo2014GoodDO} & FP\\
      \texttt{rotten\_tomatoes}~\cite{Pang+Lee:05a} & Rott.T\\
    \bottomrule
    \end{tabular}}
\end{table}

\begin{table}[t] 
    \centering
    \caption{Prompt templates used in this paper.}
    \vspace{-0.5\baselineskip}
    \label{tab:prompt}
    \resizebox{\linewidth}{!}{
    \begin{tabular}{lll}
    \toprule
      \textbf{Dataset} & \textbf{Prompt Template} & \textbf{Verbalizer} \\
    \midrule
      AGNews & Input: <x>, Label: <y> & world, sport, business, science\\
      SemE.R & Input: <x>, Aspect: <a>, Label: <y> & positive, neutral, negative\\
      SemE.L & Input: <x>, Aspect: <a>, Label: <y> & positive, neutral, negative\\
      PoemS & Input: <x>, Label: <y> & positive, neutral, negative, mix\\
      RTE & Input: <x>, Text 2: <a>, Label: <y> & include, neutral\\
      TEE & Input: <x>, Label: <y> & anger, joy, positive, sad\\
      TEH  & Input: <x>, Label: <y> & normal, hate\\
      TES & Input: <x>, Label: <y> & positive, neutral, negative\\
      FP & Input: <x>, Label: <y> & positive, neutral, negative\\
      Rott.T & Input: <x>, Label: <y> & positive, negative\\
    \bottomrule
    \end{tabular}}
\end{table}

\begin{table}[t] 
    \centering
    \caption{The 7 prompt templates used in the experiment (on \texttt{Rotten\_Tomatoes}) of Fig.~\ref{fig:sense} (left).}
    \label{tab:moreprompt}
    \resizebox{0.65\linewidth}{!}{
    \begin{tabular}{cl}
    \toprule
      \textbf{\#} & \textbf{Prompt Template}\\
    \midrule
      Original & Input: <x>, Label: <y>\\
      1 & Sentence: <x>, Label: <y>\\
      2 & sentence: <x>, Label: <y>\\
      3 & sentence:$\backslash$n <x>, Label: <y>\\
      4 & Input: <x>, Sentiment: <y>\\
      5 & Input: <x>, sentiment: <y>\\
      6 & x: <x>, y: <y>\\
    \bottomrule
    \end{tabular}}
\end{table}

In this paper, we use a minimum prompt template shown in Table~\ref{tab:prompt}. The separator between demonstrations is ``$\backslash$n''.

To facilitate the replication of label probability-based methods, we limit all the labels in the label space to one token by synonymous conversion. Note that \M~does not need to meet such a one-token requirement. 

Especially, in \S\ref{sec:4.3}, we use 6 more prompt templates to test the stability of each ICL method against the prompt templates. We list these extra templates in Table~\ref{tab:moreprompt}.

\subsection{Details of Visualization in \S\ref{sec:3.1}}
\label{Appendix:Visual}

\paragraph{Principle Component Analysis (PCA).} Given a hidden state set $\mathcal{H} = \left\{ h^{(i)} \right\}^{n}_{i=1}$, we span all the hidden state vector into a matrix $H\in\mathbb{R}^{n\times d}$. The covariance matrix is $\mathrm{cov}\left( H \right) = \frac{1}{n}\left(H-\Bar{H}\right)^T\left(H-\Bar{H}\right)$, where the $\Bar{H}$ is the matrix spanned by the element-wise average vectors $\Bar{h}$ of hidden state set $\mathcal{H}$. We conduct Eigenvalue Decomposition on $\mathrm{cov}\left( H \right)$ and adjust the dimensions to arrange the eigenvalues $\Lambda$ in a descending order along the row:
\begin{equation}
    \mathrm{cov}\left( H \right) = P\Lambda P^{T},
\end{equation}
\noindent where the $P\in\mathbb{R}^{d\times d}$ is an orthogonal matrix. Taking the top-$\Tilde{d}$ lines of $P$ and span them into $\Tilde{P}\in\mathbb{R}^{d\times\Tilde{d}}$, we get the principle component mapping:
\begin{equation}
    \mathrm{PCA}_{\mathcal{H}}(h) = \left(h-\Bar{h}\right)\Tilde{P}= h\Tilde{P} - \Bar{h}\Tilde{P}.
\end{equation}
Notice that $\Tilde{P}\Tilde{P}^T = I$, where $I$ is the identity matrix.

\paragraph{Dot-product after PCA.} Suppose we have dot-product with vector\footnote{Due to excessive superscripts, in this section, we omit the superscripts $U$ in the notation of un-embedding $E_l^U$.} $h$ and $E$ in the original space $\mathbb{R}^d$, producing the dot-product similarity classification criterion $\alpha$:
\begin{equation}
    \alpha = h\left(E^T - \mathbf{0}^T\right).
\end{equation}
When we conduct a same PCA on both $h$ and $E^T$ to get dot-product similarity in a dimensionality-reduced space similar to Fig.~\ref{fig:2_Motivation}:
\begin{align}
    \Tilde{\alpha} &= \mathrm{PCA}_{\mathcal{H}}\left(h\right)\underbrace{\left(\mathrm{PCA}_{\mathcal{H}}\left(E\right)^T - \mathrm{PCA}_{\mathcal{H}}\left(\mathbf{0}\right)^T\right)}_{\text{\textcolor[HTML]{0070c0}{Mapping direction selected after PCA}}}\\
    &= \left(h\Tilde{P} - \Bar{h}\Tilde{P}\right) \left(E\Tilde{P}\right)^T \\
    &= h\Tilde{P}\Tilde{P}^TE^T - \Bar{h}\Tilde{P}\Tilde{P}^TE^T \\
    &= \alpha - \Bar{h}E^T.
\end{align}
Notice that we use the mapping direction $\left(\mathrm{PCA}_{\mathcal{H}}\left(E\right)^T - \mathrm{PCA}_{\mathcal{H}}\left(\mathbf{0}\right)^T\right)$ after the PCA, instead of $\left(\mathrm{PCA}_{\mathcal{H}}\left(E\right)^T-\mathbf{0}^T\right)$, and this is the reason why the \textcolor[HTML]{0070c0}{oblique axis} in Fig.~\ref{fig:2_Motivation} does not necessarily pass through the coordinate origin. In such a scenario, the dot productions after PCA only differ by a fixed constant bias $-\Bar{h}E^T$ from the ones before PCA. This is the reason why the normal line of \textcolor[HTML]{0070c0}{oblique axis} on the $0$-point doesn't pass the coordinate origin of the 2D-plane in Fig.~\ref{fig:2_Motivation}.

\paragraph{Decision Boundary after PCA.} Notice that the decision boundary of two classes $l_1$ and $l_2$ in an non-rotated ICL scenario is:
\begin{equation}
    \mathcal{B} = \left\{ h| hE_{l_1}^T - hE_{l_2}^T = C \right\}.
\end{equation}
Where the $C$ is the calibration term without rotation. Notice that it is a hyperplane in $\mathbb{R}^d$ with normal vector $\left( E_{l_1} - E_{l_2} \right)^T$. Also, the normal plane which pass the $0$-point of direction $\left( E_{l_1} - E_{l_2} \right)^T$ in $\mathbb{R}^{\Tilde{d}}$ after PCA is:
\begin{equation}
\begin{split}
    \Tilde{\mathcal{B}} &= \{ \mathrm{PCA}_{\mathcal{H}}(h)| \mathrm{PCA}_{\mathcal{H}}(h)\\ &( \mathrm{PCA}_{\mathcal{H}}(E_{l_1} - E_{l_2}) - \mathrm{PCA}_{\mathcal{H}}(\mathbf{0}))^T = 0 \}.
\end{split}
\end{equation}
By the aforementioned transformation, we have:
\begin{equation}
    \Tilde{\mathcal{B}} = \left\{ \mathrm{PCA}_{\mathcal{H}}(h) | hE_{l_1}^T - hE_{l_2}^T = \Bar{h}\left(E_{l_1}^T-E_{l_2}^T\right) \right\}.
\end{equation}
That is, the dimensionality-reduced decision boundary $\Tilde{\mathcal{B}}$ is perpendicular to the mapped direction $\left( \mathrm{PCA}_{\mathcal{H}}\left(E_{l_1} - E_{l_2}\right) - \mathrm{PCA}_{\mathcal{H}}(\mathbf{0})\right)$, and biased only by a constant $\left(\Bar{h}\left(E_{l_1}^T-E_{l_2}^T\right) - C\right)$ on the classification criteria comparing to the original space. Specifically, in the two-dimensional case, it is a straight line that may not necessarily pass through the coordinate origin, as shown in Fig.~\ref{fig:2_Motivation}.

\subsection{Details of Experiment in \S\ref{sec:5.1}}
\label{Appendix:ExpDetails_5.1}

\subsubsection{Calculation Details of Averaged Overlap}

First, we divide the $\vert\mathcal{Y}\vert$-way classification task into $\mathbb{C}(\vert\mathcal{Y}\vert, 2)$ 2-way classification task\footnote{The $\mathbb{C}(m, n)$ is the $n$-combination number from $m$ elements.}, to allow us to use a scalar to characterize the classification criteria for each 2-combination (similar to what we do to the ``positive'' and ``negative'' examples in Fig.~\ref{fig:2_Motivation}). Then, for each chosen 2-combination, w.l.o.g, given labels denoted as $l_1$ and $l_2$, we build prompt-label sets\footnote{Notice that the $T$ is the prompting function.} as:
\begin{equation}
    \mathcal{S}_{l_j} = \left\{ T\left(\mathcal{D}^{de, (i)}, x^{(c_i)}\right) \Big| y^{(c_i)}={l_j}\right\}_{l_j\in\{l_1, l_2\}}^{n_{l_j}},
\end{equation}
\noindent where $c_i$ is the sampled query index. That is, we sample queries annotated with these two labels and build prompt sets, then collect the prompts with the same query label $l_j$ into $\mathcal{S}_{l_j}$, with a size $n_{l_j}$.

\begin{figure*}[t]
    \centering
    \includegraphics[width=0.9\linewidth]{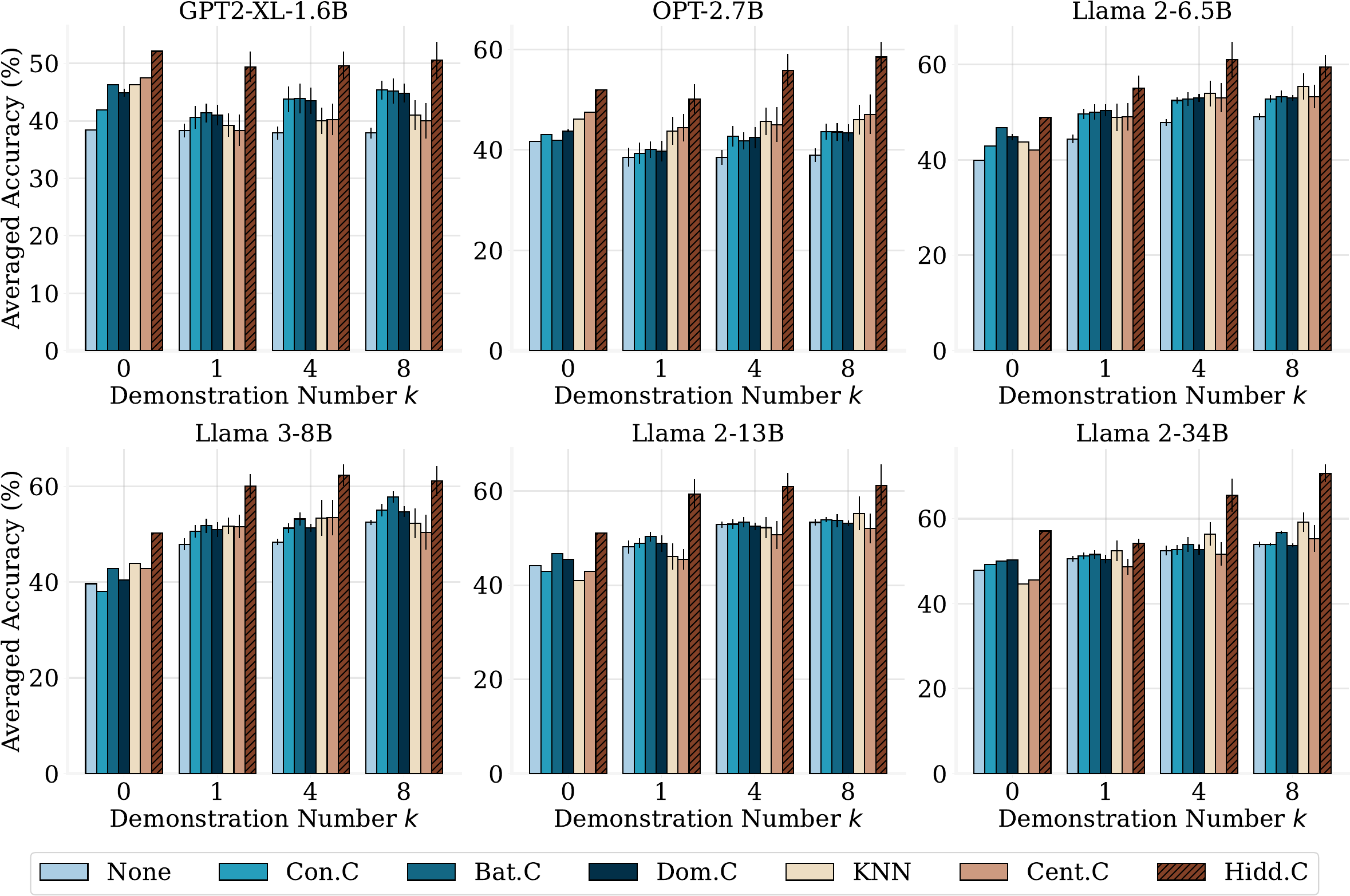}
    \caption{The classification performance (Accuracy(\%)) of 3 models averaged on 10 datasets.}
    \label{fig:Appendix.14.accuracy}
\end{figure*}

Then, for each prompt $s^{(i)} = T\left(\mathcal{D}^{de, (i)}, x^{(i)}\right)\in\mathcal{S}_{l_j}$, we run decoders (vanilla, Con.C, Dom.C and \M) with probability normlization $f_{l_1}(\cdot)$ and $f_{l_2}(\cdot)$ to get the classification probabilities of assigning label $l_1$ and $l_2$ as $\alpha^{(i)}_1=f_{l_1}\left(s^{(i)}\right)$ and $\alpha^{(i)}_2=f_{l_2}\left(s^{(i)}\right)$. We calculate the difference between $\alpha^{(i)}_1$ and $\alpha^{(i)}_2$ and collect them into a set:
\begin{equation}
    \mathcal{A}_{l_j}=\left\{ \alpha^{(i)}_1-\alpha^{(i)}_2 \Big| s^{(i)}\in\mathcal{S}_{l_j} \right\}_{i=1}^{n_{l_j}}.
\end{equation}
Now, for the 2-combination of labels $(l_1, l_2)$, we get $\mathcal{A}_{l_1}$ and $\mathcal{A}_{l_2}$, whose elements are the probabilities difference between assigning $l_1$ and assigning $l_2$ to example $s^{(i)}$. The difference between $\mathcal{A}_{l_1}$ and $\mathcal{A}_{l_2}$ is: the elements in $\mathcal{A}_{l_1}$ are from $s^{(i)}$s with queries labeled by ground-truth $l_1$, and vice versa. We obtain continuous probability density functions of $\mathcal{A}_{l_1}$ and $\mathcal{A}_{l_2}$ as $p_{l_1}(\cdot)$ and $p_{l_2}(\cdot)$ by kernel density estimation, as the curves in Fig. \ref{fig:5_Ana1_intro}.

Then, we calculate the overlap area of these curves:
\begin{equation}
    S_{l_1,l_2}=\int_{-1}^{1}\min\left[p_{l_1}(x), p_{l_2}(x)\right]\mathrm{d}x.
\end{equation}
For each combination\footnote{Notice that on $S_{\cdot,\cdot}$, the labels are rotational symmetry.} in the $\mathbb{C}(\vert\mathcal{Y}\vert, 2)$ 2-combinations, we repeat to calculate the $S_{\cdot,\cdot}$, and average them as the \textbf{Averaged Overlap} $\Bar{S}$.
\begin{equation}
    \Bar{S}=\frac{1}{\mathbb{C}(\vert\mathcal{Y}\vert, 2)}\sum_{i=1}^{\vert\mathcal{Y}\vert} \sum_{j=i+1}^{\vert\mathcal{Y}\vert} S_{l_i,l_j}.
\end{equation}

\subsubsection{Experimental Details in \S\ref{sec:5.1}}
\label{Appendix:OverlapExpDetails}

We conduct experiments resulting Fig.~\ref{fig:6_Ana1_OPT_res} on 3 models with \SER, \SEL, AGNews, Poem Sentiment, and fiancial\_phrasebank, given the demonstration number $k=4$ and calibration example numbers $m=16$. We use the whole 512 examples on the test split for each dataset and repeat 5 times.

\subsubsection{Proof: \textit{the Overlap Area is Double to the Error's Lower Bound}}
\label{Appendix:Proof}

Suppose a label combination $l_1$ and $l_2$, w.l.o.g., we have a ground truth probability density function $p_{l_1}(x)$ and $p_{l_2}(x)$ on a criterion $x\in\mathbb{X}$, same as the curves in Fig.~\ref{fig:5_Ana1_intro}. Given a specific value of criterion $x$, the upper-bound classification performance is determined by majority vote, which is the most accurate method on such a point, resulting in a density of error classification:
\begin{equation}
    e(x)_{l_1, l_2} \geqslant \min\left[p_{l_1}(x), p_{l_2}(x)\right].
\end{equation}

So, the integral error rate:
\begin{align}
    \mathcal{E}_{l_1, l_2} &\geqslant \frac{\int_{x\in\mathbb{X}}\min\left[p_{l_1}(x), p_{l_2}(x)\right]\mathrm{d}x}{\int_{x\in\mathbb{X}}p_{l_1}(x)dx+\int_{x\in\mathbb{X}}p_{l_2}(x)\mathrm{d}x} \\
    &= \frac{1}{2}\int_{x\in\mathbb{X}}\min\left[p_{l_1}(x), p_{l_2}(x)\right]\mathrm{d}x \\
    &= \frac{1}{2}S_{l_1, l_2}.
\end{align}

\subsection{Details of Experiment in \S\ref{sec:5.2}}
\label{Appendix:ExpDetails_5.2}

\subsubsection{Calculation of the Distance and Standard Error}

\begin{figure*}[t]
    \centering
    \includegraphics[width=1\linewidth]{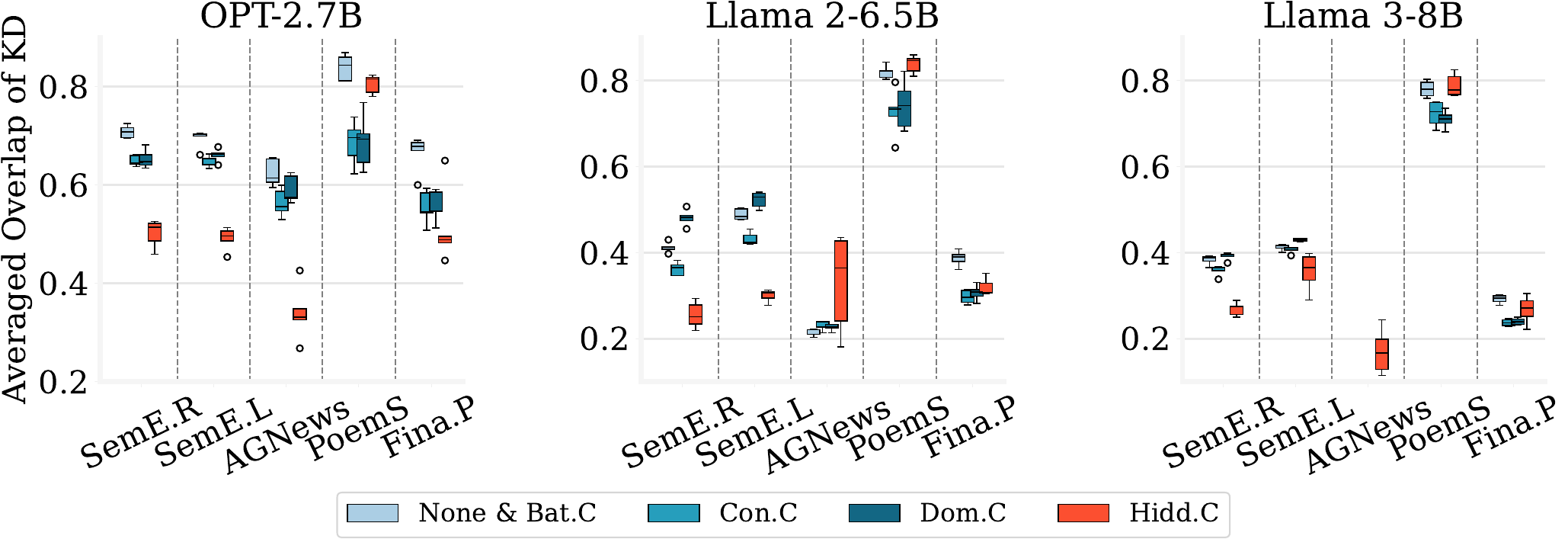}
    \caption{The augmented results on 2 models of Fig.~\ref{fig:6_Ana1_OPT_res}.}
    \label{fig:Appendix.12.overlap}
\end{figure*}

\begin{table}[t]
\centering
\caption{Transferability of centroid among various datasets with the same label space. Big numbers are the averaged improvement (MF1) compared to vanilla ICL, small numbers are standard error. Statistically significant results ($p<0.1$) are in \textbf{bold}.}
\label{table:2_D2DTransfer}
\resizebox{\columnwidth}{!}{
\begin{tabular}{@{}c|cccc@{}}
\toprule
 \diagbox[trim=lr]{\textbf{Test}}{\textbf{Cali.}}&
  \textbf{SemE.R} &
  \textbf{SemE.L} &
  \textbf{Fina.P} &
  \textbf{TES} \\ \midrule
\textbf{SemE.R} &
  \textbf{\begin{tabular}[c]{@{}c@{}}(+38.75)\\[-1.2ex]{\tiny $\pm2.28$}\end{tabular}} &
  \textbf{\begin{tabular}[c]{@{}c@{}}+29.24\\[-1.2ex]{\tiny $\pm3.19$}\end{tabular}} &
  \begin{tabular}[c]{@{}c@{}}+6.32\\[-1.2ex]{\tiny $\pm10.55$}\end{tabular} &
  \begin{tabular}[c]{@{}c@{}}+7.54\\[-1.2ex]{\tiny $\pm8.96$}\end{tabular} \\
\textbf{SemE.L} &
  \textbf{\begin{tabular}[c]{@{}c@{}}+20.78\\[-1.2ex]{\tiny $\pm7.37$}\end{tabular}} &
  \textbf{\begin{tabular}[c]{@{}c@{}}(+37.33)\\[-1.2ex]{\tiny $\pm3.47$}\end{tabular}} &
  \begin{tabular}[c]{@{}c@{}}\textcolor[HTML]{ee442f}{-0.40}\\[-1.2ex]{\tiny $\pm7.37$}\end{tabular} &
  \begin{tabular}[c]{@{}c@{}}+8.94\\[-1.2ex]{\tiny $\pm8.93$}\end{tabular} \\
\textbf{Fina.P} &
  \textbf{\begin{tabular}[c]{@{}c@{}}+7.42\\[-1.2ex]{\tiny $\pm4.98$}\end{tabular}} &
  \begin{tabular}[c]{@{}c@{}}+9.05\\[-1.2ex]{\tiny $\pm11.14$}\end{tabular} &
  \textbf{\begin{tabular}[c]{@{}c@{}}(+37.29)\\[-1.2ex]{\tiny $\pm2.30$}\end{tabular}} &
  \begin{tabular}[c]{@{}c@{}}\textcolor[HTML]{ee442f}{-4.35}\\[-1.2ex]{\tiny $\pm6.34$}\end{tabular} \\
\textbf{TES} &
  \begin{tabular}[c]{@{}c@{}}+6.95\\[-1.2ex]{\tiny $\pm7.00$}\end{tabular} &
  \textbf{\begin{tabular}[c]{@{}c@{}}+9.73\\[-1.2ex]{\tiny $\pm5.68$}\end{tabular}} &
  \begin{tabular}[c]{@{}c@{}}\textcolor[HTML]{ee442f}{-0.51}\\[-1.2ex]{\tiny $\pm3.83$}\end{tabular} &
  \textbf{\begin{tabular}[c]{@{}c@{}}(+11.83)\\[-1.2ex]{\tiny $\pm3.59$}\end{tabular}} \\ \bottomrule
\end{tabular}
}
\end{table}

\begin{table}[t]
\centering
\caption{Transferability of centroid among various $k$ on the same dataset. $k_1\rightarrow k_2$ is to use centroids estimated by $k_1$ demonstrations for inference on test examples with $k_2$ demonstrations. Other annotations are the same as Table.~\ref{table:2_D2DTransfer}}
\label{table:3_E2ETransfer}
\resizebox{\columnwidth}{!}{
\begin{tabular}{@{}c|ccc|cc@{}}
\toprule
 &
  \textbf{0$\rightarrow$1} &
  \textbf{4$\rightarrow$1} &
  \textbf{(1$\rightarrow$1)} &
  \textbf{1$\rightarrow$4} &
  \textbf{(4$\rightarrow$4)} \\ \midrule
\textbf{SemE.R} &
  \textbf{\begin{tabular}[c]{@{}c@{}}+9.46\\[-1.2ex] {\tiny $\pm1.95$}\end{tabular}} &
  \textbf{\begin{tabular}[c]{@{}c@{}}+22.50\\[-1.2ex] {\tiny $\pm14.55$}\end{tabular}} &
  \textbf{\begin{tabular}[c]{@{}c@{}}(+26.14)\\[-1.2ex] {\tiny $\pm5.16$}\end{tabular}} &
  \textbf{\begin{tabular}[c]{@{}c@{}}+17.95\\[-1.2ex] {\tiny $\pm7.51$}\end{tabular}} &
  \textbf{\begin{tabular}[c]{@{}c@{}}(+38.75)\\[-1.2ex] {\tiny $\pm2.28$}\end{tabular}} \\
\textbf{SemE.L} &
  \textbf{\begin{tabular}[c]{@{}c@{}}+26.80\\[-1.2ex] {\tiny $\pm3.20$}\end{tabular}} &
  \textbf{\begin{tabular}[c]{@{}c@{}}+17.18\\[-1.2ex] {\tiny $\pm5.61$}\end{tabular}} &
  \textbf{\begin{tabular}[c]{@{}c@{}}(+26.65)\\[-1.2ex] {\tiny $\pm2.72$}\end{tabular}} &
  \begin{tabular}[c]{@{}c@{}}+10.79\\[-1.2ex] {\tiny $\pm14.86$}\end{tabular} &
  \textbf{\begin{tabular}[c]{@{}c@{}}(+37.33)\\[-1.2ex] {\tiny $\pm3.47$}\end{tabular}} \\
\textbf{AGNews} &
  \textbf{\begin{tabular}[c]{@{}c@{}}+42.38\\[-1.2ex] {\tiny $\pm2.42$}\end{tabular}} &
  \textbf{\begin{tabular}[c]{@{}c@{}}+40.20\\[-1.2ex] {\tiny $\pm1.24$}\end{tabular}} &
  \textbf{\begin{tabular}[c]{@{}c@{}}(+41.02)\\[-1.2ex] {\tiny $\pm2.49$}\end{tabular}} &
  \textbf{\begin{tabular}[c]{@{}c@{}}+43.12\\[-1.2ex] {\tiny $\pm2.02$}\end{tabular}} &
  \textbf{\begin{tabular}[c]{@{}c@{}}(+46.66)\\[-1.2ex] {\tiny $\pm3.77$}\end{tabular}} \\
\textbf{PoemS} &
  \begin{tabular}[c]{@{}c@{}}+0.16\\[-1.2ex] {\tiny $\pm1.87$}\end{tabular} &
  \begin{tabular}[c]{@{}c@{}}+2.12\\[-1.2ex] {\tiny $\pm6.18$}\end{tabular} &
  \textbf{\begin{tabular}[c]{@{}c@{}}(+21.49)\\[-1.2ex] {\tiny $\pm2.54$}\end{tabular}} &
  \textbf{\begin{tabular}[c]{@{}c@{}}+8.79\\[-1.2ex] {\tiny $\pm1.84$}\end{tabular}} &
  \textbf{\begin{tabular}[c]{@{}c@{}}(+12.96)\\[-1.2ex] {\tiny $\pm1.52$}\end{tabular}} \\
\textbf{Fina.P} &
  \begin{tabular}[c]{@{}c@{}}\textcolor[HTML]{ee442f}{-0.13}\\[-1.2ex] {\tiny $\pm1.88$}\end{tabular} &
  \textbf{\begin{tabular}[c]{@{}c@{}}+21.40\\[-1.2ex] {\tiny $\pm2.90$}\end{tabular}} &
  \textbf{\begin{tabular}[c]{@{}c@{}}(+16.70)\\[-1.2ex] {\tiny $\pm3.80$}\end{tabular}} &
  \textbf{\begin{tabular}[c]{@{}c@{}}+10.00\\[-1.2ex] {\tiny $\pm13.68$}\end{tabular}} &
  \textbf{\begin{tabular}[c]{@{}c@{}}(+37.30)\\[-1.2ex] {\tiny $\pm2.30$}\end{tabular}} \\ \bottomrule
\end{tabular}
}
\end{table}

\paragraph{Averaged Centroid Distance.} Given a $\vert\mathcal{Y}\vert$-way classification task, for each label $l$ we build its corresponding prompt set $\mathcal{S}_l=\left\{s^{(c_i)}|y^{(c_i)}=l\right\}_{i=1}^{n_l}$, where $s^{(c_i)}$ is the prompt with query labeled by $l$, and $c_i$ is the sampled query index. We encode it into a hidden state set $\mathcal{H}_l=\left\{h^{(i)}\right\}_{i=1}^{n_l}$, and calculate its centroid $\Bar{h}_l$, as what we do in \M:
\begin{equation}
    \Bar{h}_l = \frac{1}{n_l}\sum_{h^{(i)}\in\mathcal{H}_l}h^{(i)}.
\end{equation}
For every 2-combination of labels $l$ and $l'$, we calculate the distance of their centroid, and the average among all the 2-combination is used as the Averaged Centroid Distance:
\begin{equation}
    \mathrm{ACD} = \frac{1}{\mathbb{C}(\left\vert\mathcal{Y}\right\vert, 2)}\sum_{i=1}^{\vert\mathcal{Y}\vert} \sum_{j=i+1}^{\vert\mathcal{Y}\vert} \left\Vert\Bar{h}_{i} - \Bar{h}_{j}\right\Vert_2.
\end{equation}

\begin{figure}[t]
    \centering
    \includegraphics[width=\linewidth]{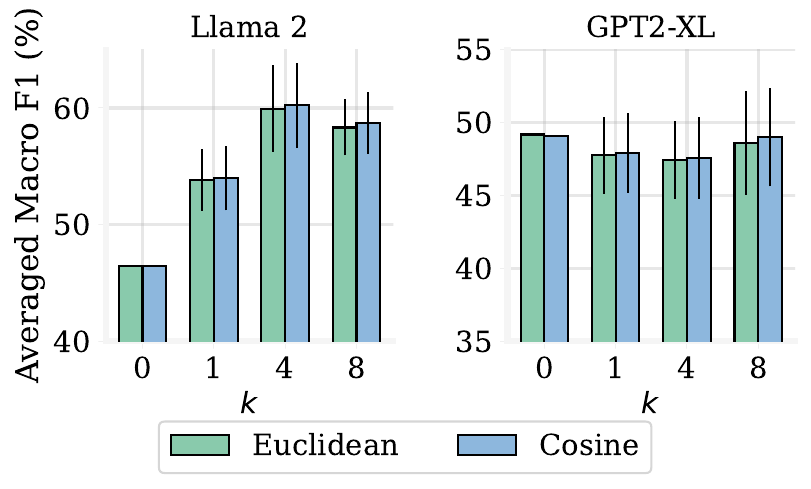}
    \caption{The classification performance (Macro F1(\%)) of \M~with difference similarity measure.}
        \label{fig:Appendix.13.similarity}
\end{figure}

\paragraph{Averaged Intra-class Standard Error.} Given the hidden state set $\mathcal{H}_l=\left\{h^{(i)}\right\}_{i=1}^{n_l}$ w.r.t. the label $l$, we span all the hidden state vectors into a matrix $H_l\in\mathbb{R}^{n_l\times d}$. The covariance matrix is $\left(H_l-\Bar{H}_l\right)^T\left(H_l-\Bar{H}_l\right)$, where the $\Bar{H}_l$ is the matrix spanned by the element-wise average vectors of hidden state set $\mathcal{H}_l$. Notice that the $\mathrm{ACD}$ is a first-order moment, for a proper comparison, we use the average on the diagonal elements of the element-wise square root of the covariance matrix as the intra-class standard error metric for label $l$. We average all the standard errors from all the classes as the Averaged Intra-class Standard Error:
\begin{equation}
    \mathrm{AIS} = \frac{1}{\vert\mathcal{Y}\vert d}\sum_{i=1}^{\vert\mathcal{Y}\vert}\mathrm{tr}\left[\sqrt{\left(H_i-\Bar{H}_i\right)^T\left(H_i-\Bar{H}_i\right)}\right].
\end{equation}

\subsubsection{Experimental Details in \S\ref{sec:5.2}}

We conduct experiments resulting Fig.~\ref{fig:10_Dynam_COV} on 4 models with \SER, \SEL, AGNews, Poem Sentiment, and fiancial\_phrasebank, given the calibration example numbers $m=16$. We use the whole 512 examples on the test split for each dataset and repeat 5 times.

\subsection{Experimental Details for Fig.~\ref{fig:11_Data_eff}}
\label{Appendix:ExpDetails_5.3}

We conduct experiments resulting Fig.~\ref{fig:11_Data_eff} on \OPT~with 4 datasets: \SER, \SEL, AGNews, and Poem Sentiment, given the demonstration numbers $k = 4$ and repeat 5 times.

\section{Detailed Results}

\subsection{Details of Main Results}
\label{Appendix:Detailed_Results_mainres}

Numerical details of Fig.~\ref{fig:4_Mainres} are shown in Table~\ref{table:Appendix.GPT21},~\ref{table:Appendix.OPT3},~\ref{table:Appendix.LLAMA22},~\ref{table:Appendix.LLAMA3},~\ref{table:Appendix.LLAMA213B} and~\ref{table:Appendix.LLAMA234B}. Accuracy results are shown in Fig.~\ref{fig:Appendix.14.accuracy}.

\subsection{Details of Averaged Overlaps Results}
\label{Appendix:ResDetails_5.1}

\begin{table*}[t] 
    \centering
    \caption{Performance of all 6 models on TREC for a more-way classification, and on Hate\_Speech18 for a biased dataset. $k=4$, top-2 results are in \textbf{bold}.}
    \label{tab:two_more_dataset}
    \resizebox{0.85\linewidth}{!}{
    \begin{tabular}{@{}ccccccc@{}}
\toprule
                & \textbf{GPT2-XL-1.6B} & \textbf{OPT-2.7B} & \textbf{Llama 2-6.5B} & \textbf{Llama 3-8B} & \textbf{Llama 2-13B} & \textbf{Llama 2-34B} \\ \midrule
\multicolumn{7}{c}{\texttt{TREC}}                                                                                                                                \\
\textbf{None}   & 13.02                 & 15.85             & 23.12                 & 16.79               & 23.22                & 21.13                \\
\textbf{Con.C}  & 14.10                 & 8.12              & 23.72                 & 18.34               & 23.56                & 22.35                \\
\textbf{Bat.C}  & 14.44                 & 17.31             & 23.26                 & 20.28               & 22.88                & 22.15                \\
\textbf{Dom.C}  & 14.10                  & 8.58              & 23.14                 & 19.10                & 23.42                & 23.23                \\
\textbf{KNN}    & \textbf{27.53}        & 33.80             & 49.25                 & 43.15               & 54.76                & \textbf{53.66}       \\
\textbf{Cent.C} & 23.46                 & \textbf{33.80}     & \textbf{49.18}        & \textbf{46.12}      & \textbf{56.79}       & 53.03                \\
\rowcolor[HTML]{EFEFEF} 
\textbf{Hidd.C} & \textbf{55.91}        & \textbf{61.14}    & \textbf{64.90}         & \textbf{71.59}      & \textbf{75.64}       & \textbf{68.39}       \\ \midrule
\multicolumn{7}{c}{\texttt{Hate\_Speech18}}                                                                                                                      \\
\textbf{None}   & \textbf{23.96}        & 23.95             & \textbf{23.96}        & \textbf{23.96}      & \textbf{23.96}       & 23.92                \\
\textbf{Con.C}  & 21.57                 & 25.47             & 14.68                 & \textbf{23.74}      & 21.49                & 21.25                \\
\textbf{Bat.C}  & 20.59                 & 21.25             & 20.54                 & 18.25               & 20.66                & 23.68                \\
\textbf{Dom.C}  & 21.23                 & 23.68             & 20.73                 & 23.67               & \textbf{24.36}       & 23.28                \\
\textbf{KNN}    & 17.30                  & 16.74             & 16.50                  & 16.26               & 16.23                & \textbf{27.05}       \\
\textbf{Cent.C} & 16.32                 & \textbf{25.47}    & 15.42                 & 5.46                & 9.15                 & 22.23                \\
\rowcolor[HTML]{EFEFEF} 
\textbf{Hidd.C} & \textbf{24.47}        & \textbf{27.03}    & \textbf{23.44}        & 20.67               & 20.06                & \textbf{29.94}       \\ \bottomrule
\end{tabular}}
\end{table*}

The augmented results on the other 3 models (we skip this experiment on the quantitated model) of Fig.~\ref{fig:6_Ana1_OPT_res} are shown in Fig.~\ref{fig:Appendix.12.overlap}.

\section{Additional Discussion}
\subsection{The Similarity Measures Used in \M}
\label{Appendix:Similarity}

In~\S\ref{sec:3.2}, we use the Euclidean distance as the similarity measure, while this is not the only option. Intuitively, we can choose other similarity measures as alternatives. Moreover, since we get inspired by observation with dot-production similarity, we should check the performance on such a measure instead of the Euclidean distance. This section uses cosine similarity as an example to illustrate that there is no significant performance difference between these measures. We use cosine similarity to repeat the results in \S\ref{sec:4.2} on \LLAMA~and \GPT.

The results are shown in Fig.~\ref{fig:Appendix.13.similarity}, where the performance based on these two measures is close, without statistical difference. This indicates that the hidden space has good properties of both metric and vector space, and \M~acts equally on these measures.

\subsection{Transferability of the Centorid}
\label{sec.appendix.transf}

We have proven that it is not advisable to use the \textit{common} token probability criteria, while, since the centroid criteria are proven to be better than token probability, we are curious: can the centroid calculated in one task be transferred to other tasks with the same label space? Among the datasets sharing the same label space ``positive'', ``neutral'', and ``negative'', we calculate centroids by one dataset and evaluate \M~with it on another dataset, on \OPT, with $k=4$, $m=16|\mathcal{Y}|$. The results are shown in Table~\ref{table:2_D2DTransfer}, where only limited transferability is demonstrated in different domains of the same task (SemE.R and SemE.L), whose behavior is similar to \textit{task vector}~\cite{ilharco2022editing, hendel2023context}, while other combination of datasets can not demonstrate considerable transferability. This further exacerbates our doubts about the token-based method: We find that the hidden state distributions have significant differences among various datasets, even if they share a common label space, then utilizing fixed token un-embedding vectors to decode these classification criteria is highly unreliable.

Moreover, we repeat this experiment on various $k$, instead of various datasets, as shown in Table~\ref{table:3_E2ETransfer}. The transferabilities among $k$ are better than on datasets, but still worse than the un-transferred scenario. Notice that $4\rightarrow1$ results are much better than $0\rightarrow1$, which support our results in~\S\ref{sec:5.2}: hidden states with higher $k$ are further converged.

\subsection{A Demonstration towards ICL Principles}

Our findings may lead to an explanation of the principle of ICL and traditional calibrations. LMs generate distributed representations into separate clusters in the last hidden state. At this point, by dot-product, any non-collinear \textit{arbitrary or plausible} mapping directions should be able to capture and classify these clusters to some extent. Note: The absolute distance in such a direction is not faithful (since the centroids of these hidden states and the coordinate origins in these mapping directions are not necessarily aligned), which leads to the generation of so-called bias, and calibrating these biases can improve the performance to a certain extent. However, in such a paradigm, high-dimensional features are discarded, resulting in overlapping originally linearly separable features in high-dimensional space, leading to a loss of classification accuracy, even if the calibration aligns the coordinate origin. 


\subsection{Applicability on More-way Classification and Biased Dataset}

To further verify the applicability of \M~on harder tasks, we test the performance of \M~and baseline methods on TREC~\cite{li-roth-2002-learning, hovy-etal-2001-toward} for more-way ($\vert\mathcal{Y}\vert$ = 6) classification and Hate\_Speech18~\cite{gibert2018hate} for a biased dataset (label frequency distribution: $[0.87, 0.11, 0.007, 0.015]$), shown in Table~\ref{tab:two_more_dataset}. For results on TREC, \M~produces a significant improvement compared to all the baselines on all the models. While, \M~outperforms on Hate\_Speech18 in most cases, and sometimes \M~is weaker than baseline methods, but consistently produces competitive results. However, we believe that this slightly weaker result cannot be fully attributed to biased datasets, given that there are biased datasets among the 10 standard datasets where \M~perform SotA results (e.g. \SER, refer \S\ref{Appendix:Detailed_Results_mainres}). 

\section{Statements}
\subsection{Author Contributions Statement}
\label{Appendix:ACS}


The vast majority of contributions of this paper are attributed to Hakaze Cho (also called Yufeng Zhao), who provided ideas, designed and conducted experiments, collected and described data, wrote papers, and revised them.

Y.S., M.K., K.T., and A.I. participated in our discussion and provided some comments, with sequentially decreasing participation and contribution. They also helped write a non-peer-viewed seminar version of this paper in Japanese.

Naoya Inoue is their supervisor, he provides an excellent laboratory environment, funding, necessary guidance, and paper revision.

\subsection{License for Artifacts}

\paragraph{Models.} \GPT~and \OPT~is under the MIT license, Llama family is under its specific license.

\paragraph{Datasets.} We list the open-source license for the datasets used in this paper as follows:

\begin{itemize}
    \item \texttt{CC-by-4.0}: \texttt{Poem Sentiment}, \SER, \SEL, \texttt{tweet\_eval\_emotion}, \texttt{tweet\_eval\_hate}, \texttt{tweet\_eval\_hate}
    \item \texttt{CC-by-SA-3.0}: \texttt{financial\_phrasebank}, \texttt{GLUE-RTE}
    \item Unknown: \texttt{AGNews}, \texttt{rotten\_tomatoes}
\end{itemize}


\paragraph{Consistency of Usage.} Models and data are used with their original usage.

\subsection{AI Agent Usage}

AI Agents are only used for writing improving and grammar checking in this paper.

\begin{sidewaystable*}[t]
\centering
\caption{Classification performance (Macro F1(\%)) on \GPT. $\mathrm{mean}_{\mathrm{std}}$, top-2 results are in \textbf{bold}.}
\resizebox{0.85\textwidth}{!}{
\begin{tabular}{@{}ccccccccccccc@{}}
\toprule
\multicolumn{2}{c}{GPT-2 XL} & \textbf{AGNews} & \textbf{SemE.R} & \textbf{SemE.L} & \textbf{PoemS} & \textbf{RTE} & \textbf{TEE} & \textbf{TEH} & \textbf{TES} & \textbf{Fina.P} & \textbf{Rott.T} & \textbf{Average} \\ \midrule
 & None & 16.53$_{0.00}$ & 9.87$_{0.00}$ & 12.31$_{0.00}$ & 8.75$_{0.00}$ & 48.31$_{0.00}$ & 19.40$_{0.00}$ & 37.56$_{0.00}$ & 21.14$_{0.00}$ & 25.36$_{0.00}$ & 34.16$_{0.00}$ & 23.34 \\
 & Con.C & 30.04$_{0.00}$ & 9.87$_{0.00}$ & 12.31$_{0.00}$ & 8.11$_{0.00}$ & 47.46$_{0.00}$ & 9.68$_{0.00}$ & 42.97$_{0.00}$ & 27.01$_{0.00}$ & 33.14$_{0.00}$ & 60.75$_{0.00}$ & 28.13 \\
 & Bat.C & 42.56$_{0.00}$ & 37.29$_{0.00}$ & \textbf{46.21$_{0.00}$} & 18.92$_{0.00}$ & \textbf{51.29$_{0.00}$} & 26.39$_{0.00}$ & \textbf{44.45$_{0.00}$} & \textbf{42.52$_{0.00}$} & 32.00$_{0.00}$ & \textbf{66.96$_{0.00}$} & 40.86 \\
 & Dom.C & 38.00$_{0.60}$ & 24.76$_{0.04}$ & 26.58$_{1.06}$ & 20.52$_{0.99}$ & 38.00$_{0.60}$ & 14.34$_{0.00}$ & 37.56$_{0.00}$ & \textbf{42.70$_{1.00}$} & 30.14$_{1.06}$ & \textbf{67.42$_{2.19}$} & 34.33 \\ \cmidrule(l){2-13} 
 & KNN & 50.63$_{0.00}$ & \textbf{39.62$_{0.00}$} & 42.31$_{0.00}$ & 26.69$_{0.00}$ & \textbf{49.57$_{0.00}$} & 30.79$_{0.00}$ & \textbf{44.68$_{0.00}$} & 34.12$_{0.00}$ & 39.30$_{0.00}$ & 62.37$_{0.00}$ & 42.01 \\
 & Cent.C & \textbf{52.54$_{0.00}$} & 39.53$_{0.00}$ & 45.37$_{0.00}$ & \textbf{27.23$_{0.00}$} & 44.92$_{0.00}$ & \textbf{32.71$_{0.00}$} & 41.93$_{0.00}$ & 32.15$_{0.00}$ & \textbf{43.15$_{0.00}$} & 63.32$_{0.00}$ & \textbf{42.29} \\
\multirow{-7}{*}{$k=0$} & \cellcolor[HTML]{EFEFEF}\textbf{Hidd.C} & \cellcolor[HTML]{EFEFEF}\textbf{82.02$_{0.00}$} & \cellcolor[HTML]{EFEFEF}\textbf{44.73$_{0.00}$} & \cellcolor[HTML]{EFEFEF}\textbf{54.45$_{0.00}$} & \cellcolor[HTML]{EFEFEF}\textbf{32.81$_{0.00}$} & \cellcolor[HTML]{EFEFEF}47.36$_{0.00}$ & \cellcolor[HTML]{EFEFEF}\textbf{43.45$_{0.00}$} & \cellcolor[HTML]{EFEFEF}42.95$_{0.00}$ & \cellcolor[HTML]{EFEFEF}36.78$_{0.00}$ & \cellcolor[HTML]{EFEFEF}\textbf{47.38$_{0.00}$} & \cellcolor[HTML]{EFEFEF}59.83$_{0.00}$ & \cellcolor[HTML]{EFEFEF}\textbf{49.18} \\ \midrule
 & None & 20.95$_{1.29}$ & 36.72$_{1.19}$ & 31.60$_{1.33}$ & 21.21$_{1.72}$ & 49.47$_{2.22}$ & 22.81$_{0.94}$ & 37.56$_{0.00}$ & 31.74$_{1.43}$ & 30.15$_{1.97}$ & 36.20$_{0.56}$ & 31.84 \\
 & Con.C & 24.15$_{1.13}$ & 41.88$_{1.73}$ & 38.92$_{3.32}$ & 24.82$_{2.55}$ & 47.64$_{3.38}$ & 20.70$_{1.20}$ & 37.56$_{0.00}$ & 33.06$_{1.29}$ & 34.93$_{1.31}$ & 60.21$_{3.98}$ & 36.39 \\
 & Bat.C & 30.02$_{1.49}$ & \textbf{45.04$_{0.77}$} & \textbf{41.10$_{4.41}$} & 25.20$_{0.48}$ & \textbf{49.58$_{1.88}$} & 25.85$_{2.51}$ & \textbf{48.02$_{0.94}$} & \textbf{34.92$_{0.91}$} & \textbf{35.02$_{1.14}$} & \textbf{64.72$_{1.92}$} & \textbf{39.95} \\
 & Dom.C & 22.17$_{1.15}$ & 44.86$_{1.16}$ & 41.02$_{4.91}$ & \textbf{25.61$_{1.27}$} & 46.81$_{2.09}$ & 18.92$_{1.58}$ & 37.56$_{0.00}$ & \textbf{33.96$_{0.90}$} & 34.72$_{1.56}$ & 62.27$_{3.33}$ & 36.79 \\ \cmidrule(l){2-13} 
 & KNN & \textbf{32.14$_{0.85}$} & 36.88$_{2.76}$ & 37.29$_{3.20}$ & 21.44$_{1.57}$ & 48.18$_{2.09}$ & \textbf{27.44$_{1.75}$} & 43.52$_{3.73}$ & 33.85$_{1.08}$ & \textbf{36.23$_{2.07}$} & 52.41$_{1.44}$ & 36.94 \\
 & Cent.C & 26.74$_{2.18}$ & 33.00$_{1.48}$ & 32.02$_{3.63}$ & 18.75$_{2.78}$ & 47.82$_{3.27}$ & 23.82$_{3.95}$ & 44.07$_{1.52}$ & 30.68$_{2.12}$ & 29.58$_{1.98}$ & 58.41$_{4.30}$ & 34.49 \\
\multirow{-7}{*}{$k=1$} & \cellcolor[HTML]{EFEFEF}\textbf{Hidd.C} & \cellcolor[HTML]{EFEFEF}\textbf{65.15$_{1.77}$} & \cellcolor[HTML]{EFEFEF}\textbf{49.16$_{3.43}$} & \cellcolor[HTML]{EFEFEF}\textbf{51.56$_{2.83}$} & \cellcolor[HTML]{EFEFEF}\textbf{32.83$_{2.32}$} & \cellcolor[HTML]{EFEFEF}\textbf{50.47$_{1.41}$} & \cellcolor[HTML]{EFEFEF}\textbf{36.17$_{2.74}$} & \cellcolor[HTML]{EFEFEF}\textbf{49.16$_{1.47}$} & \cellcolor[HTML]{EFEFEF}33.55$_{4.45}$ & \cellcolor[HTML]{EFEFEF}44.02$_{3.36}$ & \cellcolor[HTML]{EFEFEF}\textbf{65.55$_{2.47}$} & \cellcolor[HTML]{EFEFEF}\textbf{47.76} \\ \midrule
 & None & 21.87$_{4.32}$ & 33.14$_{1.46}$ & 41.03$_{2.14}$ & 20.11$_{1.47}$ & 40.48$_{1.15}$ & 17.98$_{0.33}$ & 38.19$_{1.41}$ & 29.06$_{1.70}$ & 28.86$_{2.07}$ & 33.81$_{0.72}$ & 30.45 \\
 & Con.C & 24.22$_{10.00}$ & 44.76$_{0.98}$ & \textbf{48.90$_{2.71}$} & 21.95$_{1.28}$ & 36.33$_{1.11}$ & 24.40$_{0.82}$ & 37.51$_{0.11}$ & \textbf{37.30$_{2.49}$} & 41.65$_{2.96}$ & 69.57$_{1.11}$ & 38.66 \\
 & Bat.C & 26.97$_{9.22}$ & 44.48$_{1.67}$ & 46.94$_{1.72}$ & 21.93$_{1.24}$ & 47.17$_{1.96}$ & \textbf{28.86$_{1.88}$} & 46.94$_{2.76}$ & \textbf{36.49$_{2.64}$} & \textbf{44.82$_{2.54}$} & \textbf{71.94$_{1.24}$} & \textbf{41.65} \\
 & Dom.C & 25.30$_{10.23}$ & \textbf{45.44$_{1.54}$} & 47.01$_{2.20}$ & \textbf{23.05$_{1.05}$} & 36.79$_{1.68}$ & 27.11$_{1.72}$ & 37.79$_{0.51}$ & 34.67$_{0.81}$ & 42.54$_{2.53}$ & 67.00$_{3.89}$ & 38.67 \\ \cmidrule(l){2-13} 
 & KNN & \textbf{33.93$_{2.04}$} & 37.57$_{3.54}$ & 38.35$_{2.35}$ & 21.72$_{3.05}$ & \textbf{48.55$_{2.97}$} & 26.20$_{1.96}$ & \textbf{48.71$_{2.74}$} & 30.10$_{2.12}$ & 36.57$_{2.56}$ & 57.36$_{1.72}$ & 37.91 \\
 & Cent.C & 32.98$_{2.56}$ & 37.24$_{5.80}$ & 32.71$_{4.50}$ & 18.47$_{1.88}$ & 45.78$_{2.93}$ & 24.21$_{4.17}$ & \textbf{48.83$_{3.57}$} & 29.75$_{2.50}$ & 33.30$_{7.56}$ & 58.81$_{2.86}$ & 36.21 \\
\multirow{-7}{*}{$k=4$} & \cellcolor[HTML]{EFEFEF}\textbf{Hidd.C} & \cellcolor[HTML]{EFEFEF}\textbf{49.55$_{3.29}$} & \cellcolor[HTML]{EFEFEF}\textbf{50.81$_{2.16}$} & \cellcolor[HTML]{EFEFEF}\textbf{54.16$_{3.62}$} & \cellcolor[HTML]{EFEFEF}\textbf{24.96$_{2.56}$} & \cellcolor[HTML]{EFEFEF}\textbf{49.28$_{2.42}$} & \cellcolor[HTML]{EFEFEF}\textbf{39.13$_{2.80}$} & \cellcolor[HTML]{EFEFEF}48.48$_{1.62}$ & \cellcolor[HTML]{EFEFEF}34.43$_{2.46}$ & \cellcolor[HTML]{EFEFEF}\textbf{50.80$_{3.51}$} & \cellcolor[HTML]{EFEFEF}\textbf{72.70$_{2.19}$} & \cellcolor[HTML]{EFEFEF}\textbf{47.43} \\ \midrule
 & None & 19.23$_{0.78}$ & 32.79$_{2.13}$ & 33.36$_{1.02}$ & 17.93$_{1.08}$ & 37.37$_{2.07}$ & 15.75$_{0.75}$ & 37.56$_{0.00}$ & 25.15$_{0.99}$ & 26.45$_{1.13}$ & 33.55$_{0.45}$ & 27.91 \\
 & Con.C & 18.38$_{0.28}$ & \textbf{47.06$_{3.77}$} & \textbf{52.40$_{0.96}$} & 19.70$_{2.05}$ & 35.98$_{0.00}$ & 26.80$_{2.00}$ & 37.56$_{0.00}$ & \textbf{38.12$_{3.15}$} & 41.05$_{2.25}$ & \textbf{75.05$_{2.12}$} & 39.21 \\
 & Bat.C & 22.32$_{1.22}$ & 45.96$_{2.08}$ & 48.25$_{0.79}$ & \textbf{21.20$_{1.55}$} & \textbf{45.82$_{4.68}$} & \textbf{29.23$_{1.22}$} & 47.84$_{1.36}$ & \textbf{38.30$_{3.09}$} & \textbf{48.78$_{2.06}$} & 74.81$_{1.58}$ & \textbf{42.25} \\
 & Dom.C & 21.85$_{0.95}$ & 45.78$_{2.48}$ & 49.91$_{0.74}$ & 20.80$_{1.97}$ & 35.98$_{0.00}$ & 28.50$_{1.30}$ & 37.56$_{0.60}$ & 36.20$_{2.44}$ & 47.99$_{1.84}$ & 69.87$_{4.70}$ & 39.44 \\ \cmidrule(l){2-13} 
 & KNN & \textbf{39.81$_{1.81}$} & 38.83$_{4.38}$ & 34.38$_{3.72}$ & 20.92$_{2.98}$ & 43.03$_{5.31}$ & 26.49$_{3.84}$ & \textbf{49.00$_{1.23}$} & 29.33$_{3.74}$ & 39.00$_{1.91}$ & 60.08$_{3.64}$ & 38.09 \\
 & Cent.C & 41.52$_{4.18}$ & 37.73$_{3.15}$ & 32.21$_{2.80}$ & 14.66$_{4.00}$ & 45.11$_{6.67}$ & 21.83$_{6.60}$ & 48.97$_{3.71}$ & 25.81$_{2.72}$ & 31.32$_{3.89}$ & 55.23$_{4.35}$ & 35.44 \\
\multirow{-7}{*}{$k=8$} & \cellcolor[HTML]{EFEFEF}\textbf{Hidd.C} & \cellcolor[HTML]{EFEFEF}\textbf{57.10$_{5.54}$} & \cellcolor[HTML]{EFEFEF}\textbf{49.86$_{2.55}$} & \cellcolor[HTML]{EFEFEF}\textbf{58.26$_{3.15}$} & \cellcolor[HTML]{EFEFEF}\textbf{24.42$_{1.28}$} & \cellcolor[HTML]{EFEFEF}\textbf{48.48$_{4.48}$} & \cellcolor[HTML]{EFEFEF}\textbf{35.48$_{5.04}$} & \cellcolor[HTML]{EFEFEF}\textbf{51.50$_{2.18}$} & \cellcolor[HTML]{EFEFEF}27.46$_{6.30}$ & \cellcolor[HTML]{EFEFEF}\textbf{56.99$_{1.72}$} & \cellcolor[HTML]{EFEFEF}\textbf{76.50$_{3.14}$} & \cellcolor[HTML]{EFEFEF}\textbf{48.60} \\ \bottomrule
\end{tabular}}
\label{table:Appendix.GPT21}
\end{sidewaystable*} 

\begin{sidewaystable*}[t]
\centering
\caption{Classification performance (Macro F1(\%)) on \OPT. $\mathrm{mean}_{\mathrm{std}}$, top-2 results are in \textbf{bold}.}
\resizebox{0.85\textwidth}{!}{
\begin{tabular}{@{}ccccccccccccc@{}}
\toprule
\multicolumn{2}{c}{OPT 2.7B} & \textbf{AGNews} & \textbf{SemE.R} & \textbf{SemE.L} & \textbf{PoemS} & \textbf{RTE} & \textbf{TEE} & \textbf{TEH} & \textbf{TES} & \textbf{Fina.P} & \textbf{Rott.T} & \textbf{Average} \\ \midrule
 & None & 27.67$_{0.00}$ & 24.62$_{0.00}$ & 31.07$_{0.00}$ & 25.00$_{0.00}$ & 51.19$_{0.00}$ & 24.43$_{0.00}$ & 39.31$_{0.00}$ & 34.08$_{0.00}$ & \textbf{47.23$_{0.00}$} & 40.33$_{0.00}$ & 34.49 \\
 & Con.C & 20.72$_{0.00}$ & 20.77$_{0.00}$ & 29.33$_{0.00}$ & 16.51$_{0.00}$ & 34.63$_{0.00}$ & 16.25$_{0.00}$ & 31.37$_{0.00}$ & 30.65$_{0.00}$ & 44.88$_{0.00}$ & \textbf{67.48$_{0.00}$} & 31.26 \\
 & Bat.C & 28.58$_{0.00}$ & 31.32$_{0.00}$ & 34.31$_{0.00}$ & 25.44$_{0.00}$ & \textbf{53.23$_{0.00}$} & 26.81$_{0.00}$ & 43.60$_{0.00}$ & 34.34$_{0.00}$ & 45.25$_{0.00}$ & \textbf{66.51$_{0.00}$} & 38.94 \\
 & Dom.C & 27.55$_{0.06}$ & 20.53$_{0.05}$ & 30.90$_{0.84}$ & 16.72$_{0.40}$ & 45.07$_{0.12}$ & 18.01$_{0.18}$ & 43.30$_{0.12}$ & 31.11$_{0.24}$ & \textbf{46.16$_{0.47}$} & 66.23$_{1.28}$ & 34.56 \\ \cmidrule(l){2-13} 
 & KNN & 52.50$_{0.00}$ & 31.39$_{0.00}$ & \textbf{44.16$_{0.00}$} & 25.72$_{0.00}$ & \textbf{52.08$_{0.00}$} & 36.15$_{0.00}$ & 51.60$_{0.00}$ & \textbf{39.45$_{0.00}$} & 42.26$_{0.00}$ & 57.16$_{0.00}$ & 43.25 \\
 & Cent.C & \textbf{55.97$_{0.00}$} & \textbf{31.65$_{0.00}$} & 43.23$_{0.00}$ & \textbf{35.86$_{0.00}$} & 41.98$_{0.00}$ & \textbf{41.27$_{0.00}$} & \textbf{53.44$_{0.00}$} & 35.25$_{0.00}$ & 37.54$_{0.00}$ & 58.60$_{0.00}$ & \textbf{43.48} \\
\multirow{-7}{*}{$k=0$} & \cellcolor[HTML]{EFEFEF}\textbf{Hidd.C} & \cellcolor[HTML]{EFEFEF}\textbf{75.01$_{0.00}$} & \cellcolor[HTML]{EFEFEF}\textbf{41.94$_{0.00}$} & \cellcolor[HTML]{EFEFEF}\textbf{52.14$_{0.00}$} & \cellcolor[HTML]{EFEFEF}\textbf{39.92$_{0.00}$} & \cellcolor[HTML]{EFEFEF}45.64$_{0.00}$ & \cellcolor[HTML]{EFEFEF}\textbf{45.90$_{0.00}$} & \cellcolor[HTML]{EFEFEF}\textbf{52.93$_{0.00}$} & \cellcolor[HTML]{EFEFEF}\textbf{35.67$_{0.00}$} & \cellcolor[HTML]{EFEFEF}43.24$_{0.00}$ & \cellcolor[HTML]{EFEFEF}61.71$_{0.00}$ & \cellcolor[HTML]{EFEFEF}\textbf{49.41} \\ \midrule
 & None & 24.16$_{0.71}$ & 18.76$_{0.50}$ & 24.98$_{1.74}$ & 12.23$_{1.19}$ & \textbf{50.75$_{3.01}$} & 20.72$_{3.33}$ & \textbf{51.55$_{4.28}$} & 23.64$_{1.19}$ & 31.40$_{1.83}$ & 48.94$_{1.90}$ & 30.71 \\
 & Con.C & 22.57$_{1.38}$ & 20.15$_{1.78}$ & 24.48$_{1.88}$ & 13.17$_{0.63}$ & \textbf{50.75$_{3.01}$} & 24.30$_{1.69}$ & \textbf{51.69$_{4.03}$} & 23.02$_{0.00}$ & 28.88$_{1.95}$ & 65.63$_{1.72}$ & 32.46 \\
 & Bat.C & 27.43$_{1.13}$ & 20.48$_{0.90}$ & 26.78$_{1.17}$ & 14.21$_{1.05}$ & 50.25$_{3.25}$ & 26.44$_{1.41}$ & 50.45$_{1.76}$ & 23.34$_{2.30}$ & 28.81$_{0.24}$ & \textbf{70.88$_{0.65}$} & 33.91 \\
 & Dom.C & 24.06$_{0.94}$ & 19.68$_{1.26}$ & 24.24$_{1.73}$ & 13.78$_{2.27}$ & \textbf{50.75$_{3.01}$} & 23.58$_{2.54}$ & 51.54$_{3.86}$ & 23.14$_{0.83}$ & 28.81$_{1.72}$ & \textbf{69.98$_{1.81}$} & 32.96 \\ \cmidrule(l){2-13} 
 & KNN & 48.15$_{2.50}$ & \textbf{42.35$_{4.06}$} & 39.01$_{5.24}$ & 25.52$_{1.94}$ & \textbf{53.07$_{3.06}$} & 32.62$_{3.02}$ & 49.78$_{0.58}$ & \textbf{31.90$_{3.64}$} & 36.59$_{3.67}$ & 57.31$_{4.89}$ & \textbf{41.63} \\
 & Cent.C & \textbf{49.21$_{2.81}$} & 39.46$_{4.26}$ & \textbf{42.30$_{4.65}$} & \textbf{29.60$_{5.30}$} & 48.48$_{3.85}$ & \textbf{34.32$_{2.51}$} & 50.69$_{1.21}$ & 31.00$_{1.34}$ & \textbf{36.69$_{2.28}$} & 53.70$_{2.51}$ & 41.54 \\
\multirow{-7}{*}{$k=1$} & \cellcolor[HTML]{EFEFEF}\textbf{Hidd.C} & \cellcolor[HTML]{EFEFEF}\textbf{65.18$_{2.39}$} & \cellcolor[HTML]{EFEFEF}\textbf{44.91$_{5.14}$} & \cellcolor[HTML]{EFEFEF}\textbf{51.62$_{2.09}$} & \cellcolor[HTML]{EFEFEF}\textbf{33.72$_{2.24}$} & \cellcolor[HTML]{EFEFEF}50.40$_{1.80}$ & \cellcolor[HTML]{EFEFEF}\textbf{45.00$_{2.79}$} & \cellcolor[HTML]{EFEFEF}49.53$_{2.92}$ & \cellcolor[HTML]{EFEFEF}\textbf{35.02$_{0.47}$} & \cellcolor[HTML]{EFEFEF}\textbf{48.10$_{3.33}$} & \cellcolor[HTML]{EFEFEF}57.72$_{1.42}$ & \cellcolor[HTML]{EFEFEF}\textbf{48.12} \\ \midrule
 & None & 22.91$_{1.05}$ & 20.84$_{1.16}$ & 25.44$_{1.60}$ & 12.46$_{1.29}$ & 49.70$_{3.22}$ & 14.83$_{0.16}$ & 40.68$_{0.44}$ & 23.78$_{0.99}$ & 28.62$_{1.13}$ & 53.73$_{0.95}$ & 29.30 \\
 & Con.C & 22.81$_{1.04}$ & 20.70$_{1.02}$ & 27.74$_{1.79}$ & 12.62$_{1.32}$ & \textbf{50.06$_{3.86}$} & 21.92$_{1.03}$ & 41.21$_{0.70}$ & 25.86$_{1.65}$ & 36.74$_{1.68}$ & \textbf{83.50$_{1.82}$} & 34.32 \\
 & Bat.C & 25.40$_{1.04}$ & 20.07$_{1.40}$ & 26.49$_{0.86}$ & 11.46$_{1.24}$ & 47.61$_{2.78}$ & 23.92$_{0.96}$ & 45.71$_{1.33}$ & 26.91$_{1.24}$ & 38.88$_{1.70}$ & 82.22$_{1.13}$ & 34.87 \\
 & Dom.C & 22.32$_{1.17}$ & 20.71$_{1.39}$ & 25.96$_{1.97}$ & 12.64$_{1.24}$ & \textbf{50.37$_{3.22}$} & 21.74$_{1.30}$ & 41.30$_{0.54}$ & 26.15$_{1.41}$ & 38.01$_{1.69}$ & \textbf{83.99$_{1.45}$} & 34.32 \\ \cmidrule(l){2-13} 
 & KNN & 49.36$_{2.28}$ & 49.71$_{5.26}$ & 47.66$_{3.87}$ & 22.18$_{1.17}$ & 48.74$_{3.76}$ & \textbf{32.05$_{4.50}$} & \textbf{47.35$_{3.18}$} & \textbf{30.76$_{3.92}$} & \textbf{40.78$_{1.86}$} & 66.33$_{4.19}$ & \textbf{43.49} \\
 & Cent.C & \textbf{49.91$_{4.46}$} & \textbf{51.72$_{2.80}$} & \textbf{48.96$_{3.00}$} & \textbf{21.63$_{3.55}$} & 47.02$_{2.37}$ & 29.84$_{3.43}$ & 46.99$_{2.05}$ & 25.39$_{3.05}$ & 40.23$_{8.32}$ & 65.80$_{4.54}$ & 42.75 \\
\multirow{-7}{*}{$k=4$} & \cellcolor[HTML]{EFEFEF}\textbf{Hidd.C} & \cellcolor[HTML]{EFEFEF}\textbf{69.56$_{3.62}$} & \cellcolor[HTML]{EFEFEF}\textbf{59.59$_{1.97}$} & \cellcolor[HTML]{EFEFEF}\textbf{62.77$_{3.08}$} & \cellcolor[HTML]{EFEFEF}\textbf{25.42$_{0.79}$} & \cellcolor[HTML]{EFEFEF}47.94$_{3.16}$ & \cellcolor[HTML]{EFEFEF}\textbf{47.47$_{5.86}$} & \cellcolor[HTML]{EFEFEF}\textbf{49.46$_{1.80}$} & \cellcolor[HTML]{EFEFEF}\textbf{35.61$_{3.45}$} & \cellcolor[HTML]{EFEFEF}\textbf{65.91$_{2.01}$} & \cellcolor[HTML]{EFEFEF}79.47$_{5.88}$ & \cellcolor[HTML]{EFEFEF}\textbf{54.32} \\ \midrule
 & None & 21.52$_{1.03}$ & 21.24$_{1.48}$ & 24.11$_{1.51}$ & 12.84$_{0.73}$ & 48.05$_{3.65}$ & 14.54$_{0.00}$ & 37.54$_{0.03}$ & 21.85$_{1.07}$ & 28.47$_{1.48}$ & 59.86$_{1.63}$ & 29.00 \\
 & Con.C & 22.26$_{0.33}$ & 23.84$_{1.26}$ & 30.27$_{1.27}$ & 13.02$_{0.89}$ & 47.58$_{3.27}$ & 23.06$_{1.10}$ & 37.80$_{0.48}$ & 25.91$_{0.79}$ & 37.67$_{1.44}$ & 83.42$_{1.74}$ & 34.48 \\
 & Bat.C & 24.38$_{1.49}$ & 22.28$_{1.04}$ & 28.05$_{1.65}$ & 13.31$_{0.50}$ & 47.50$_{2.86}$ & 24.46$_{0.50}$ & 49.30$_{1.63}$ & 27.96$_{1.51}$ & 40.39$_{1.53}$ & 84.03$_{1.31}$ & 36.16 \\
 & Dom.C & 22.38$_{0.70}$ & 21.33$_{1.57}$ & 26.63$_{1.36}$ & 13.18$_{1.08}$ & 47.48$_{3.87}$ & 23.63$_{1.08}$ & 37.59$_{0.25}$ & 25.25$_{1.11}$ & 39.27$_{2.20}$ & \textbf{86.67$_{1.55}$} & 34.34 \\ \cmidrule(l){2-13} 
 & KNN & 47.45$_{1.64}$ & 55.08$_{1.89}$ & 48.16$_{3.38}$ & 22.61$_{2.20}$ & \textbf{49.56$_{2.68}$} & \textbf{34.80$_{3.18}$} & 49.60$_{2.26}$ & \textbf{30.77$_{2.79}$} & 38.73$_{3.52}$ & 66.65$_{5.67}$ & 44.34 \\
 & Cent.C & \textbf{47.50$_{4.46}$} & \textbf{55.83$_{4.46}$} & \textbf{51.06$_{8.42}$} & \textbf{22.95$_{2.22}$} & 48.22$_{2.35}$ & 33.57$_{4.12}$ & \textbf{49.66$_{2.73}$} & 29.18$_{4.37}$ & \textbf{43.87$_{2.57}$} & 67.49$_{8.07}$ & \textbf{44.93} \\
\multirow{-7}{*}{$k=8$} & \cellcolor[HTML]{EFEFEF}\textbf{Hidd.C} & \cellcolor[HTML]{EFEFEF}\textbf{65.27$_{1.64}$} & \cellcolor[HTML]{EFEFEF}\textbf{63.95$_{2.51}$} & \cellcolor[HTML]{EFEFEF}\textbf{66.48$_{2.53}$} & \cellcolor[HTML]{EFEFEF}\textbf{23.70$_{1.76}$} & \cellcolor[HTML]{EFEFEF}\textbf{49.86$_{2.86}$} & \cellcolor[HTML]{EFEFEF}\textbf{52.24$_{2.39}$} & \cellcolor[HTML]{EFEFEF}\textbf{50.57$_{3.28}$} & \cellcolor[HTML]{EFEFEF}\textbf{41.94$_{6.42}$} & \cellcolor[HTML]{EFEFEF}\textbf{71.16$_{4.13}$} & \cellcolor[HTML]{EFEFEF}\textbf{85.78$_{4.39}$} & \cellcolor[HTML]{EFEFEF}\textbf{57.09} \\ \bottomrule
\end{tabular}}
\label{table:Appendix.OPT3}
\end{sidewaystable*}   


\begin{sidewaystable*}[t]
\centering
\caption{Classification performance (Macro F1(\%)) on \LLAMA. $\mathrm{mean}_{\mathrm{std}}$, top-2 results are in \textbf{bold}.}
\resizebox{0.85\textwidth}{!}{
\begin{tabular}{@{}ccccccccccccc@{}}
\toprule
\multicolumn{2}{c}{Llama 2} & \textbf{AGNews} & \textbf{SemE.R} & \textbf{SemE.L} & \textbf{PoemS} & \textbf{RTE} & \textbf{TEE} & \textbf{TEH} & \textbf{TES} & \textbf{Fina.P} & \textbf{Rott.T} & \textbf{Average} \\ \midrule
 & None & 23.74$_{0.00}$ & 44.36$_{0.00}$ & 32.48$_{0.00}$ & 16.23$_{0.00}$ & 35.88$_{0.00}$ & 17.01$_{0.00}$ & 46.06$_{0.00}$ & 30.59$_{0.00}$ & 19.64$_{0.00}$ & 32.90$_{0.00}$ & 29.89 \\
 & Con.C & 22.76$_{0.00}$ & 44.50$_{0.00}$ & 42.83$_{0.00}$ & 10.53$_{0.00}$ & 43.88$_{0.00}$ & 24.06$_{0.00}$ & 42.42$_{0.00}$ & 33.32$_{0.00}$ & 54.92$_{0.00}$ & \textbf{58.97$_{0.00}$} & 37.82 \\
 & Bat.C & 37.26$_{0.00}$ & 49.03$_{0.00}$ & \textbf{54.33$_{0.00}$} & 19.70$_{0.00}$ & \textbf{47.34$_{0.00}$} & \textbf{31.64$_{0.00}$} & \textbf{51.06$_{0.00}$} & \textbf{39.82$_{0.00}$} & \textbf{55.76$_{0.00}$} & \textbf{63.07$_{0.00}$} & \textbf{44.90} \\
 & Dom.C & 28.11$_{1.43}$ & 47.88$_{0.45}$ & 49.74$_{1.40}$ & \textbf{26.58$_{1.27}$} & 39.99$_{0.13}$ & 23.00$_{1.11}$ & 37.48$_{0.00}$ & 33.33$_{0.21}$ & \textbf{61.03$_{0.76}$} & 54.96$_{2.78}$ & 40.21 \\ \cmidrule(l){2-13} 
 & KNN & \textbf{40.24$_{0.00}$} & 47.65$_{0.00}$ & 49.15$_{0.00}$ & 20.56$_{0.00}$ & \textbf{50.93$_{0.00}$} & 26.65$_{0.00}$ & 42.35$_{0.00}$ & 33.63$_{0.00}$ & 46.44$_{0.00}$ & 53.26$_{0.00}$ & 41.09 \\
 & Cent.C & 34.36$_{0.00}$ & \textbf{50.87$_{0.00}$} & 46.36$_{0.00}$ & 17.57$_{0.00}$ & 42.15$_{0.00}$ & 25.09$_{0.00}$ & 42.31$_{0.00}$ & 28.21$_{0.00}$ & 43.81$_{0.00}$ & 49.38$_{0.00}$ & 38.01 \\
\multirow{-7}{*}{$k=0$} & \cellcolor[HTML]{EFEFEF}\textbf{Hidd.C} & \cellcolor[HTML]{EFEFEF}\textbf{62.46$_{0.00}$} & \cellcolor[HTML]{EFEFEF}\textbf{50.90$_{0.00}$} & \cellcolor[HTML]{EFEFEF}\textbf{55.77$_{0.00}$} & \cellcolor[HTML]{EFEFEF}\textbf{22.49$_{0.00}$} & \cellcolor[HTML]{EFEFEF}46.94$_{0.00}$ & \cellcolor[HTML]{EFEFEF}\textbf{34.06$_{0.00}$} & \cellcolor[HTML]{EFEFEF}\textbf{47.37$_{0.00}$} & \cellcolor[HTML]{EFEFEF}\textbf{34.29$_{0.00}$} & \cellcolor[HTML]{EFEFEF}54.84$_{0.00}$ & \cellcolor[HTML]{EFEFEF}56.37$_{0.00}$ & \cellcolor[HTML]{EFEFEF}\textbf{46.49} \\ \midrule
 & None & 13.97$_{1.15}$ & 51.57$_{0.30}$ & 51.50$_{0.47}$ & 22.28$_{0.98}$ & 36.26$_{0.80}$ & 27.80$_{1.58}$ & 40.16$_{0.83}$ & 30.78$_{2.77}$ & 27.89$_{2.18}$ & 55.03$_{1.91}$ & 35.72 \\
 & Con.C & 13.29$_{1.62}$ & 52.36$_{0.58}$ & 54.12$_{0.50}$ & 24.16$_{1.29}$ & 38.53$_{0.92}$ & 27.77$_{1.16}$ & 40.63$_{1.52}$ & 37.39$_{1.86}$ & 37.23$_{0.68}$ & \textbf{75.91$_{2.44}$} & 40.14 \\
 & Bat.C & 22.05$_{1.07}$ & 52.24$_{0.44}$ & 53.77$_{1.27}$ & \textbf{24.75$_{1.51}$} & \textbf{49.77$_{3.25}$} & 28.94$_{0.90}$ & 47.72$_{1.19}$ & \textbf{40.25$_{2.30}$} & 38.43$_{1.91}$ & 75.77$_{1.37}$ & 43.37 \\
 & Dom.C & 11.89$_{1.01}$ & 51.80$_{0.38}$ & 53.98$_{0.45}$ & 24.23$_{1.13}$ & \textbf{49.33$_{3.82}$} & 29.17$_{1.44}$ & 38.23$_{0.85}$ & 39.02$_{1.72}$ & 34.95$_{1.35}$ & \textbf{79.44$_{0.58}$} & 41.20 \\ \cmidrule(l){2-13} 
 & KNN & \textbf{52.50$_{3.58}$} & 56.19$_{5.59}$ & 61.24$_{2.42}$ & 21.91$_{0.91}$ & 48.82$_{1.90}$ & \textbf{30.66$_{3.56}$} & \textbf{49.65$_{3.15}$} & 34.86$_{1.11}$ & \textbf{40.26$_{3.40}$} & 75.26$_{3.34}$ & \textbf{46.87} \\
 & Cent.C & 46.17$_{0.93}$ & \textbf{62.09$_{3.04}$} & \textbf{64.50$_{1.76}$} & 23.29$_{2.57}$ & 46.66$_{3.16}$ & 27.80$_{2.36}$ & 45.63$_{2.33}$ & 33.10$_{2.42}$ & 38.99$_{3.56}$ & 75.16$_{2.33}$ & 46.34 \\
\multirow{-7}{*}{$k=1$} & \cellcolor[HTML]{EFEFEF}\textbf{Hidd.C} & \cellcolor[HTML]{EFEFEF}\textbf{61.88$_{0.68}$} & \cellcolor[HTML]{EFEFEF}\textbf{64.83$_{1.85}$} & \cellcolor[HTML]{EFEFEF}\textbf{69.05$_{2.29}$} & \cellcolor[HTML]{EFEFEF}\textbf{24.26$_{1.37}$} & \cellcolor[HTML]{EFEFEF}48.32$_{2.35}$ & \cellcolor[HTML]{EFEFEF}\textbf{42.77$_{2.38}$} & \cellcolor[HTML]{EFEFEF}\textbf{51.55$_{1.27}$} & \cellcolor[HTML]{EFEFEF}\textbf{40.72$_{1.91}$} & \cellcolor[HTML]{EFEFEF}\textbf{63.79$_{0.93}$} & \cellcolor[HTML]{EFEFEF}71.00$_{5.23}$ & \cellcolor[HTML]{EFEFEF}\textbf{53.82} \\ \midrule
 & None & 9.57$_{0.26}$ & 53.45$_{0.34}$ & 53.76$_{0.62}$ & 19.82$_{0.51}$ & 35.90$_{0.18}$ & 29.30$_{1.47}$ & 37.56$_{0.00}$ & 36.55$_{0.97}$ & 37.86$_{0.78}$ & 69.31$_{1.43}$ & 38.31 \\
 & Con.C & 9.35$_{0.23}$ & 53.49$_{0.33}$ & 54.21$_{0.30}$ & \textbf{27.01$_{1.15}$} & 37.65$_{1.74}$ & \textbf{31.37$_{1.07}$} & 37.56$_{0.00}$ & 39.34$_{0.47}$ & 39.28$_{0.55}$ & 90.86$_{0.35}$ & 42.01 \\
 & Bat.C & 19.33$_{1.51}$ & 52.58$_{0.64}$ & 54.40$_{0.69}$ & 24.62$_{0.99}$ & 46.81$_{5.22}$ & 31.26$_{1.58}$ & 48.81$_{2.93}$ & 38.47$_{0.33}$ & 37.05$_{0.60}$ & 86.62$_{1.15}$ & 44.00 \\
 & Dom.C & 9.49$_{0.32}$ & 52.12$_{0.79}$ & 54.14$_{0.28}$ & \textbf{26.47$_{0.87}$} & 39.59$_{2.03}$ & 30.75$_{1.06}$ & 37.56$_{1.43}$ & \textbf{39.52$_{0.22}$} & 34.71$_{0.58}$ & \textbf{91.84$_{0.88}$} & 41.62 \\ \cmidrule(l){2-13} 
 & KNN & \textbf{66.47$_{6.82}$} & 60.00$_{0.92}$ & 62.87$_{2.54}$ & 21.76$_{3.20}$ & 54.41$_{4.95}$ & 29.96$_{2.19}$ & 49.46$_{2.22}$ & 31.90$_{3.97}$ & \textbf{49.09$_{5.20}$} & 88.18$_{1.17}$ & \textbf{51.41} \\
 & Cent.C & 60.62$_{4.62}$ & \textbf{63.61$_{2.06}$} & \textbf{64.55$_{3.77}$} & 21.42$_{4.73}$ & \textbf{55.50$_{5.61}$} & 28.41$_{2.51}$ & \textbf{50.72$_{3.29}$} & 21.12$_{5.70}$ & 42.48$_{2.81}$ & 89.55$_{0.74}$ & 49.80 \\
\multirow{-7}{*}{$k=4$} & \cellcolor[HTML]{EFEFEF}\textbf{Hidd.C} & \cellcolor[HTML]{EFEFEF}\textbf{68.22$_{6.82}$} & \cellcolor[HTML]{EFEFEF}\textbf{65.64$_{1.38}$} & \cellcolor[HTML]{EFEFEF}\textbf{71.38$_{2.00}$} & \cellcolor[HTML]{EFEFEF}25.47$_{2.10}$ & \cellcolor[HTML]{EFEFEF}\textbf{58.00$_{4.38}$} & \cellcolor[HTML]{EFEFEF}\textbf{48.48$_{4.32}$} & \cellcolor[HTML]{EFEFEF}\textbf{51.83$_{3.03}$} & \cellcolor[HTML]{EFEFEF}\textbf{49.44$_{3.20}$} & \cellcolor[HTML]{EFEFEF}\textbf{68.79$_{9.71}$} & \cellcolor[HTML]{EFEFEF}\textbf{92.26$_{0.18}$} & \cellcolor[HTML]{EFEFEF}\textbf{59.95} \\ \midrule
 & None & 8.03$_{0.00}$ & 53.90$_{0.31}$ & 54.12$_{0.38}$ & 20.47$_{0.49}$ & 35.98$_{0.00}$ & 30.77$_{1.50}$ & 37.56$_{0.00}$ & 38.59$_{1.26}$ & 39.25$_{1.04}$ & 73.77$_{0.60}$ & 39.24 \\
 & Con.C & 8.15$_{0.26}$ & 53.72$_{0.30}$ & 54.26$_{0.37}$ & \textbf{28.72$_{1.91}$} & 35.98$_{0.00}$ & 31.40$_{1.94}$ & 37.56$_{0.00}$ & 39.33$_{0.53}$ & 38.80$_{0.46}$ & 91.52$_{0.38}$ & 41.94 \\
 & Bat.C & 17.58$_{1.19}$ & 53.06$_{0.52}$ & 54.15$_{0.28}$ & 25.62$_{0.85}$ & 51.85$_{3.70}$ & 30.66$_{1.48}$ & 48.45$_{2.63}$ & 38.44$_{0.46}$ & 36.28$_{0.37}$ & 88.11$_{0.62}$ & 44.42 \\
 & Dom.C & 8.03$_{0.00}$ & 52.26$_{0.43}$ & 54.26$_{0.47}$ & 26.50$_{0.99}$ & 36.44$_{1.05}$ & 29.28$_{2.06}$ & 37.56$_{0.00}$ & \textbf{39.72$_{0.41}$} & 35.21$_{0.86}$ & \textbf{91.91$_{0.49}$} & 41.12 \\ \cmidrule(l){2-13} 
 & KNN & \textbf{69.41$_{4.38}$} & 59.12$_{4.16}$ & 62.69$_{3.74}$ & 23.15$_{2.09}$ & \textbf{54.86$_{4.00}$} & \textbf{34.89$_{2.06}$} & 48.44$_{3.62}$ & 36.27$_{4.36}$ & \textbf{50.27$_{5.38}$} & 89.60$_{1.32}$ & \textbf{52.87} \\
 & Cent.C & \textbf{60.04$_{8.07}$} & \textbf{61.80$_{1.44}$} & \textbf{66.16$_{2.05}$} & 23.17$_{1.31}$ & 53.64$_{3.23}$ & 32.16$_{2.62}$ & \textbf{51.74$_{1.94}$} & 30.29$_{3.91}$ & 36.55$_{6.71}$ & 90.12$_{0.52}$ & 50.57 \\
\multirow{-7}{*}{$k=8$} & \cellcolor[HTML]{EFEFEF}\textbf{Hidd.C} & \cellcolor[HTML]{EFEFEF}59.17$_{4.08}$ & \cellcolor[HTML]{EFEFEF}\textbf{65.88$_{0.84}$} & \cellcolor[HTML]{EFEFEF}\textbf{70.07$_{2.32}$} & \cellcolor[HTML]{EFEFEF}\textbf{27.85$_{2.56}$} & \cellcolor[HTML]{EFEFEF}\textbf{55.79$_{3.05}$} & \cellcolor[HTML]{EFEFEF}\textbf{46.70$_{5.37}$} & \cellcolor[HTML]{EFEFEF}\textbf{53.95$_{2.36}$} & \cellcolor[HTML]{EFEFEF}\textbf{43.90$_{0.48}$} & \cellcolor[HTML]{EFEFEF}\textbf{67.69$_{2.29}$} & \cellcolor[HTML]{EFEFEF}\textbf{92.26$_{0.63}$} & \cellcolor[HTML]{EFEFEF}\textbf{58.33} \\ \bottomrule
\end{tabular}}
\label{table:Appendix.LLAMA22}
\end{sidewaystable*}

\begin{sidewaystable*}[t]
\centering
\caption{Classification performance (Macro F1(\%)) on Llama 3-8B. $\mathrm{mean}_{\mathrm{std}}$, top-2 results are in \textbf{bold}.}
\resizebox{0.85\textwidth}{!}{
\begin{tabular}{@{}ccccccccccccc@{}}
\toprule
\multicolumn{2}{c}{Llama 3} & \textbf{AGNews} & \textbf{SemE.R} & \textbf{SemE.L} & \textbf{PoemS} & \textbf{RTE} & \textbf{TEE} & \textbf{TEH} & \textbf{TES} & \textbf{Fina.P} & \textbf{Rott.T} & \textbf{Average} \\ \midrule
 & None & 19.55$_{0.00}$ & 27.61$_{0.00}$ & 24.38$_{0.00}$ & 11.25$_{0.00}$ & 43.68$_{0.00}$ & 26.33$_{0.00}$ & 46.69$_{0.00}$ & 30.67$_{0.00}$ & 44.45$_{0.00}$ & 61.19$_{0.00}$ & 33.58 \\
 & Con.C & 23.20$_{0.00}$ & 17.16$_{0.00}$ & 14.55$_{0.00}$ & 2.33$_{0.00}$ & \textbf{47.50$_{0.00}$} & 21.66$_{0.00}$ & 33.34$_{0.00}$ & 29.66$_{0.00}$ & 28.78$_{0.00}$ & 60.71$_{0.00}$ & 27.89 \\
 & Bat.C & 26.59$_{0.00}$ & 43.52$_{0.00}$ & 38.62$_{0.00}$ & 17.40$_{0.00}$ & 47.25$_{0.00}$ & \textbf{32.02$_{0.00}$} & \textbf{49.12$_{0.00}$} & \textbf{38.17$_{0.00}$} & \textbf{50.32$_{0.00}$} & \textbf{67.19$_{0.00}$} & 41.02 \\
 & Dom.C & 19.17$_{0.00}$ & 25.92$_{0.00}$ & 24.36$_{0.00}$ & 3.97$_{0.00}$ & 32.86$_{0.00}$ & 26.46$_{0.00}$ & 39.35$_{0.00}$ & \textbf{37.95$_{0.00}$} & \textbf{48.39$_{0.00}$} & 59.50$_{0.00}$ & 31.79 \\
\cmidrule(l){2-13} 
 & KNN & \textbf{49.70$_{0.00}$} & 51.20$_{0.00}$ & 47.55$_{0.00}$ & 20.74$_{0.00}$ & 47.25$_{0.00}$ & 28.97$_{0.00}$ & 51.08$_{0.00}$ & 31.26$_{0.00}$ & 47.80$_{0.00}$ & 52.64$_{0.00}$ & \textbf{42.82} \\
 & Cent.C & 39.40$_{0.00}$ & \textbf{53.41$_{0.00}$} & \textbf{48.67$_{0.00}$} & \textbf{23.68$_{0.00}$} & \textbf{47.41$_{0.00}$} & 27.00$_{0.00}$ & \textbf{49.58$_{0.00}$} & 25.19$_{0.00}$ & 40.38$_{0.00}$ & 51.14$_{0.00}$ & 40.58 \\
\multirow{-7}{*}{$k=0$} & \cellcolor[HTML]{EFEFEF}\textbf{Hidd.C} & \cellcolor[HTML]{EFEFEF}\textbf{81.52$_{0.00}$} & \cellcolor[HTML]{EFEFEF}\textbf{52.75$_{0.00}$} & \cellcolor[HTML]{EFEFEF}\textbf{49.43$_{0.00}$} & \cellcolor[HTML]{EFEFEF}\textbf{27.67$_{0.00}$} & \cellcolor[HTML]{EFEFEF}44.53$_{0.00}$ & \cellcolor[HTML]{EFEFEF}\textbf{34.04$_{0.00}$} & \cellcolor[HTML]{EFEFEF}44.53$_{0.00}$ & \cellcolor[HTML]{EFEFEF}37.20$_{0.00}$ & \cellcolor[HTML]{EFEFEF}47.74$_{0.00}$ & \cellcolor[HTML]{EFEFEF}\textbf{63.04$_{0.00}$} & \cellcolor[HTML]{EFEFEF}\textbf{48.25 }\\
\midrule
 & None & 19.97$_{0.70}$ & 53.47$_{0.13}$ & 53.75$_{0.74}$ & \textbf{25.41$_{1.06}$} & 48.60$_{2.81}$ & 17.02$_{1.72}$ & \textbf{50.68$_{2.19}$} & 28.30$_{1.50}$ & 39.28$_{1.36}$ & 81.04$_{2.00}$ & 41.75 \\
 & Con.C & 15.57$_{1.02}$ & 53.37$_{0.69}$ & 54.42$_{0.44}$ & 20.37$_{5.23}$ & 48.60$_{3.06}$ & 19.27$_{1.28}$ & 49.71$_{3.07}$ & 37.32$_{1.11}$ & 39.07$_{0.33}$ & \textbf{90.32$_{0.74}$} & 42.80 \\
 & Bat.C & 27.50$_{1.91}$ & 52.87$_{0.17}$ & 55.71$_{0.60}$ & 23.73$_{0.65}$ & 49.23$_{2.95}$ & 26.48$_{0.99}$ & 49.80$_{2.77}$ & \textbf{39.54$_{2.26}$} & 39.06$_{0.76}$ & 87.01$_{0.79}$ & 45.09 \\
 & Dom.C & 12.96$_{1.62}$ & 52.23$_{1.07}$ & 54.41$_{0.56}$ & 19.94$_{4.35}$ & \textbf{49.37$_{3.03}$} & 20.24$_{0.86}$ & 49.73$_{3.05}$ & 37.86$_{1.05}$ & 36.40$_{0.36}$ & \textbf{90.27$_{0.56}$} & 42.34 \\
\cmidrule(l){2-13} 
 & KNN & \textbf{61.64$_{0.94}$} & 54.58$_{3.37}$ & 61.94$_{1.25}$ & 21.49$_{1.40}$ & 48.15$_{2.60}$ & 34.73$_{2.96}$ & 48.13$_{2.61}$ & 39.11$_{1.54}$ & \textbf{43.88$_{1.85}$} & 87.67$_{1.78}$ & \textbf{50.13} \\
 & Cent.C & 60.16$_{4.68}$ & \textbf{56.81$_{5.45}$} & \textbf{62.72$_{1.87}$} & 22.74$_{3.75}$ & 45.77$_{4.80}$ & \textbf{35.10$_{3.40}$} & 44.59$_{5.76}$ & 35.95$_{2.01}$ & 38.15$_{2.52}$ & 88.06$_{1.14}$ & 49.00 \\
\multirow{-7}{*}{$k=1$} & \cellcolor[HTML]{EFEFEF}\textbf{Hidd.C} & \cellcolor[HTML]{EFEFEF}\textbf{83.37$_{2.76}$} & \cellcolor[HTML]{EFEFEF}\textbf{60.59$_{5.42}$} & \cellcolor[HTML]{EFEFEF}\textbf{64.12$_{1.73}$} & \cellcolor[HTML]{EFEFEF}\textbf{27.82$_{1.99}$} & \cellcolor[HTML]{EFEFEF}\textbf{49.24$_{1.88}$} & \cellcolor[HTML]{EFEFEF}\textbf{53.37$_{3.10}$} & \cellcolor[HTML]{EFEFEF}\textbf{51.77$_{2.59}$} & \cellcolor[HTML]{EFEFEF}\textbf{41.87$_{1.07}$} & \cellcolor[HTML]{EFEFEF}\textbf{65.62$_{5.94}$} & \cellcolor[HTML]{EFEFEF}90.17$_{0.58}$ & \cellcolor[HTML]{EFEFEF}\textbf{58.79} \\

\midrule
 & None & 9.75$_{0.22}$ & 53.90$_{0.25}$ & 54.36$_{0.52}$ & \textbf{24.45$_{1.23}$} & 37.09$_{1.14}$ & 14.54$_{0.00}$ & 39.42$_{1.76}$ & 33.23$_{1.55}$ & 40.28$_{0.18}$ & 84.80$_{0.93}$ & 39.18 \\
 & Con.C & 9.17$_{0.00}$ & 53.72$_{0.64}$ & 54.42$_{0.51}$ & \textbf{23.58$_{4.38}$} & 37.20$_{1.21}$ & 15.56$_{0.72}$ & 43.43$_{4.03}$ & 38.31$_{1.40}$ & 40.17$_{1.84}$ & \textbf{92.71$_{0.63}$} & 40.83 \\
 & Bat.C & 23.64$_{0.50}$ & 52.70$_{0.67}$ & 54.35$_{0.33}$ & 22.95$_{1.60}$ & 46.73$_{5.45}$ & 27.62$_{1.01}$ & \textbf{51.46$_{0.44}$} & \textbf{38.65$_{0.99}$} & 37.11$_{0.41}$ & 90.51$_{1.13}$ & 44.57 \\
 & Dom.C & 9.17$_{0.01}$ & 52.51$_{0.47}$ & 54.35$_{0.38}$ & 21.85$_{1.37}$ & 37.52$_{1.69}$ & 15.94$_{0.73}$ & 43.01$_{5.37}$ & 38.04$_{1.01}$ & 37.76$_{0.82}$ & \textbf{92.57$_{0.48}$} & 40.27 \\
\cmidrule(l){2-13} 
 & KNN & \textbf{74.94$_{3.56}$} & \textbf{54.73$_{5.26}$} & \textbf{59.53$_{2.92}$ }& 23.01$_{3.28}$ & \textbf{49.54$_{5.90}$} & \textbf{40.01$_{1.41}$} & 49.43$_{3.14}$ & 37.00$_{4.50}$ & 42.38$_{2.07}$ & 86.28$_{1.79}$ & \textbf{51.69} \\
 & Cent.C & 73.26$_{3.26}$ & 54.31$_{3.12}$ & 61.26$_{4.08}$ & 21.02$_{2.43}$ & 45.64$_{10.22}$ & 38.73$_{5.55}$ & 50.74$_{5.16}$ & 37.05$_{3.14}$ & \textbf{44.61$_{2.96}$} & 85.45$_{2.18}$ & 51.21 \\
\multirow{-7}{*}{$k=4$} & \cellcolor[HTML]{EFEFEF}\textbf{Hidd.C} & \cellcolor[HTML]{EFEFEF}\textbf{84.53$_{1.35}$} & \cellcolor[HTML]{EFEFEF}\textbf{61.73$_{4.56}$} & \cellcolor[HTML]{EFEFEF}\textbf{69.28$_{1.43}$} & \cellcolor[HTML]{EFEFEF}23.03$_{1.82}$ & \cellcolor[HTML]{EFEFEF}\textbf{53.49$_{3.61}$} & \cellcolor[HTML]{EFEFEF}\textbf{54.70$_{2.03}$} & \cellcolor[HTML]{EFEFEF}\textbf{52.96$_{3.84}$} & \cellcolor[HTML]{EFEFEF}\textbf{40.71$_{2.35}$} & \cellcolor[HTML]{EFEFEF}\textbf{72.18$_{3.19}$} & \cellcolor[HTML]{EFEFEF}92.18$_{1.39}$ & \cellcolor[HTML]{EFEFEF}\textbf{60.48} \\

\midrule
 & None & - & 53.77$_{0.42}$ & 54.82$_{0.12}$ & \textbf{27.10$_{1.62}$} & 35.98$_{0.00}$ & 14.54$_{0.00}$ & 37.69$_{0.27}$ & 35.96$_{1.51}$ & 42.65$_{0.85}$ & 88.40$_{0.59}$ & 39.09 \\
 & Con.C & - & 53.52$_{0.53}$ & 54.42$_{0.53}$ & \textbf{25.07$_{4.81}$} & 35.98$_{0.00}$ & 14.54$_{0.00}$ & 47.99$_{9.84}$ & 38.74$_{1.35}$ & 41.19$_{2.12}$ & \textbf{92.81$_{1.09}$} & 40.43 \\
 & Bat.C & - & 52.40$_{0.72}$ & 54.82$_{0.35}$ & 24.68$_{0.67}$ & \textbf{60.38$_{5.04}$} & 31.50$_{0.44}$ & \textbf{55.00$_{0.54}$} & 38.59$_{0.78}$ & 37.95$_{0.35}$ & 91.59$_{0.53}$ & 44.69 \\
 & Dom.C & - & 52.03$_{0.35}$ & 54.11$_{0.66}$ & 21.95$_{0.83}$ & 35.98$_{0.00}$ & 14.67$_{0.22}$ & 45.15$_{9.11}$ & 38.61$_{1.13}$ & 37.30$_{1.39}$ & 93.11$_{0.74}$ & 39.29 \\
\cmidrule(l){2-13} 
 & KNN & - & 54.47$_{2.60}$ & \textbf{58.17$_{2.84}$} & 21.49$_{1.69}$ & 50.75$_{4.92}$ & \textbf{47.22$_{2.50}$} & 51.75$_{2.65}$ & \textbf{41.66$_{3.19}$} & \textbf{46.47$_{4.53}$} & 84.53$_{1.91}$ & \textbf{45.65} \\
 & Cent.C & - & \textbf{56.03$_{3.55}$} & 49.78$_{5.61}$ & 22.15$_{1.36}$ & 48.30$_{9.65}$ & 45.41$_{6.52}$ & 48.34$_{4.57}$ & 38.65$_{1.60}$ & 39.58$_{7.11}$ & 85.23$_{2.05}$ & 43.35 \\
\multirow{-7}{*}{$k=8$} & \cellcolor[HTML]{EFEFEF}\textbf{Hidd.C} & \cellcolor[HTML]{EFEFEF}- & \cellcolor[HTML]{EFEFEF}\textbf{63.34$_{1.61}$} & \cellcolor[HTML]{EFEFEF}\textbf{67.48$_{1.00}$} & \cellcolor[HTML]{EFEFEF}23.84$_{1.22}$ & \cellcolor[HTML]{EFEFEF}\textbf{60.89$_{3.38}$} & \cellcolor[HTML]{EFEFEF}\textbf{56.67$_{7.36}$} & \cellcolor[HTML]{EFEFEF}\textbf{53.45$_{4.92}$} & \cellcolor[HTML]{EFEFEF}\textbf{39.74$_{2.66}$} & \cellcolor[HTML]{EFEFEF}\textbf{79.28$_{4.45}$} & \cellcolor[HTML]{EFEFEF}\textbf{91.64$_{1.59}$} & \cellcolor[HTML]{EFEFEF}\textbf{53.63}
\\ \bottomrule
\end{tabular}}
\label{table:Appendix.LLAMA3}
\end{sidewaystable*}

\begin{sidewaystable*}[t]
\centering
\caption{Classification performance (Macro F1(\%)) on Llama 2-13B. $\mathrm{mean}_{\mathrm{std}}$, top-2 results are in \textbf{bold}.}
\resizebox{0.85\textwidth}{!}{
\begin{tabular}{@{}ccccccccccccc@{}}
\toprule
\multicolumn{2}{c}{Llama 2} & \textbf{AGNews} & \textbf{SemE.R} & \textbf{SemE.L} & \textbf{PoemS} & \textbf{RTE} & \textbf{TEE} & \textbf{TEH} & \textbf{TES} & \textbf{Fina.P} & \textbf{Rott.T} & \textbf{Average} \\ \midrule
 & None & 9.61$_{0.00}$ & 47.67$_{0.00}$ & 51.12$_{0.00}$ & 8.01$_{0.00}$ & 36.04$_{0.00}$ & 31.92$_{0.00}$ & 37.56$_{0.00}$ & 46.19$_{0.00}$ & 30.94$_{0.00}$ & 54.98$_{0.00}$ & 35.40 \\
 & Con.C & 22.36$_{0.00}$ & 48.44$_{0.00}$ & \textbf{57.96$_{0.00}$} & 15.71$_{0.00}$ & 41.45$_{0.00}$ & 23.57$_{0.00}$ & 37.56$_{0.00}$ & 39.59$_{0.00}$ & 44.97$_{0.00}$ & 42.61$_{0.00}$ & 37.42 \\
 & Bat.C & 30.84$_{0.00}$ & \textbf{52.16$_{0.00}$} & \textbf{61.45$_{0.00}$} & 20.50$_{0.00}$ & 43.54$_{0.00}$ & \textbf{37.88$_{0.00}$} & 43.73$_{0.00}$ & \textbf{47.31$_{0.00}$} & \textbf{50.83$_{0.00}$} & \textbf{61.26$_{0.00}$} & \textbf{44.95} \\
 & Dom.C & 23.21$_{0.00}$ & 46.54$_{0.00}$ & 51.51$_{0.00}$ & 8.88$_{0.00}$ & 41.87$_{0.00}$ & 29.11$_{0.00}$ & 37.56$_{0.00}$ & \textbf{47.22$_{0.00}$} & 31.98$_{0.00}$ & \textbf{60.59$_{0.00}$} & 37.85 \\
\cmidrule(l){2-13} 
 & KNN & \textbf{46.44$_{0.00}$} & 42.06$_{0.00}$ & 43.18$_{0.00}$ & \textbf{24.03$_{0.00}$} & \textbf{50.33$_{0.00}$} & 28.36$_{0.00}$ & \textbf{55.76$_{0.00}$} & 35.27$_{0.00}$ & 45.74$_{0.00}$ & 51.95$_{0.00}$ & 42.31 \\
 & Cent.C & 43.48$_{0.00}$ & 44.64$_{0.00}$ & 41.05$_{0.00}$ & 21.23$_{0.00}$ & 34.02$_{0.00}$ & 30.41$_{0.00}$ & 49.02$_{0.00}$ & 30.44$_{0.00}$ & 43.17$_{0.00}$ & 52.75$_{0.00}$ & 39.02 \\
\multirow{-7}{*}{$k=0$} & \cellcolor[HTML]{EFEFEF}\textbf{Hidd.C} & \cellcolor[HTML]{EFEFEF}\textbf{69.78$_{0.00}$} & \cellcolor[HTML]{EFEFEF}\textbf{54.08$_{0.00}$ }& \cellcolor[HTML]{EFEFEF}50.67$_{0.00}$ & \cellcolor[HTML]{EFEFEF}\textbf{33.01$_{0.00}$} & \cellcolor[HTML]{EFEFEF}\textbf{54.41$_{0.00}$} & \cellcolor[HTML]{EFEFEF}\textbf{41.23$_{0.00}$} & \cellcolor[HTML]{EFEFEF}\textbf{51.14$_{0.00}$} & \cellcolor[HTML]{EFEFEF}35.45$_{0.00}$ & \cellcolor[HTML]{EFEFEF}\textbf{48.87$_{0.00}$} & \cellcolor[HTML]{EFEFEF}57.61$_{0.00}$ & \cellcolor[HTML]{EFEFEF}\textbf{49.63} \\

\midrule
 & None & 9.17$_{0.00}$ & 50.67$_{0.80}$ & 53.51$_{0.08}$ & 15.49$_{0.67}$ & 35.11$_{0.00}$ & 33.79$_{1.09}$ & 37.56$_{0.00}$ & 37.11$_{0.78}$ & 34.73$_{0.90}$ & 81.61$_{0.84}$ & 38.88 \\
 & Con.C & 9.17$_{0.00}$ & 53.38$_{0.30}$ & \textbf{54.04$_{0.20}$} & \textbf{25.20$_{0.64}$} & 35.11$_{0.00}$ & 33.61$_{0.46}$ & 37.52$_{0.05}$ & \textbf{38.00$_{0.53}$} & 35.73$_{0.58}$ & \textbf{87.40$_{1.46}$} & 40.92 \\
 & Bat.C & 24.81$_{2.72}$ & 49.21$_{0.87}$ & 53.72$_{0.14}$ & 23.39$_{0.33}$ & 47.24$_{1.16}$ & \textbf{35.68$_{0.19}$} & \textbf{51.66$_{1.75}$} & 36.21$_{0.81}$ & 34.35$_{0.39}$ & 85.99$_{1.19}$ & 44.23 \\
 & Dom.C & 9.55$_{0.38}$ & 51.02$_{0.55}$ & 52.92$_{0.29}$ & 19.66$_{1.43}$ & 35.11$_{0.00}$ & 32.55$_{2.39}$ & 37.56$_{0.00}$ & 38.43$_{0.72}$ & 35.55$_{0.68}$ & \textbf{86.42$_{1.75}$} & 39.88 \\
\cmidrule(l){2-13} 
 & KNN & \textbf{58.35$_{3.83}$} & \textbf{54.98$_{5.17}$} & 48.86$_{0.44}$ & 24.03$_{1.43}$ & \textbf{47.50$_{2.83}$} & 32.11$_{1.50}$ & 50.95$_{3.28}$ & 33.33$_{4.95}$ & \textbf{37.93$_{3.72}$} & 82.19$_{1.11}$ & \textbf{47.02} \\
 & Cent.C & 51.96$_{0.88}$ & 49.83$_{4.34}$ & 46.51$_{2.10}$ & 25.20$_{2.94}$ & 43.15$_{3.01}$ & 27.11$_{1.80}$ & 47.81$_{6.69}$ & 31.46$_{6.15}$ & 31.22$_{4.39}$ & 82.31$_{1.10}$ & 43.65 \\
\multirow{-7}{*}{$k=1$} & \cellcolor[HTML]{EFEFEF}\textbf{Hidd.C} & \cellcolor[HTML]{EFEFEF}\textbf{76.45$_{5.43}$} & \cellcolor[HTML]{EFEFEF}\textbf{64.70$_{1.97}$} & \cellcolor[HTML]{EFEFEF}\textbf{65.70$_{0.16}$} & \cellcolor[HTML]{EFEFEF}\textbf{30.22$_{3.11}$} & \cellcolor[HTML]{EFEFEF}\textbf{49.42$_{1.78}$} & \cellcolor[HTML]{EFEFEF}\textbf{48.97$_{0.32}$} & \cellcolor[HTML]{EFEFEF}\textbf{52.81$_{4.66}$} & \cellcolor[HTML]{EFEFEF}\textbf{41.05$_{4.32}$} & \cellcolor[HTML]{EFEFEF}\textbf{59.29$_{4.43}$} & \cellcolor[HTML]{EFEFEF}84.02$_{0.88}$ & \cellcolor[HTML]{EFEFEF}\textbf{57.26} \\

\midrule
 & None & 9.17$_{0.00}$ & 49.99$_{0.32}$ & 53.69$_{0.44}$ & 19.29$_{0.86}$ & 35.11$_{0.00}$ & \textbf{36.23$_{2.22}$} & 37.56$_{0.00}$ & 37.63$_{0.58}$ & 36.02$_{0.42}$ & 91.82$_{0.41}$ & 40.65 \\
 & Con.C & 9.37$_{0.22}$ & 53.12$_{0.62}$ & \textbf{54.25$_{0.26}$} & \textbf{26.98$_{1.34}$} & 35.11$_{0.00}$ & 34.64$_{2.69}$ & 37.56$_{0.00}$ & 37.64$_{1.28}$ & 36.38$_{0.15}$ & \textbf{92.52$_{0.76}$} & 41.76 \\
 & Bat.C & 19.64$_{1.36}$ & 49.33$_{0.12}$ & 53.69$_{0.13}$ & 23.93$_{0.41}$ & 46.56$_{3.27}$ & 34.29$_{1.84}$ & 46.95$_{1.79}$ & 35.92$_{0.83}$ & 34.14$_{0.46}$ & 91.68$_{0.52}$ & 43.61 \\
 & Dom.C & 9.56$_{0.31}$ & 49.61$_{0.38}$ & 53.79$_{0.35}$ & 16.85$_{0.80}$ & 35.11$_{0.00}$ & 33.62$_{1.59}$ & 37.56$_{0.00}$ & \textbf{38.20$_{1.38}$} & 35.95$_{0.35}$ & \textbf{92.46$_{0.83}$} & 40.27 \\
\cmidrule(l){2-13} 
 & KNN & \textbf{76.62$_{0.69}$} & \textbf{53.18$_{1.01}$} & 53.38$_{4.35}$ & 22.73$_{1.17}$ & 47.94$_{1.65}$ & 35.09$_{5.00}$ & \textbf{50.17$_{1.01}$} & 38.26$_{1.25}$ & 40.40$_{2.69}$ & 83.18$_{3.32}$ & \textbf{50.09} \\
 & Cent.C & 74.72$_{7.17}$ & 52.52$_{2.90}$ & 48.20$_{3.83}$ & 22.08$_{2.73}$ & \textbf{48.59$_{2.55}$} & 28.15$_{3.24}$ & 49.53$_{1.53}$ & 29.35$_{4.83}$ & \textbf{43.69$_{5.11}$} & 82.73$_{1.48}$ & 47.96 \\
\multirow{-7}{*}{$k=4$} & \cellcolor[HTML]{EFEFEF}\textbf{Hidd.C} & \cellcolor[HTML]{EFEFEF}\textbf{80.76$_{4.75}$} & \cellcolor[HTML]{EFEFEF}\textbf{66.91$_{1.22}$} & \cellcolor[HTML]{EFEFEF}\textbf{68.44$_{2.54}$} & \cellcolor[HTML]{EFEFEF}\textbf{25.39$_{1.27}$} & \cellcolor[HTML]{EFEFEF}\textbf{52.69$_{3.96}$} & \cellcolor[HTML]{EFEFEF}\textbf{53.41$_{3.40}$} & \cellcolor[HTML]{EFEFEF}\textbf{52.29$_{4.93}$} & \cellcolor[HTML]{EFEFEF}\textbf{44.47$_{4.05}$} & \cellcolor[HTML]{EFEFEF}\textbf{59.48$_{1.42}$} & \cellcolor[HTML]{EFEFEF}91.36$_{0.79}$ & \cellcolor[HTML]{EFEFEF}\textbf{59.52} \\

\midrule
 & None & 9.17$_{0.00}$ & 50.01$_{0.25}$ & 53.82$_{0.37}$ & 19.54$_{0.37}$ & 35.11$_{0.00}$ & 36.93$_{1.76}$ & 37.56$_{0.00}$ & 36.64$_{0.75}$ & 36.49$_{0.04}$ & \textbf{92.80$_{0.63}$} & 40.81 \\
 & Con.C & 9.17$_{0.00}$ & \textbf{52.89$_{0.18}$} & \textbf{54.31$_{0.29}$ }& \textbf{29.11$_{0.49}$} & 35.11$_{0.00}$ & 36.44$_{1.99}$ & 37.56$_{0.00}$ & 38.24$_{0.79}$ & 38.10$_{0.70}$ & 92.72$_{0.85}$ & 42.36 \\
 & Bat.C & 17.94$_{0.26}$ & 48.95$_{0.69}$ & 53.64$_{0.45}$ & 24.35$_{0.44}$ & \textbf{52.35$_{3.27}$} & 34.94$_{2.46}$ & 47.18$_{3.05}$ & 36.11$_{0.40}$ & 35.04$_{0.09}$ & 92.23$_{0.64}$ & 44.27 \\
 & Dom.C & 9.17$_{0.00}$ & 50.55$_{0.37}$ & 53.95$_{0.41}$ & 18.12$_{1.11}$ & 35.11$_{0.00}$ & 34.14$_{2.53}$ & 37.56$_{0.00}$ & 38.43$_{0.17}$ & 35.70$_{0.27}$ & \textbf{92.91$_{0.56}$} & 40.56 \\
\cmidrule(l){2-13} 
 & KNN & \textbf{79.57$_{4.10}$} & 52.03$_{2.71}$ & 51.68$_{8.67}$ & 23.48$_{0.62}$ & 50.35$_{3.07}$ & \textbf{39.90$_{6.24}$} & \textbf{53.54$_{4.10}$} & \textbf{41.95$_{2.23}$} & 50.11$_{3.56}$ & 86.47$_{1.81}$ & \textbf{52.91} \\
 & Cent.C & 76.52$_{2.63}$ & 51.46$_{5.58}$ & 42.86$_{5.16}$ & 22.13$_{4.60}$ & 49.28$_{0.85}$ & 34.97$_{5.39}$ & 53.09$_{4.32}$ & 33.61$_{7.13}$ & \textbf{50.29$_{2.71}$} & 84.48$_{2.15}$ & 49.87 \\
\multirow{-7}{*}{$k=8$} & \cellcolor[HTML]{EFEFEF}\textbf{Hidd.C} & \cellcolor[HTML]{EFEFEF}\textbf{80.47$_{2.80}$} & \cellcolor[HTML]{EFEFEF}\textbf{66.49$_{2.19}$} & \cellcolor[HTML]{EFEFEF}\textbf{62.69$_{8.43}$} & \cellcolor[HTML]{EFEFEF}\textbf{25.58$_{3.44}$} & \cellcolor[HTML]{EFEFEF}\textbf{53.02$_{4.91}$} & \cellcolor[HTML]{EFEFEF}\textbf{50.88$_{5.15}$} & \cellcolor[HTML]{EFEFEF}\textbf{57.19$_{7.28}$} & \cellcolor[HTML]{EFEFEF}\textbf{41.99$_{3.69}$} & \cellcolor[HTML]{EFEFEF}\textbf{69.04$_{5.48}$} & \cellcolor[HTML]{EFEFEF}91.31$_{0.25}$ & \cellcolor[HTML]{EFEFEF}\textbf{59.87}
\\ \bottomrule
\end{tabular}}
\label{table:Appendix.LLAMA213B}
\end{sidewaystable*}

\begin{sidewaystable*}[t]
\centering
\caption{Classification performance (Macro F1(\%)) on Llama 2-34B. $\mathrm{mean}_{\mathrm{std}}$, top-2 results are in \textbf{bold}.}
\resizebox{0.85\textwidth}{!}{
\begin{tabular}{@{}ccccccccccccc@{}}
\toprule
\multicolumn{2}{c}{Llama 2} & \textbf{AGNews} & \textbf{SemE.R} & \textbf{SemE.L} & \textbf{PoemS} & \textbf{RTE} & \textbf{TEE} & \textbf{TEH} & \textbf{TES} & \textbf{Fina.P} & \textbf{Rott.T} & \textbf{Average} \\ \midrule
 & None & 18.28$_{0.00}$ & 50.09$_{0.00}$ & 51.30$_{0.00}$ & 15.50$_{0.00}$ & 39.75$_{0.00}$ & 24.91$_{0.00}$ & 40.14$_{0.00}$ & 31.05$_{0.00}$ & 34.67$_{0.00}$ & \textbf{71.67$_{0.00}$} & 37.73 \\
 & Con.C & 23.06$_{0.00}$ & 50.75$_{0.00}$ & \textbf{58.71$_{0.00}$} & 8.81$_{0.00}$ & 35.11$_{0.00}$ & 30.96$_{0.00}$ & 44.55$_{0.00}$ & 34.93$_{0.00}$ & 48.68$_{0.00}$ & 69.32$_{0.00}$ & 40.49 \\
 & Bat.C & 26.52$_{0.00}$ & \textbf{54.51$_{0.00}$} & \textbf{60.19$_{0.00}$} & 18.91$_{0.00}$ & 48.47$_{0.00}$ & 31.62$_{0.00}$ & 45.90$_{0.00}$ & \textbf{40.37$_{0.00}$} & \textbf{53.45$_{0.00}$} & 70.69$_{0.00}$ & \textbf{45.06} \\
 & Dom.C & 22.58$_{0.00}$ & 52.91$_{0.00}$ & 55.96$_{0.00}$ & 24.06$_{0.00}$ & 47.61$_{0.00}$ & 31.18$_{0.00}$ & 37.30$_{0.00}$ & 36.45$_{0.00}$ & 50.42$_{0.00}$ & \textbf{71.77$_{0.00}$} & 43.02 \\
\cmidrule(l){2-13} 
 & KNN & 31.79$_{0.00}$ & 45.85$_{0.00}$ & 45.03$_{0.00}$ & \textbf{27.38$_{0.00}$} & \textbf{53.04$_{0.00}$} & 31.64$_{0.00}$ & \textbf{55.29$_{0.00}$} & 34.10$_{0.00}$ & 46.57$_{0.00}$ & 57.60$_{0.00}$ & \textbf{42.83} \\
 & Cent.C & \textbf{35.28$_{0.00}$} & 50.83$_{0.00}$ & 51.72$_{0.00}$ & 23.44$_{0.00}$ & 51.15$_{0.00}$ & \textbf{32.02$_{0.00}$} & 53.12$_{0.00}$ & 35.33$_{0.00}$ & 44.25$_{0.00}$ & 48.47$_{0.00}$ & 42.56 \\
\multirow{-7}{*}{$k=0$} & \cellcolor[HTML]{EFEFEF}\textbf{Hidd.C} & \cellcolor[HTML]{EFEFEF}\textbf{65.31$_{0.00}$} & \cellcolor[HTML]{EFEFEF}\textbf{56.22$_{0.00}$} & \cellcolor[HTML]{EFEFEF}54.83$_{0.00}$ & \cellcolor[HTML]{EFEFEF}\textbf{31.82$_{0.00}$} & \cellcolor[HTML]{EFEFEF}\textbf{54.76$_{0.00}$} & \cellcolor[HTML]{EFEFEF}\textbf{50.49$_{0.00}$} & \cellcolor[HTML]{EFEFEF}\textbf{65.71$_{0.00}$} & \cellcolor[HTML]{EFEFEF}\textbf{45.46$_{0.00}$} & \cellcolor[HTML]{EFEFEF}\textbf{69.22$_{0.00}$} & \cellcolor[HTML]{EFEFEF}65.14$_{0.00}$ & \cellcolor[HTML]{EFEFEF}\textbf{55.90} \\

\midrule
 & None & 9.59$_{0.03}$ & 53.48$_{0.49}$ & 54.97$_{0.04}$ & 22.51$_{0.39}$ & 35.53$_{0.18}$ & \textbf{33.32$_{1.18}$} & 37.56$_{0.38}$ & 36.38$_{1.24}$ & 35.41$_{0.18}$ & 81.78$_{0.31}$ & 40.05 \\
 & Con.C & 10.38$_{0.91}$ & 53.97$_{0.80}$ & 54.97$_{0.55}$ & 25.44$_{0.20}$ & 35.11$_{2.42}$ & 32.92$_{0.56}$ & 37.56$_{0.00}$ & 37.08$_{2.18}$ & 38.73$_{0.28}$ & \textbf{85.60$_{0.25}$} & 41.17 \\
 & Bat.C & 21.41$_{0.28}$ & 51.86$_{0.43}$ & 54.81$_{0.10}$ & 23.47$_{1.50}$ & 52.97$_{1.18}$ & 33.02$_{1.00}$ & 48.06$_{1.46}$ & 37.07$_{0.67}$ & 35.23$_{0.21}$ & 84.63$_{0.44}$ & 44.25 \\
 & Dom.C & 10.00$_{0.29}$ & 53.46$_{0.90}$ & 54.96$_{0.11}$ & \textbf{29.23$_{2.93}$} & 37.32$_{0.40}$ & 32.70$_{0.82}$ & 37.56$_{0.00}$ & 37.21$_{1.47}$ & 39.96$_{0.16}$ & \textbf{85.80$_{1.51}$} & 41.82 \\
\cmidrule(l){2-13} 
 & KNN & \textbf{65.38$_{1.25}$} & \textbf{58.91$_{1.13}$} & \textbf{61.28$_{2.68}$} & 29.09$_{4.68}$ & \textbf{62.63$_{3.40}$} & 30.21$_{1.64}$ & 54.69$_{0.85}$ & \textbf{40.78$_{2.03}$} & \textbf{68.78$_{1.13}$} & 68.77$_{3.17}$ & \textbf{54.05} \\
 & Cent.C & 63.03$_{3.16}$ & 55.45$_{2.61}$ & 54.41$_{4.82}$ & 24.23$_{1.29}$ & 62.47$_{0.04}$ & 31.16$_{0.39}$ & \textbf{56.13$_{2.05}$} & 30.55$_{3.73}$ & 55.08$_{3.23}$ & 58.61$_{3.26}$ & 49.11 \\
\multirow{-7}{*}{$k=1$} & \cellcolor[HTML]{EFEFEF}\textbf{Hidd.C} & \cellcolor[HTML]{EFEFEF}\textbf{79.39$_{0.48}$} & \cellcolor[HTML]{EFEFEF}\textbf{67.23$_{2.91}$} & \cellcolor[HTML]{EFEFEF}\textbf{69.03$_{0.53}$} & \cellcolor[HTML]{EFEFEF}\textbf{33.12$_{0.35}$} & \cellcolor[HTML]{EFEFEF}\textbf{74.89$_{1.33}$} & \cellcolor[HTML]{EFEFEF}\textbf{52.50$_{4.09}$} & \cellcolor[HTML]{EFEFEF}\textbf{61.88$_{0.91}$} & \cellcolor[HTML]{EFEFEF}\textbf{46.63$_{1.04}$} & \cellcolor[HTML]{EFEFEF}\textbf{76.83$_{2.01}$} & \cellcolor[HTML]{EFEFEF}73.14$_{0.06}$ & \cellcolor[HTML]{EFEFEF}\textbf{63.46} \\

\midrule
 & None & 9.59$_{0.55}$ & 53.48$_{0.39}$ & 54.97$_{0.64}$ & 22.51$_{0.69}$ & 35.53$_{0.06}$ & \textbf{33.32$_{3.35}$} & 37.56$_{0.00}$ & 36.38$_{1.92}$ & 35.41$_{0.51}$ & 81.78$_{0.55}$ & 40.05 \\
 & Con.C & 10.38$_{0.01}$ & 53.97$_{0.16}$ & 54.97$_{0.61}$ & 25.44$_{0.06}$ & 35.11$_{0.00}$ & 32.92$_{3.78}$ & 37.56$_{0.00}$ & 37.08$_{0.97}$ & 38.73$_{0.55}$ & \textbf{85.60$_{2.37}$} & 41.17 \\
 & Bat.C & 21.41$_{1.08}$ & 51.86$_{0.34}$ & 54.81$_{0.44}$ & 23.47$_{0.33}$ & 52.97$_{0.10}$ & 33.02$_{4.05}$ & 48.06$_{2.65}$ & 37.07$_{1.89}$ & 35.23$_{0.77}$ & 84.63$_{1.97}$ & 44.25 \\
 & Dom.C & 10.00$_{0.55}$ & 53.46$_{0.36}$ & 54.96$_{0.66}$ & \textbf{29.23$_{0.90}$} & 37.32$_{1.28}$ & 32.70$_{4.18}$ & 37.56$_{0.00}$ & 37.21$_{0.80}$ & 39.96$_{0.04}$ & \textbf{85.80$_{2.09}$} & 41.82 \\
\cmidrule(l){2-13} 
 & KNN & \textbf{65.38$_{3.35}$} & \textbf{58.91$_{0.83}$} & \textbf{61.28$_{0.90}$} & 29.09$_{2.60}$ & \textbf{62.63$_{2.71}$} & 30.21$_{3.04}$ & 54.69$_{2.18}$ & \textbf{40.78$_{2.16}$} & \textbf{68.78$_{1.44}$} & 68.77$_{8.22}$ & \textbf{54.05} \\
 & Cent.C & 63.03$_{10.02}$ & 55.45$_{0.33}$ & 54.41$_{2.47}$ & 24.23$_{0.35}$ & 62.47$_{6.66}$ & 31.16$_{2.87}$ & \textbf{56.13$_{0.93}$} & 30.55$_{0.73}$ & 55.08$_{1.63}$ & 58.61$_{1.20}$ & 49.11 \\
\multirow{-7}{*}{$k=4$} & \cellcolor[HTML]{EFEFEF}\textbf{Hidd.C} & \cellcolor[HTML]{EFEFEF}\textbf{79.39$_{3.49}$} & \cellcolor[HTML]{EFEFEF}\textbf{67.23$_{1.13}$} & \cellcolor[HTML]{EFEFEF}\textbf{69.03$_{9.36}$} & \cellcolor[HTML]{EFEFEF}\textbf{33.12$_{2.59}$} & \cellcolor[HTML]{EFEFEF}\textbf{74.89$_{1.75}$} & \cellcolor[HTML]{EFEFEF}\textbf{52.50$_{2.17}$} & \cellcolor[HTML]{EFEFEF}\textbf{61.88$_{1.74}$} & \cellcolor[HTML]{EFEFEF}\textbf{46.63$_{5.67}$} & \cellcolor[HTML]{EFEFEF}\textbf{76.83$_{9.16}$} & \cellcolor[HTML]{EFEFEF}73.14$_{11.06}$ & \cellcolor[HTML]{EFEFEF}\textbf{63.46} \\

\midrule
 & None & 9.18$_{0.01}$ & 53.56$_{0.99}$ & 55.06$_{0.70}$ & 26.48$_{0.16}$ & 35.11$_{0.00}$ & 35.06$_{0.43}$ & 37.56$_{0.00}$ & 37.93$_{0.81}$ & 36.12$_{0.27}$ & 88.65$_{2.13}$ & 41.47 \\
 & Con.C & 11.00$_{1.59}$ & 52.04$_{0.51}$ & 54.55$_{0.11}$ & 27.01$_{0.23}$ & 35.11$_{0.00}$ & 34.60$_{0.17}$ & 37.56$_{0.00}$ & 38.81$_{0.29}$ & 38.68$_{0.74}$ & \textbf{89.87$_{0.41}$} & 41.92 \\
 & Bat.C & 18.25$_{0.02}$ & 52.17$_{0.44}$ & 54.91$_{0.11}$ & 25.89$_{0.60}$ & \textbf{66.77$_{0.63}$} & 35.08$_{0.61}$ & 57.00$_{0.26}$ & 36.73$_{0.86}$ & 35.85$_{0.25}$ & 90.17$_{1.13}$ & 47.28 \\
 & Dom.C & 9.39$_{0.30}$ & 53.02$_{1.61}$ & 55.18$_{0.52}$ & \textbf{32.92$_{1.16}$} & 35.11$_{0.00}$ & 33.10$_{2.08}$ & 37.56$_{0.00}$ & 37.79$_{0.23}$ & 41.86$_{1.38}$ & \textbf{92.08$_{0.41}$} & 42.80 \\
\cmidrule(l){2-13} 
 & KNN & \textbf{63.23$_{1.12}$} & \textbf{60.79$_{0.34}$} & \textbf{61.09$_{4.12}$} & 30.84$_{0.06}$ & 63.67$_{1.20}$ & \textbf{37.53$_{0.81}$} & 55.26$_{5.26}$ & \textbf{42.30$_{4.48}$} & \textbf{80.93$_{0.92}$} & 78.58$_{0.41}$ & \textbf{57.42} \\
 & Cent.C & 57.73$_{6.27}$ & 57.02$_{0.78}$ & 54.73$_{5.85}$ & 26.35$_{2.44}$ & 62.27$_{5.14}$ & 35.94$_{1.68}$ & \textbf{59.23$_{6.62}$} & 38.25$_{6.66}$ & 58.83$_{10.95}$ & 74.64$_{0.80}$ & 52.50 \\
\multirow{-7}{*}{$k=8$} & \cellcolor[HTML]{EFEFEF}\textbf{Hidd.C} & \cellcolor[HTML]{EFEFEF}\textbf{83.00$_{0.52}$} & \cellcolor[HTML]{EFEFEF}\textbf{68.71$_{0.42}$} & \cellcolor[HTML]{EFEFEF}\textbf{77.09$_{0.85}$} & \cellcolor[HTML]{EFEFEF}\textbf{31.59$_{8.39}$} & \cellcolor[HTML]{EFEFEF}\textbf{78.91$_{0.40}$} & \cellcolor[HTML]{EFEFEF}\textbf{60.97$_{0.90}$} & \cellcolor[HTML]{EFEFEF}\textbf{64.20$_{1.65}$} & \cellcolor[HTML]{EFEFEF}\textbf{50.52$_{0.85}$} & \cellcolor[HTML]{EFEFEF}\textbf{91.98$_{0.17}$} & \cellcolor[HTML]{EFEFEF}84.78$_{3.82}$ & \cellcolor[HTML]{EFEFEF}\textbf{69.17} \\ \bottomrule
\end{tabular}}
\label{table:Appendix.LLAMA234B}
\end{sidewaystable*}

\end{document}